\documentclass[10pt,twocolumn,letterpaper]{article}

\usepackage[pagenumbers]{wacv} %

\usepackage{graphicx}
\usepackage{subcaption}
\usepackage{amsmath}
\usepackage{amssymb}
\usepackage{booktabs}
\usepackage{natbib}
\usepackage{array}
\usepackage{graphicx}
\usepackage{booktabs}
\usepackage{makecell}
\usepackage{float}
\usepackage{makecell}
\usepackage{multirow, multicol}
\usepackage{colortbl}
\usepackage{cuted} 
\usepackage{soul}
\usepackage{wrapfig}
\usepackage{tikz}
\usepackage{caption}
\usepackage{float}
\usetikzlibrary{tikzmark, arrows.meta}
\definecolor{wacvblue}{rgb}{0.21,0.49,0.74}
\usepackage[pagebackref,breaklinks,colorlinks,allcolors=wacvblue]{hyperref}

\usepackage[capitalize]{cleveref}
\crefname{section}{Sec.}{Secs.}
\Crefname{section}{Section}{Sections}
\Crefname{table}{Table}{Tables}
\crefname{table}{Tab.}{Tabs.}

\def\wacvPaperID{1867} %
\def\confName{WACV}
\def\confYear{2027}

\begin{document}

\title{Breaking the Privacy Promise: Gradient Inversion Attacks on Federated Low-Rank Adapted Vision Transformers}

\author{
Ritik K. Badiya \quad
Siladittya Manna \quad
Ashmit Sinha \quad
M Yashwanth \quad
Anirban Chakraborty\\
Indian Institute of Science, Bengaluru, India\\
{\tt\small \{ritikbadiya, siladittyam, yashwanthm, anirban\}@iisc.ac.in \quad sin.ashmit@gmail.com}
}

\maketitle

\begin{abstract}
Parameter-efficient fine-tuning (PEFT) has been increasingly adopted to fine-tune pretrained models with low computational cost and is therefore also preferred in the federated learning paradigm to reduce computational and communication overhead. {In this work, we investigate the sparsely explored problem of the risk of private data leakage in LoRA-adapted Vision Transformers (ViTs), where low-rank updates weaken the gradient signal available to a server with access to individual client updates}. We show that quality of reconstruction is limited by two coupled ambiguities: reduced information in LoRA gradients and missing spatial cues from frozen positional embeddings. To address this, we propose a spatially anchored generative inversion framework that incorporates {server-prescribed} trainable positional embeddings into ViT blocks during fine-tuning at clients. The resulting positional gradients act as layer-wise spatial anchors, {providing additional positional cues}, while the generative prior keeps reconstructions on a realistic image manifold, reducing {duplicate}-data and patch-permutation artifacts, {which produce similar gradients as those produced when the target client images are used}. Adding our proposed spatial anchoring to GIAS improves PSNR by 10.14 dB, SSIM by 0.45, and LPIPS by 0.098 on CIFAR-10, as well as PSNR gains of 7.83 and 11.74 dB on high-resolution CelebA and CARS datasets, respectively, in the FedSGD setting. Such gains in reconstructed image quality metrics are also maintained at higher batch sizes and in FedAvg settings. These results indicate that LoRA-based PEFT does not, by itself, prevent data leakage in FL, and carefully incorporated positional embeddings can help in exposing private information.
\end{abstract}

\section{Introduction}

\begin{figure}[!ht] %
    \centering
    \includegraphics[width=0.9\linewidth]{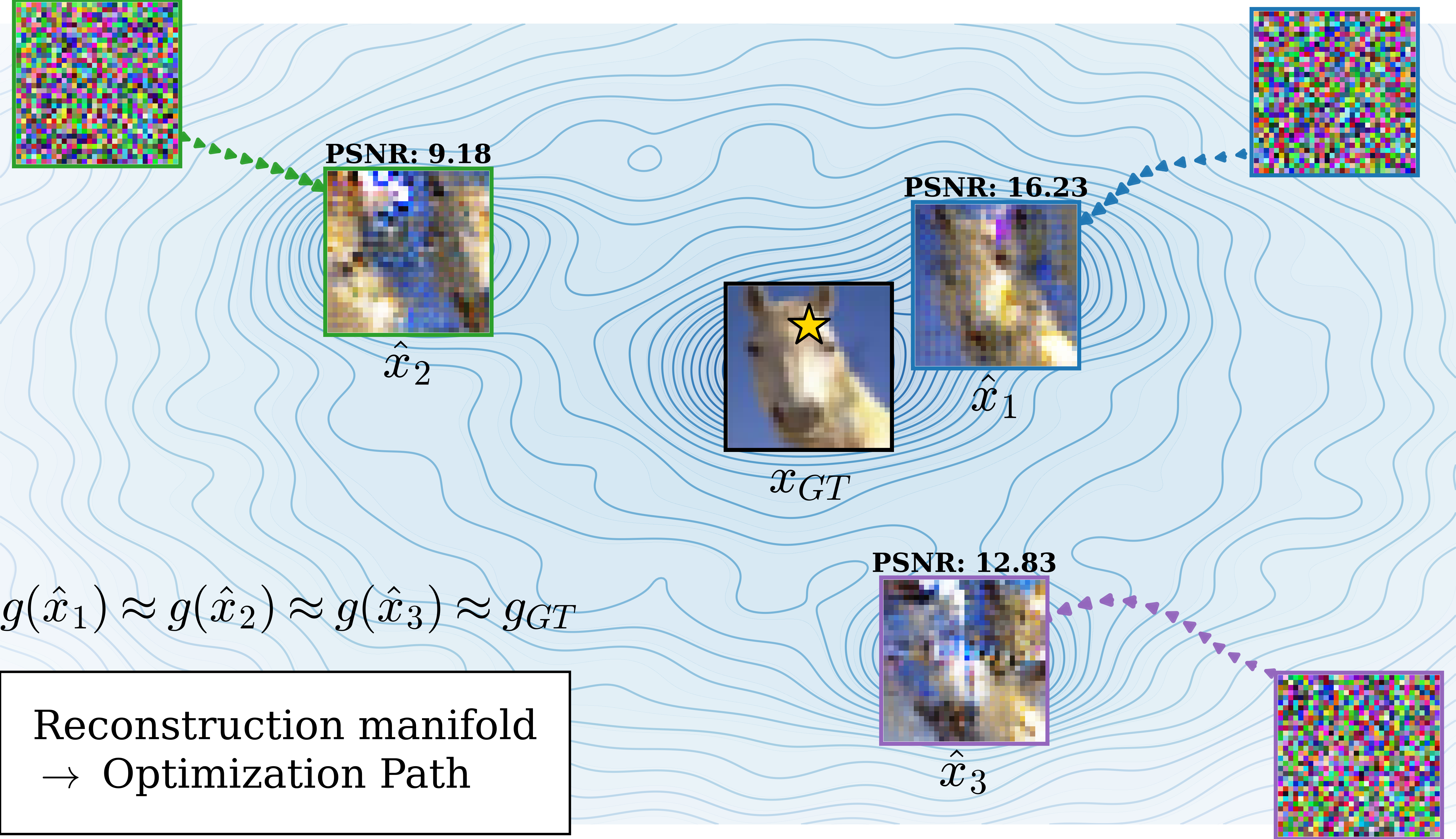}
    \caption{Conceptual illustration of duplicate sample issue in GIAs. Multiple local minima in the gradient-matching objective lead to reconstruction ambiguity, where multiple images yield nearly identical gradients to the target sample. All initializations and reconstructed images are real samples from our experiments.}
    \label{fig:conceptgap}
\end{figure}

Federated Learning (FL) enables privacy-preserving decentralized training by keeping individual users' training data private and by training a global model by sharing users' local gradients \cite{mcmahan17fedavg, zhang2024flhetbench}. However, recent studies have shown that it is possible to breach the user data privacy through several types of data privacy attacks, such as Gradient Inversion Attacks (GIAs) \cite{geiping2020igfl, Zhu020dlg, ali2022gradvit}, which focus on reconstructing the user data, either through an honest-but-curious server interested in recovering user data in the FL system \cite{Zhu020dlg, geiping2020igfl}, or by modifying the model architecture or model weights \cite{fowl2022robbing, fowl2023decepticons, zhao2024loki} and sending the compromised model to the users through a malicious server.

Early works on GIAs targeted fully connected and convolutional networks \cite{Zhu020dlg, geiping2020igfl, ZhuB21rgap, Chaudhry2025egap}, while subsequent work extended these attacks to Vision Transformers (ViTs), revealing the critical role of positional encodings in reconstruction quality \cite{jiahao2022april, ali2022gradvit}. However, these works assumed full fine-tuning.
In contrast, modern federated systems increasingly adopt parameter-efficient fine-tuning (PEFT), particularly LoRA \cite{HuSWALWWC22lora}, which updates only a small low-rank subset of parameters. This weakens the gradient signal, making inversion harder and providing stronger privacy guarantees \cite{zhang2023fedpetuning}. A similar phenomenon is also observed for `Membership Inference Attacks' (MIAs) in Liu et al. \cite{Liu2026nrr}, where decreasing effective rank decreases the attack success rate.

\begin{figure}[!ht]
    \centering
    \includegraphics[width=0.98\linewidth]{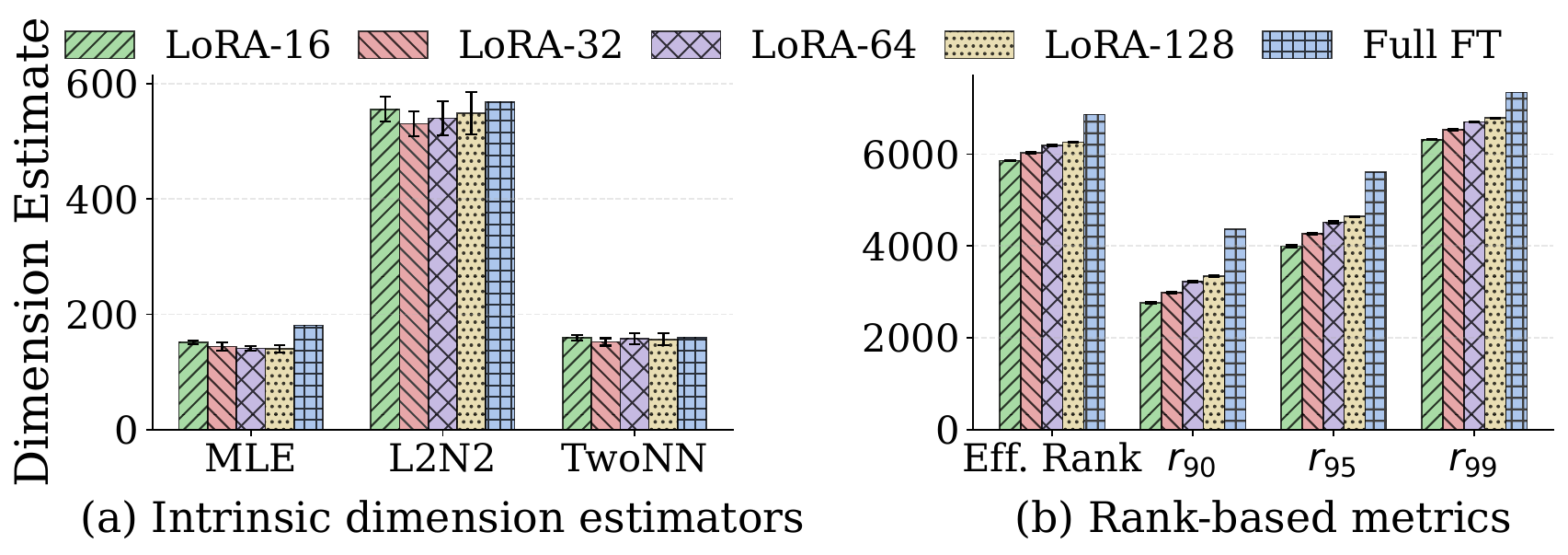}
    \caption{Comparison of intrinsic dimensionality of the gradients obtained from full fine-tuning and the low-rank-adapted base model (ViT-Small), which supports our hypothesis that the dimensionality of the manifold on which the shared gradient lies decreases with reduction in the number of trainable parameters. The metrics are discussed in detail in Supp. Sec. \ref{app:gradient_dimension_estimators}.}
    \label{fig:intdim}
\end{figure}

The feasibility of a gradient inversion attack closely depends on the number of trainable parameters exposed by the client model \cite{geiping2020igfl, ZhuB21rgap}. 
As the number of trainable parameters decreases, the effective manifold on which the shared gradients lie also shrinks \cite{li2018intdim, aghajanyan2021intrinsic}. Consequently, the inversion problem becomes more challenging, since the observed gradient no longer uniquely determines the private input. This results in different initializations of the candidate solutions converging to different optima, as shown in Fig. \ref{fig:conceptgap}. This issue is further exacerbated in PEFT settings, where LoRA-based fine-tuning updates only a small fraction of the base model's total parameters and the intrinsic gradient dimension decreases too, as evident from Fig. \ref{fig:intdim}. Drawing inspiration from gradient inversion attacks on language models like DAGER \cite{petrov2024dager}, LAMP \cite{BalunovicD0V22lamp}, and GRAB \cite{feng2024grab}, which use a token vocabulary to filter the plausible set of tokens, we adopt a generative prior trained on images drawn from a distribution similar to, but non-overlapping with, that of the target data. 
This prior constrains reconstruction to an overlapping image manifold \cite{li2022ggl}, discouraging non-identifiable solutions that match the gradient signal. %

\begin{figure}[!ht]
\centering
\scriptsize
\setlength{\tabcolsep}{1pt}

\begin{tabular}{c|ccc|ccc|ccc}
\toprule
\multirow{2}{*}{\textbf{GT}}
& \multicolumn{3}{c|}{\textbf{APRIL}}
& \multicolumn{3}{c|}{\textbf{GradViT}}
& \multicolumn{3}{c}{\textbf{GIAS}}
\\
& 1 & 2 & 3& 1 & 2 & 3& 1 & 2 & 3 \\
\midrule

\includegraphics[width=0.09\linewidth]{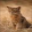}
& \includegraphics[width=0.09\linewidth]{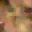}
& \includegraphics[width=0.09\linewidth]{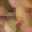}
& \includegraphics[width=0.09\linewidth]{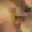}
& \includegraphics[width=0.09\linewidth]{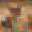}
& \includegraphics[width=0.09\linewidth]{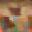}
& \includegraphics[width=0.09\linewidth]{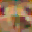}
& \includegraphics[width=0.09\linewidth]{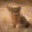}
& \includegraphics[width=0.09\linewidth]{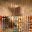}
& \includegraphics[width=0.09\linewidth]{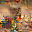}
\\

\includegraphics[width=0.09\linewidth]{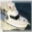}
& \includegraphics[width=0.09\linewidth]{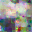}
& \includegraphics[width=0.09\linewidth]{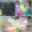}
& \includegraphics[width=0.09\linewidth]{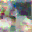}
& \includegraphics[width=0.09\linewidth]{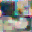}
& \includegraphics[width=0.09\linewidth]{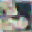}
& \includegraphics[width=0.09\linewidth]{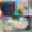}
& \includegraphics[width=0.09\linewidth]{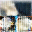}
& \includegraphics[width=0.09\linewidth]{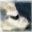}
& \includegraphics[width=0.09\linewidth]{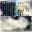}
\\

\includegraphics[width=0.09\linewidth]{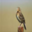}
& \includegraphics[width=0.09\linewidth]{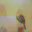}
& \includegraphics[width=0.09\linewidth]{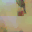}
& \includegraphics[width=0.09\linewidth]{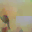}
& \includegraphics[width=0.09\linewidth]{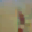}
& \includegraphics[width=0.09\linewidth]{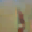}
& \includegraphics[width=0.09\linewidth]{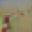}
& \includegraphics[width=0.09\linewidth]{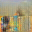}
& \includegraphics[width=0.09\linewidth]{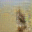}
& \includegraphics[width=0.09\linewidth]{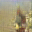}
\\

\bottomrule
\end{tabular}

\caption{Examples of jumbled patches in both optimization and generative GIAs for different initializations on low rank-adapted ($r =128$) ViT-Small on the CIFAR-10 dataset.}
\label{fig:mot_grid}
\end{figure}

In LoRA-based PEFT, the non-identifiability of gradient inversion is also reflected in the occurrence of \emph{duplicate} data samples, where spatially misaligned patches or {distorted and noisy inputs} can induce the same gradients as the true private sample \cite{ZhuB21rgap, jiahao2022april}, as shown in Fig. \ref{fig:mot_grid}. Since positional embeddings encode spatial information, their ambiguity can permit such misaligned, {distorted, or noisy} reconstructions to satisfy the same gradient constraints. From Fig. \ref{fig:mot_grid}, we observe that this non-identifiability is also observed in generative gradient inversion attacks, such as GIAS \cite{gias}, where the use of a learned image prior reduces the search space but does not guarantee a unique reconstruction.

\colorlet{framecolor}{green}

These observations suggest that successful gradient inversion under LoRA-based PEFT requires addressing two interacting sources of ambiguity: (i) reduced information from low-rank parameter updates and (ii) lost positional information from frozen positional embeddings, leading to unstable reconstructions across optimization restarts.

Motivated by these observations, we propose a generative gradient inversion framework that combines LoRA fine-tuning with layer-wise trainable positional embeddings. Although standard ViT architectures do not include layer-wise positional embeddings, adding them enables efficient recovery of positional details without hurting model performance (see Supp. Sec. \ref{sec:backbonecomp}). This design allows the attacker to reconstruct spatial data while respecting the generative prior's realism constraints. Experimental results demonstrate substantial improvements in reconstruction fidelity, attack stability, and resistance to \emph{duplicate} artefacts across low-rank and training configurations. In summary, our main contributions are as follows: 
\begin{itemize}
\setlength{\itemsep}{0.5pt}
    \item We show that LoRA-induced parameter sparsity increases the non-identifiability of GIA by reducing the information content available to the attacker.
    \item We identify the ambiguity of frozen positional embeddings as a fundamental source of \emph{duplicate} reconstructions, revealing that spatially distinct or distorted inputs can satisfy identical gradient constraints under PEFT.
    \item We design a layer-wise trainable positional embedding mechanism for ViTs that exposes positional gradients during federated PEFT, providing spatial supervision that resolves positional ambiguity and consistently improves reconstruction quality.
    \item We extensively evaluate the approach across multiple ranks and training configurations, demonstrating substantial improvements over prior GIA methods.
\end{itemize}

\section{Related Works}

\noindent\textbf{Optimization-based Gradient inversion.}
Early optimization-based GIAs focused on fully connected neural networks (FCNNs) \cite{Rumelhart1986fcnn} and convolutional neural networks (CNNs) \cite{lecun1989cnn}. DLG \cite{Zhu020dlg} showed that inputs and labels can be recovered by optimizing dummy data, while IG \cite{geiping2020igfl} improved reconstruction quality using gradient-matching objectives and image priors. R-GAP \cite{ZhuB21rgap} related reconstruction ambiguity to non-unique pre-activations and loss-gradient responses. GradInversion \cite{yin2021gradinversion} exploits batch-normalization statistics as image priors but struggles with repeated classes in a batch. HFGradInversion \cite{Ye2024stepsiseGIA} addresses this limitation through layer-wise inversion, potentially overlooking cross-layer gradient coupling. SPEAR \cite{dimitar2024spear} instead leverages gradient sparsity in ReLU-based FCNNs using gradient decomposition and thresholding.

\noindent\textbf{Generative Methods for Gradient Leakage.}
GAN-based priors \cite{goodfellow2014gans} are widely used for gradient inversion because they can model realistic image distributions. GGL \cite{li2022ggl} optimizes only the latent code, often missing fine details. GIAS \cite{gias} optimizes the latent code and the generator parameters in two steps, while GIFD \cite{fang2023gifd} progressively optimizes intermediate generator embeddings. %

\noindent\textbf{Gradient inversion in Transformers and Language Models.}
With the adoption of Vision Transformers (ViTs) \cite{dosovitskiy2021vit}, gradient inversion methods began exploiting transformer-specific characteristics. APRIL \cite{jiahao2022april} showed that positional embeddings strongly influence reconstruction quality through patch alignment and demonstrated the \emph{twin-data} phenomenon, where spatially plausible but misarranged reconstructions can arise, while GradViT \cite{ali2022gradvit} further improved inversion fidelity using multiple regularizers. Beyond image classification, TAG \cite{DengWLWSLRD21TAG}, LAMP \cite{BalunovicD0V22lamp}, DAGER \cite{petrov2024dager}, GRAB \cite{feng2024grab}, and FILM \cite{gupta2022film} demonstrated gradient leakage in text and multimodal models.

\noindent\textbf{Gradient leakage under parameter-efficient fine-tuning.}
Unlike full fine-tuning, parameter-efficient fine-tuning (PEFT) shares gradients over only a small subset of parameters, reducing the information available to an attacker. PEFTLeak \cite{Sami2025gipeft} studies leakage from adapter-based PEFT, while MineGRAD \cite{sami2026minegrad} targets LoRA-based PEFT. However, both rely on maliciously modifying model weights, and {DAGER further requires the number of tokens to be no greater than the LoRA rank}. Although these works demonstrate that PEFT does not eliminate privacy risks, their attacks are easier to detect and may interfere with normal federated learning \cite{feng2024privacybckdoor}.

In contrast, gradient inversion for visual data in LoRA-adapted ViTs remains largely unexplored, particularly for pixel-level reconstruction without weight modification. This setting matters for vision encoders in modern VLMs, motivating our study of trainable positional embeddings to resolve patch misalignment.

\section{Methodology}

\subsection{Threat Model}

We consider an \textit{honest-but-curious} server whose goal is to infer private user data from the information exchanged during federated learning (FL) without interfering with the training process \cite{geiping2020igfl, jiahao2022april}. The server is allowed to store and independently analyze updates received from individual clients, following the FL protocol. In particular, the server defines and distributes the global model architecture to all clients before training \cite{BonawitzEGHIIKK19fedlscale, mcmahan17fedavg}. Our threat model allows the server to prescribe the architecture, including trainable positional embeddings. But unlike model poisoning attacks like PEFTLeak \mbox{\cite{Sami2025gipeft}}, it does not modify the clients' local training procedure or introduce undisclosed malicious parameters. Because the server-broadcast model improves task performance, it also builds trust with clients. %
Since model quality (See Sec. \ref{sec:backbonecomp} of the supplementary material) and robustness are key considerations in deploying trustworthy FL systems \mbox{\cite{Tariq2024tflsurvey, Liu2022carefl, SnchezSnchez2024fedtrust}}, such architectures are likely to be adopted in practice. %
Thus, our proposed architecture represents a practically relevant design rather than an artificial modification introduced solely to facilitate an attack. {We show in Sec. \ref{sec:expres} that these additional positional embeddings with insignificant computational overhead lead to ease in retrieval of the client data without sacrificing performance.}

\subsection{Problem Formulation}

In this section, we formally describe the gradient inversion (GI) problem. Consider a standard supervised classification task that optimizes a neural network $f_\theta$ parameterized by $\theta$:
\begin{equation}
\min_{\theta} \sum_{(x,y)\in\mathcal{D}} \ell\!\left(f_\theta(x), y\right),
\label{eq:training_objective}
\end{equation}
where $\ell$ denotes a point-wise loss function and $\mathcal{D}$ is a dataset consisting of inputs $x\in\mathbb{R}^m$ and labels $y\in\{0,1\}^L$ represented as one-hot vectors.

In the federated learning setting, clients communicate gradients of the loss $\ell(f_\theta(x),y)$ rather than transmitting raw training data. The goal of gradient inversion is to reconstruct the training samples that generated the observed gradients. Suppose a client computes a gradient,
\begin{equation}
g=\frac{1}{B}\sum_{j=1}^{B}
\nabla_{\theta} \ell\!\left(f_\theta(x_j^*),y_j^*\right)
\end{equation}
from a mini-batch $\{(x_1^*,y_1^*),\ldots,(x_B^*,y_B^*)\}$. The gradient inversion problem can then be formulated as
\begin{equation}
\min_{(x_1,y_1),\ldots,(x_B,y_B)\in\mathbb{R}^m\times\{0,1\}^L} d\!\left(\frac{1}{B} \sum_{j=1}^{B} \nabla_{\theta} \ell\!\left(f_\theta(x_j),y_j\right),g\right)
\label{eq:gradient_matching}
\end{equation}
where $d(\cdot,\cdot)$ measures the discrepancy between two gradients, such as the $\ell_2$ distance or negative cosine similarity.

Following prior work \cite{zhao2020idlg, yin2021gradinversion} that shows that labels $y$ can often be recovered directly from the gradients of the final layer, we focus on recovering the inputs assuming the labels are available. This leads to the optimization problem
\begin{equation}
\min_{x_1,\ldots,x_B\in\mathbb{R}^m} c(x_1,\ldots,x_B;\theta,g),
\label{eq:input_reconstruction}
\end{equation}
where $c(x_1,\ldots,x_B;\theta,g)$ denotes the objective in Equation~\eqref{eq:gradient_matching} with $y_j=y_j^*$ for all $j=1,\ldots,B$.

\noindent\textbf{Generative GIA.} Given a pre-trained generative model $G_\omega:\mathbb{R}^k \rightarrow \mathbb{R}^m$ trained on a public dataset, a natural approach is to use the generator as an image prior during gradient inversion. Instead of directly optimizing over the input space $\mathbb{R}^m$, we constrain the reconstruction to the image manifold induced by the generator, i.e., $\{G_\omega(z): z\in\mathbb{R}^k\}$. Since the latent dimension $k$ is typically much smaller than the image dimension $m$, searching in the latent space significantly reduces the solution space and alleviates the under-determined nature of the inversion problem. Consequently, the inversion objective can be reformulated as an optimization over the latent code,
\begin{equation}
    z^*=\arg\min_z d\!\left(\nabla_{\theta}\ell(f_\theta(G_\omega(z)),\Tilde{y}),g\right)+R_{\text{prior}}(z;G_\omega),
\end{equation}
where $d(\cdot,\cdot)$ measures the discrepancy between generated and observed gradients, $\Tilde{y}$ is the label inferred by the server and $R_{\text{prior}}$ enforces consistency with the generative model.

\begin{figure}[!ht]
    \centering
    \includegraphics[width=0.9\linewidth]{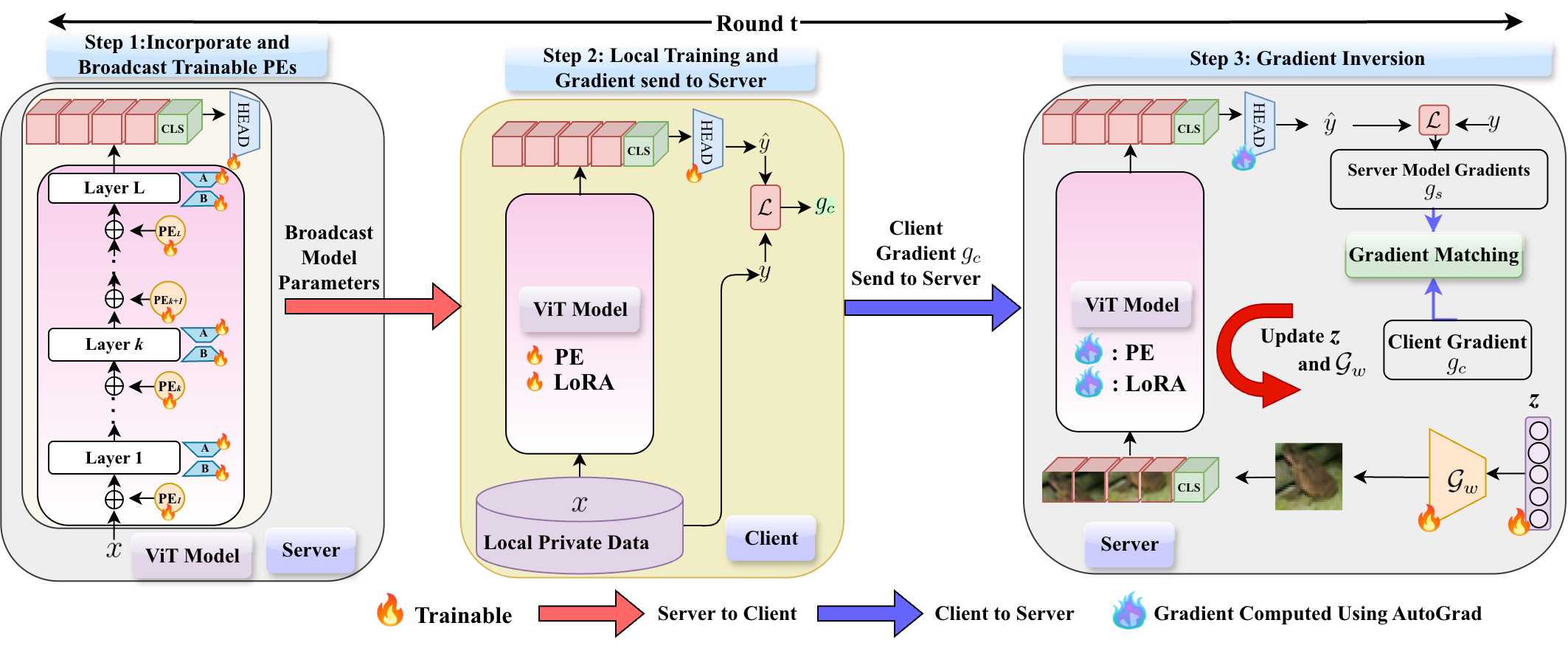}
    \caption{A high-level overview of our method showing how the baseline generative GIA framework works. $g^C$ and $g^S$ denote gradients from the client and server models.}
    \label{fig:highlvlimg}
\end{figure}

In this work, we adopt GIAS \cite{gias}, one of the earliest generative GIA frameworks, as our primary baseline. {Fig. \mbox{\ref{fig:highlvlimg}} provides a high-level overview of the method used here.}

\subsection{Generative vs. Optimization-Based Gradient Inversion under LoRA}

\noindent
Recovering an input sample $x \in \mathbb{R}^{m}$ from its gradient $\nabla_{\theta}\ell(f_{\theta}(x),y)\in\mathbb{R}^{p}$ constitutes an inverse problem whose difficulty depends on the amount of information contained in the observed gradients. When $p<m$, the problem is under-determined due to incomplete observations \cite{BenningB18mrminvprobs, CandesRT06RUP}. Even when $p>m$, reconstruction remains challenging because the gradient mapping is highly non-linear and often ill-conditioned \cite{geiping2020igfl, ZhuB21rgap, jiahao2022april}. To mitigate these issues, gradient inversion methods incorporate regularization and priors \cite{yin2021gradinversion, ali2022gradvit, sanjay2023cpa} to restrict the search space and encourage realistic solutions.

Optimization-based GIAs start from a randomly initialized dummy input $\hat{x}_0$, typically far from the target image $x^*$ in the image space. As the number of observed parameters decreases, the inverse problem becomes increasingly underdetermined because less gradient information is available for reconstruction \cite{BenningB18mrminvprobs, CandesRT06RUP}. In LoRA-adapted fine-tuning, only a small subset of gradients is available, substantially increasing reconstruction ambiguity.

Generative GIAs address this challenge by introducing a strong image prior through a pretrained generator \cite{li2022ggl}. %
Consequently, the optimization starts from a plausible image estimate rather than an arbitrary random initialization. Furthermore, optimizing the latent code $z$ allows the reconstruction to move within this manifold toward regions more consistent with the observed gradients \cite{gias}. By restricting the solution space to realistic images and providing a favourable initialization, generative priors reduce the likelihood of convergence to poor local minima.

\subsection{Spatial Anchoring through Layer-wise Trainable Positional Embedding}

While generative priors substantially reduce the inversion search space, they do not fully resolve the ambiguities introduced by LoRA fine-tuning, as shown in Fig.~\ref{fig:mot_grid}. Self-attention contextualizes token representations through interactions among all tokens \cite{abnar2020,ferrando2022measuring}, causing the shared LoRA gradients to accumulate contributions from multiple contextualized patch representations rather than corresponding to any single spatial location. We hypothesize that this aggregation reduces the explicit spatial cues available for gradient inversion. Furthermore, ViTs rely on positional embeddings to encode patch order, while adding positional information only at the input limits expressiveness, motivating the layer-wise propagation of positional embeddings \cite{lape, LeeHK25mpvg}.

{To alleviate this issue, we modify the target ViT by adding trainable positional embeddings to the input of each transformer encoder layer, exposing richer spatial cues through its gradients. This design enhances positional expressiveness and improves model performance and robustness, with empirical evidence supporting the suitability of the proposed architecture provided in supplementary Sec. \mbox{\ref{sec:backbonecomp}}. We further investigate whether it also exposes richer spatial information through the shared gradients, enabling more accurate gradient inversion attacks.}

\begin{figure}[!ht]
    \centering
    \includegraphics[width=0.9\linewidth]{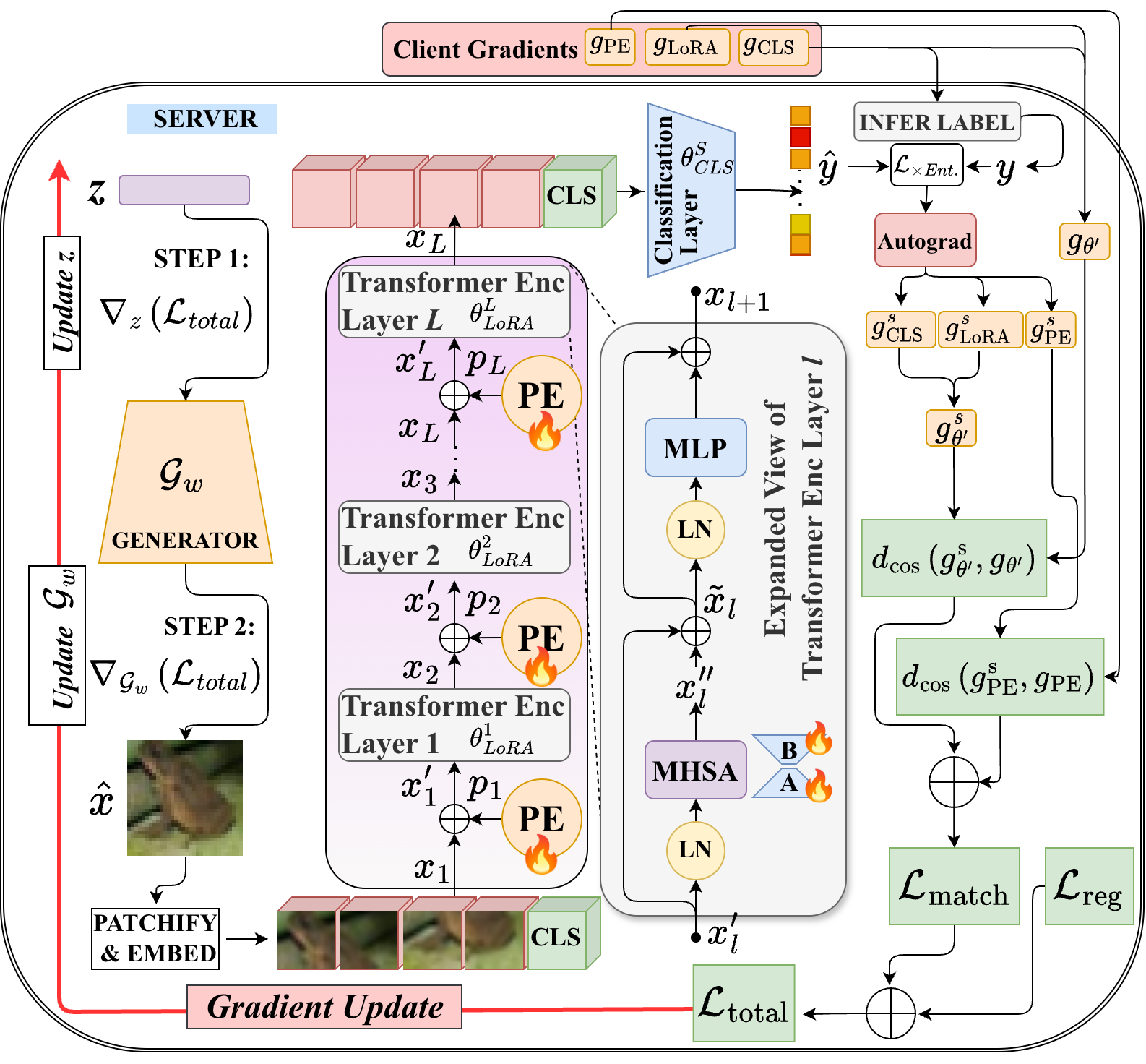}
    \caption{Schematic of the proposed GIA. The overall optimization proceeds in two stages, from client gradients to image reconstruction and subsequent updates of the latent code and generator parameters. The gradients of the trainable positional embeddings enable improved reconstruction quality.
    $g^s_{\cdot} = \nabla_{\cdot}\ell(\hat{y},y)$}
    \label{fig:archifigure}
\end{figure}

Specifically, we augment the token representation in each Transformer block except the first with additional positional information. Given the input tokens $x_l$ to layer $l$, the modified encoder computes
\begin{equation}
x_{l+1} = \tilde{x}_l + \operatorname{MLP}\Biggl( \operatorname{LN}'\Bigl( \smash{\underbrace{\operatorname{MHSA}\bigl( \operatorname{LN}(\overbrace{x_l+p_l}^{x'}) \bigr) + x'}_{\tilde{x}_l}} \Bigr) \Biggr)%
\label{eqn:allposembed}
\end{equation}

where MLP, LN, MHSA, $p_l$ denote \textbf{M}ulti-\textbf{L}ayer \textbf{P}erceptron, \textbf{L}ayer \textbf{N}ormalization, \textbf{M}ulti-\textbf{H}ead \textbf{S}elf-\textbf{A}ttention layers and the positional embedding at layer $l$ as shown in Fig. \ref{fig:archifigure}. %
This modification is intentionally introduced by the server at the time of initialization of local models before FL fine-tuning starts. Consequently, the gradients shared by the client now contain explicit spatial information at each layer and further constrain the space of possible inputs $x$. This substantially reduces patch permutation artefacts and leads to more faithful reconstructions, particularly in settings where LoRA gradients alone are insufficient to uniquely determine the spatial structure of the image. {Furthermore, adding trainable positional embedding to ViT-Small results in a minor increase of 0.016\% in GFLOPS.}

\subsection{Training Objectives and Regularization}

During reconstruction, the attacker optimizes a candidate image produced by a generator, $\hat{x}=G_\omega(z)$, so that the gradients induced by $\hat{x}$ match the gradients shared by the client. {Let $g_{\mathrm{CLS}}, g_{\mathrm{LoRA}}, g_{\mathrm{PE}}$ denote the observed gradients of the classification layer, LoRA and positional embedding parameters, respectively, corresponding to $\theta_{\mathrm{CLS}}, \theta_{\mathrm{LoRA}}, \theta_{\mathrm{PE}}$. Corresponding parameters and their gradients on the server are denoted by $\theta^S_{\mathrm{CLS}}, \theta^S_{\mathrm{LoRA}}, \theta^S_{\mathrm{PE}}$, and $g^S_{\mathrm{CLS}}, g^S_{\mathrm{LoRA}}, g^S_{\mathrm{PE}}$}. For image-label pair $(\hat{x},y)$, we minimize the objective
\begin{equation}
\mathcal{L}_{\mathrm{match}}
= d_{\cos}\!\left(\nabla_{\theta'^S}\ell(\hat{y},y), g_{\theta'}\right) + d_{\cos}\!\left(\nabla_{\theta^S_{\mathrm{PE}}}\ell(\hat{y},y), g_{\mathrm{PE}}\right),%
\label{eqn:lmatch}
\end{equation}

\noindent
where $d_{\cos}(\cdot,\cdot)$ is the negative cosine similarity between two gradient vectors, $\hat{y} = f_\theta(\hat{x})$, $\theta' = \{\theta_{\mathrm{CLS}}, \theta_{\mathrm{LoRA}}\}$ and $\theta'^{S} = \{\theta^S_{\mathrm{CLS}}, \theta^S_{\mathrm{LoRA}}\}$. The final objective is
\begin{equation}
    \min_{z,G_\omega} \; \mathcal{L}_{\mathrm{match}} + \lambda_{\mathrm{reg}} R_{\mathrm{base}}(z,G_\omega),
\end{equation}
where $R_{\mathrm{base}}$ and $\lambda_{reg}$ denote the regularization terms used by the corresponding generative baseline and the regularization coefficient, respectively. For a fair comparison, we use the same regularization choices as the baseline method GIAS~\cite{gias}. In addition to the standard regularizers from GIAS, we employ the group lazy regularization framework of GradViT \cite{ali2022gradvit} for evaluations conducted both with and without our proposed modification (refer to Supp. Sec. \ref{sec:glr} for further discussion). Thus, performance differences reflect the effect of exposing positional information under LoRA-based PEFT.

\section{Experiments and Results}
\label{sec:expres}
\subsection{Experimental Details}
\noindent
To evaluate the effectiveness of our proposed approach, we conduct gradient inversion experiments on the CIFAR-10, CelebA, and CARS datasets. Following prior work, all images from the CIFAR, CelebA and CARS datasets are kept at resolutions $32 \times 32$, $128 \times 128$ and $64 \times 64$, respectively.

For the generator, we use a pre-trained DCGAN generator following the strategy used in GIAS \cite{gias}. Since our objective is to investigate the effectiveness of generative GIAs on LoRA-adapted ViT, the client model is initialized with an ImageNet-pretrained ViT-Small. Unless otherwise specified, we consider a single local optimization step with batch size (B=1). {The details for other datasets are provided in the Appendix Sec. \mbox{\ref{sec:expxtra}} in the supplementary}. We additionally evaluate the attacks under larger batch sizes for single-step FedSGD and small-batch multi-step FedAvg settings.

\noindent\textbf{Evaluation Metrics:} We assess reconstruction quality using three standard image similarity metrics: Peak Signal-to-Noise Ratio (PSNR), Structural Similarity Index Measure (SSIM), and Learned Perceptual Image Patch Similarity (LPIPS). Higher PSNR and SSIM values indicate better reconstruction fidelity, whereas lower LPIPS values correspond to smaller discrepancies between reconstructed images and the ground-truth private samples. 

\subsection{Experimental Results}
\subsubsection{Attack on Single Step Update Gradient}
\label{subsubsec:fedsgdattack}

We first evaluate single-step FedSGD ($B=1$), isolating per-sample leakage for controlled comparison of inversion methods. Results are averaged over 10 users, with three independent reconstructions per sample for multiple trials; standard deviations are reported in parentheses.

\begin{table}[!ht]
\centering
\caption{Performance on CIFAR-10 for Batch Size (BS) = 1. $\Delta$ indicates the change brought by (+Ours). In the case of LPIPS (L), a negative difference means improvement, while a positive difference indicates improvement for PSNR (P) and SSIM (S).}
\label{tab:cifar_bs1_differences}
\scriptsize
\setlength{\tabcolsep}{1pt}

\newcommand{\posdiff}[1]{\textcolor{green!60!black}{\textbf{#1}}}
\newcommand{\negdiff}[1]{\textcolor{red!80!black}{\textbf{#1}}}

\begin{tabular}{l|ccc|ccc|ccc}
\toprule
\textbf{Met.} & \textbf{APRIL} & \textbf{+Ours} & $\mathbf{\Delta}$ & \textbf{GradViT} & \textbf{+Ours} & $\mathbf{\Delta}$ & \textbf{GIAS} & \textbf{+Ours} & $\mathbf{\Delta}$ \\\midrule
\textbf{P} $\uparrow$ & 
\makecell{12.75 \\ \tiny{(2.19)}} & \makecell{12.03 \\ \tiny{(2.39)}} & \negdiff{-0.72} & 
\makecell{15.03 \\ \tiny{(4.68)}} & \makecell{16.62 \\ \tiny{(4.5)}} & \posdiff{+1.59} & 
\makecell{15.77 \\ \tiny{(4.52)}} & \makecell{25.91 \\ \tiny{(6.29)}} & \posdiff{+10.14} \\ \midrule
\textbf{S} $\uparrow$ & 
\makecell{0.40 \\ \tiny{(0.152)}} & \makecell{0.45 \\ \tiny{(0.11)}} & \posdiff{+0.05} & 
\makecell{0.397 \\ \tiny{(0.226)}} & \makecell{0.48 \\ \tiny{(0.266)}} & \posdiff{+0.083} & 
\makecell{0.34 \\ \tiny{(0.221)}} & \makecell{0.79 \\ \tiny{(0.17)}} & \posdiff{+0.45} \\ \midrule
\textbf{L} $\downarrow$ & 
\makecell{0.24 \\ \tiny{(0.077)}} & \makecell{0.19 \\ \tiny{(0.224)}} & \posdiff{-0.05} & 
\makecell{0.124 \\ \tiny{(0.05)}} & \makecell{0.12 \\ \tiny{(0.051)}} & \posdiff{-0.004} & 
\makecell{0.12 \\ \tiny{(0.075)}} & \makecell{0.022 \\ \tiny{(0.033)}} & \posdiff{-0.098} \\ 
\bottomrule
\end{tabular}
\end{table}

\begin{figure}[!ht]
\centering
\scriptsize
\setlength{\tabcolsep}{1pt}
\begin{tabular}{c| cccc|cccc}
\toprule
\textbf{Method} &
\multicolumn{4}{c|}{4 samples from a batch of 4}& \multicolumn{4}{c}{4 samples from a batch of 16} \\ 
\midrule
GT
& \includegraphics[width=0.09\linewidth]{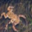}
& \includegraphics[width=0.09\linewidth]{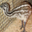}
& \includegraphics[width=0.09\linewidth]{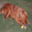}
& \includegraphics[width=0.09\linewidth]{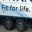}
& \includegraphics[width=0.09\linewidth]{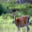}
& \includegraphics[width=0.09\linewidth]{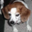}
& \includegraphics[width=0.09\linewidth]{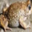}
& \includegraphics[width=0.09\linewidth]{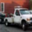}
\\

GIAS
& \includegraphics[width=0.09\linewidth]{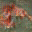}
& \includegraphics[width=0.09\linewidth]{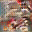}
& \includegraphics[width=0.09\linewidth]{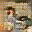}
& \includegraphics[width=0.09\linewidth]{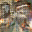}
& \includegraphics[width=0.09\linewidth]{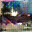}
& \includegraphics[width=0.09\linewidth]{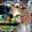}
& \includegraphics[width=0.09\linewidth]{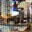}
& \includegraphics[width=0.09\linewidth]{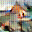}
\\

GIAS + Ours
& \includegraphics[width=0.09\linewidth]{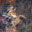}
& \includegraphics[width=0.09\linewidth]{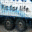}
& \includegraphics[width=0.09\linewidth]{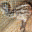}
& \includegraphics[width=0.09\linewidth]{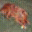}
& \includegraphics[width=0.09\linewidth]{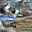}
& \includegraphics[width=0.09\linewidth]{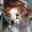}
& \includegraphics[width=0.09\linewidth]{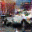}
& \includegraphics[width=0.09\linewidth]{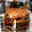}
\\
\bottomrule
\end{tabular}

\caption{Qualitative comparison for GIAS and GIAS+Ours for batch sizes 4 and 16 in the single-step setting with rank 128.}
\label{fig:largebatch}
\end{figure}

\noindent
Tab. \ref{tab:cifar_bs1_differences} shows that Low rank-adapted ViTs remain vulnerable even when the attacker observes only low-rank gradients. The proposed positional-gradient anchoring provides the largest gains for the generative baseline: GIAS+Ours improves PSNR from 15.77 to 25.91 and SSIM from 0.34 to 0.79, while reducing LPIPS from 0.12 to 0.022. GradViT also benefits from the additional spatial signal, while APRIL shows improved SSIM and LPIPS. These results suggest that positional-embedding gradients are most effective when combined with a generative prior. Qualitative examples in Fig. \ref{fig:c10_qual_grid} further demonstrate reduced patch-permutation and duplicate-data artifacts. In supplementary Sec. \ref{sec:tpeoptgia}, we present qualitative results on optimization-based GIAs too. We also present a few visual results from a single-step gradient with larger batch sizes in Fig. \ref{fig:largebatch}, which show that even for larger batch sizes, the proposed strategy helps in recovering high-quality reconstructions (More samples in Supplementary Sec. \ref{sec:largebsmore}). Quantitative results for larger batch sizes are reported in Sec. \ref{subsubsec:ablbs}.

\begin{figure}[!ht]
\centering
\scriptsize
\setlength{\tabcolsep}{1pt}

\begin{tabular}{c cccccccc}
\toprule
\textbf{Method}
& \textbf{1}
& \textbf{2}
& \textbf{3}
& \textbf{4}
& \textbf{5}
& \textbf{6}
& \textbf{7}
& \textbf{8}\\
\midrule

GT
& \includegraphics[width=0.09\linewidth]{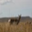}
& \includegraphics[width=0.09\linewidth]{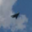}
& \includegraphics[width=0.09\linewidth]{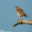}
& \includegraphics[width=0.09\linewidth]{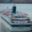}
& \includegraphics[width=0.09\linewidth]{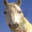}
& \includegraphics[width=0.09\linewidth]{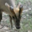}
& \includegraphics[width=0.09\linewidth]{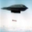}
& \includegraphics[width=0.09\linewidth]{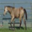}
\\

GIAS
& \includegraphics[width=0.09\linewidth]{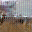}
& \includegraphics[width=0.09\linewidth]{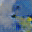}
& \includegraphics[width=0.09\linewidth]{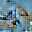}
& \includegraphics[width=0.09\linewidth]{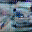}
& \includegraphics[width=0.09\linewidth]{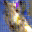}
& \includegraphics[width=0.09\linewidth]{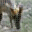}
& \includegraphics[width=0.09\linewidth]{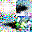}
& \includegraphics[width=0.09\linewidth]{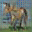}
\\

GIAS + Ours
& \includegraphics[width=0.09\linewidth]{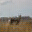}
& \includegraphics[width=0.09\linewidth]{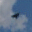}
& \includegraphics[width=0.09\linewidth]{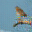}
& \includegraphics[width=0.09\linewidth]{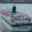}
& \includegraphics[width=0.09\linewidth]{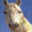}
& \includegraphics[width=0.09\linewidth]{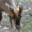}
& \includegraphics[width=0.09\linewidth]{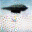}
& \includegraphics[width=0.09\linewidth]{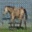}
\\
\bottomrule
\end{tabular}
\caption{Qualitative comparison on the CIFAR-10 dataset. We can observe that the spatial information obscuration issue is solved after applying the proposed modification in the ViT-Small model.}
\label{fig:c10_qual_grid}
\end{figure}

\begin{figure}[!ht]
\centering
\scriptsize
\setlength{\tabcolsep}{1pt}

\begin{tabular}{c cccc|cccc}
\toprule
\textbf{Method}
& \multicolumn{4}{c|}{CelebA ($128\times128$)}
& \multicolumn{4}{c}{CARS ($64\times64$)}\\
\midrule

GT
& \includegraphics[width=0.09\linewidth]{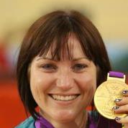}
& \includegraphics[width=0.09\linewidth]{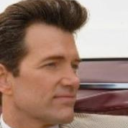}
& \includegraphics[width=0.09\linewidth]{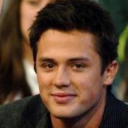}
& \includegraphics[width=0.09\linewidth]{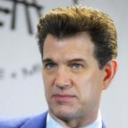}
& \includegraphics[width=0.09\linewidth]{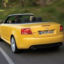}
& \includegraphics[width=0.09\linewidth]{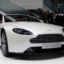}
& \includegraphics[width=0.09\linewidth]{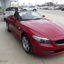}
& \includegraphics[width=0.09\linewidth]{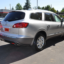}
\\

GIAS
& \includegraphics[width=0.09\linewidth]{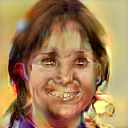}
& \includegraphics[width=0.09\linewidth]{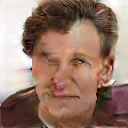}
& \includegraphics[width=0.09\linewidth]{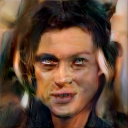}
& \includegraphics[width=0.09\linewidth]{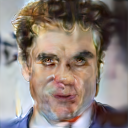}
& \includegraphics[width=0.09\linewidth]{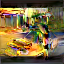}
& \includegraphics[width=0.09\linewidth]{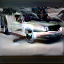}
& \includegraphics[width=0.09\linewidth]{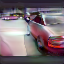}
& \includegraphics[width=0.09\linewidth]{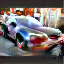}
\\

GIAS+Ours
& \includegraphics[width=0.09\linewidth]{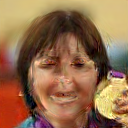}
& \includegraphics[width=0.09\linewidth]{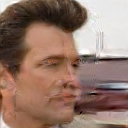}
& \includegraphics[width=0.09\linewidth]{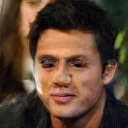}
& \includegraphics[width=0.09\linewidth]{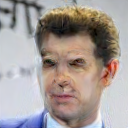}
& \includegraphics[width=0.09\linewidth]{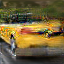}
& \includegraphics[width=0.09\linewidth]{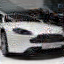}
& \includegraphics[width=0.09\linewidth]{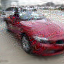}
& \includegraphics[width=0.09\linewidth]{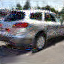}
\\

\bottomrule
\end{tabular}

\caption{{Qualitative comparison on CelebA and CARS datasets. We observe significant improvements in reconstruction quality with trainable positional embeddings applied to GIAS, resolving distorted patches in high-resolution images. Notably, instance-level identities of humans and cars are accurately retrieved, unlike competing techniques, highlighting the attack's potential impact with minimal computational overhead.}
}
\label{fig:celeba_qual_grid}
\end{figure}

{We further present additional qualitative results on the CelebA and Stanford CARS datasets in the single-step gradient update setup in Fig. \mbox{\ref{fig:celeba_qual_grid}}, where we observe that the proposed method is capable of recovering the client data with better clarity than the baseline method. Furthermore, from Tab. \mbox{\ref{tab:celebacars}}, we observe that the proposed method achieves higher scores in terms of PSNR, SSIM, and lower LPIPS on both datasets. The performance on the CARS dataset also reflects the effect of the distribution gap between the generator and the client data. More empirical evidence is presented in Secs. \mbox{\ref{sec:moreceleba}} and \mbox{\ref{sec:distdist}} of the supplementary material.}

\begin{table}[!ht]
    \centering
    \scriptsize
    \caption{Quantitative comparison on CelebA and CARS datasets. Evidently, we can observe the significant jump in performance when trainable positional embedding is applied to GIAS.}
    \label{tab:celebacars}
    \begin{tabular}{l|c|c|c|c|c|c}
    \toprule
     Datasets & \multicolumn{3}{c|}{\textbf{CelebA}} & \multicolumn{3}{c}{\textbf{CARS}} \\ \midrule
    Method & PSNR \tiny{$\uparrow$} & SSIM\tiny{$\uparrow$} & LPIPS\tiny{$\downarrow$} & PSNR\tiny{$\uparrow$} & SSIM\tiny{$\uparrow$} & LPIPS\tiny{$\downarrow$} \\\midrule
     GIAS   
     & \makecell{11.71 \\ \tiny{(2.8)}} 
     &\makecell{0.24 \\ \tiny{(0.10)}} 
     &\makecell{0.31\\ \tiny {(0.08)}} 
     &\makecell{9.04 \\ \tiny{(1.90)}} 
     &\makecell{0.07 \\ \tiny{(0.08)}} 
     &\makecell{0.36 \\ \tiny{(0.07)}}\\
     \makecell{GIAS \\ +Ours}
     & \makecell{\textbf{19.54} \\\tiny{(3.72)}} &\makecell{\textbf{0.585} \\ \tiny{(0.125)}} &\makecell{\textbf{0.125} \\ \tiny{(0.082)}} &\makecell{\textbf{20.783} \\ \tiny{(3.166)}} &\makecell{\textbf{0.72} \\ \tiny{(0.128)}} &\makecell{\textbf{0.089} \\ \tiny{(0.058)}}\\ \bottomrule
    \end{tabular}%
\end{table}

\subsubsection{Attack on Multi-Step Update Gradient}

We next evaluate the multi-step setting, where the server observes the accumulated client update after multiple local optimization steps instead of a single gradient update. Since LoRA-based PEFT updates only a small subset of parameters, each client is assumed to hold 16 private samples per communication round to provide sufficient reconstruction information. We vary the local batch size ($B\in\{4,8\}$) and local epochs ($E\in\{1,2\}$), resulting in $E\times16/B$ local update steps before communication. Compared with the single-step setting, the accumulated updates contain information from multiple samples and optimization steps, making gradient inversion more challenging.

Tab. \ref{tab:steps_results} summarizes the results. Adding trainable positional embeddings provides the largest improvement for GIAS ($\Delta_{PSNR}>2.0 \texttt{dB}$), and the gain is statistically significant (Welch's t-test, $p<0.05$), whereas the differences for APRIL and GradViT are not ($p>0.05$; Supp. Sec. \ref{sec:statguarfedavg}). Unlike the optimization-based attacks, GIAS combined with the proposed spatial anchoring continues to improve with more local updates, suggesting that the additional spatial cues become increasingly beneficial over longer optimization trajectories.

GradViT shows a slight, statistically insignificant performance decrease in the multi-step (FedAvg) setting, converging to a marginally higher gradient-matching loss. Overall, the proposed architecture consistently benefits generative gradient inversion while maintaining comparable performance for optimization-based methods.

\begin{table}[!ht]
\centering
\caption{Evaluation of performance on CIFAR-10 for multiple local updates. Parentheses denote standard deviation. $\Delta$ tracks difference in performance. Metrics (Met.) P, S, and L refer to PSNR, SSIM, and LPIPS, respectively.}
\label{tab:steps_results}
\scriptsize
\setlength{\tabcolsep}{1pt}

\newcommand{\posdiff}[1]{\textcolor{green!60!black}{\textbf{#1}}}
\newcommand{\negdiff}[1]{\textcolor{red!80!black}{\textbf{#1}}}

\resizebox{\linewidth}{!}{
\begin{tabular}{ll|c|ccc|ccc|ccc}
\toprule
 & & & \multicolumn{3}{c|}{\textbf{APRIL}} & \multicolumn{3}{c|}{\textbf{GradViT}} & \multicolumn{3}{c}{\textbf{GIAS}} \\
\textbf{E} & \textbf{B} & \textbf{Met.} & \textbf{w/o Ours} & \textbf{+Ours} & $\mathbf{\Delta}$ & \textbf{w/o Ours} & \textbf{+Ours} & $\mathbf{\Delta}$ & \textbf{w/o Ours} & \textbf{+Ours} & $\mathbf{\Delta}$ \\
\midrule

& & \textbf{P} $\uparrow$ & \makecell{13.41 \\ \tiny{(0.585)}} & \makecell{13.56 \\ \tiny{(0.44)}} & \posdiff{+0.15} & \makecell{14.01 \\ \tiny{(0.290)}} & \makecell{13.91 \\ \tiny{(0.517)}} &\negdiff{-0.1} & \makecell{11.04 \\ \tiny{(0.648)}} & \makecell{13.12 \\ \tiny{(1.58)}} & \posdiff{+2.08} \\

\textbf{1} & \textbf{8} & \textbf{S} $\uparrow$ & \makecell{0.31 \\ \tiny{(0.028)}} & \makecell{0.333 \\ \tiny{(0.02)}} & \posdiff{+0.023} & \makecell{0.243 \\ \tiny{(0.033)}} & \makecell{0.243 \\ \tiny{(0.037)}} & \posdiff{0.0} & \makecell{0.075 \\ \tiny{(0.0173)}} & \makecell{0.22 \\ \tiny{(0.105)}} & \posdiff{+0.15} \\

& & \textbf{L} $\downarrow$ & \makecell{0.265 \\ \tiny{(0.013)}} & \makecell{0.263 \\ \tiny{(0.014)}} & \posdiff{-2e-3} &  \makecell{0.199 \\ \tiny{(0.018)}}& \makecell{0.199 \\ \tiny{(0.022)}} & \posdiff{0.0} & \makecell{0.133 \\ \tiny{(0.014)}} & \makecell{0.106 \\ \tiny{(0.024)}} & \posdiff{-0.027} \\
\midrule

& & \textbf{P} $\uparrow$ & \makecell{13.34 \\ \tiny{(0.583)}} &\makecell{13.79 \\ \tiny{(0.352)}}  & \posdiff{+0.45} & \makecell{14.09 \\ \tiny{(0.948)}} & \makecell{13.86 \\ \tiny{(0.527)}} & \negdiff{-0.23} & \makecell{10.82 \\ \tiny{(0.557)}} & \makecell{13.21 \\ \tiny{(1.59)}} & \posdiff{+2.39}\\

\textbf{1} & \textbf{4} & \textbf{S} $\uparrow$ &  \makecell{0.30 \\ \tiny{(2.5e-2)}} &\makecell{0.33 \\ \tiny{(2.2e-2)}} & \posdiff{+0.03} & \makecell{0.257 \\ \tiny{(0.058)}} & \makecell{0.244 \\ \tiny{(0.021)}} & \negdiff{-0.013} &  \makecell{0.076 \\ \tiny{(0.014)}} & \makecell{0.225 \\ \tiny{(0.095)}} & \posdiff{+0.15} \\

& & \textbf{L} $\downarrow$ & \makecell{0.274 \\ \tiny{(0.014)}} &\makecell{0.267 \\ \tiny{(0.012)}} & \posdiff{-7e-3} &  \makecell{0.195 \\ \tiny{(0.005)}} & \makecell{0.195 \\ \tiny{(0.014)}} & \posdiff{0.0} & \makecell{0.131 \\ \tiny{(0.014)}} & \makecell{0.102 \\ \tiny{(0.02)}} & \posdiff{-0.029} \\
\midrule

& & \textbf{P} $\uparrow$ & \makecell{13.44 \\ \tiny{(0.681)}} & \makecell{13.66 \\ \tiny{(0.219)}} & \posdiff{+0.22} & \makecell{14.01 \\ \tiny{(0.118)}} & \makecell{13.85 \\ \tiny{(0.53)}} & \negdiff{-0.16} &  \makecell{10.84 \\ \tiny{(0.529)}} & \makecell{13.23 \\ \tiny{(1.637)}} & \posdiff{+2.39} \\

\textbf{2} & \textbf{8} & \textbf{S} $\uparrow$ & \makecell{0.313 \\ \tiny{(0.031)}} & \makecell{0.338 \\ \tiny{(0.03)}} & \posdiff{+0.025} & \makecell{0.267 \\ \tiny{(0.007)}} & \makecell{0.24 \\ \tiny{(0.041)}} & \negdiff{-0.027} & \makecell{0.074 \\ \tiny{(0.014)}} & \makecell{0.225 \\ \tiny{(0.095)}} & \posdiff{+0.15}\\

& & \textbf{L} $\downarrow$ & \makecell{0.267 \\ \tiny{(0.02)}} & \makecell{0.263 \\ \tiny{(0.01)}} & \posdiff{-4e-3} & \makecell{0.207 \\ \tiny{(0.007)}} & \makecell{0.207 \\ \tiny{(0.019)}} & \posdiff{0.0}  &  \makecell{0.131 \\ \tiny{(0.014)}}& \makecell{0.102 \\ \tiny{(0.022)}} & \posdiff{-0.029}\\ \midrule

& & \textbf{P} $\uparrow$ & \makecell{13.40 \\ \tiny{(0.593)}} & \makecell{13.46 \\ \tiny{(0.356)}} & \posdiff{+0.06} & \makecell{14.02 \\ \tiny{(0.413)}} & \makecell{13.82 \\ \tiny{(0.68)}} & \negdiff{-0.2} & \makecell{10.79 \\ \tiny{(0.748)}} & \makecell{13.58 \\ \tiny{(1.584)}} & \posdiff{+2.79}\\

\textbf{2} & \textbf{4} & \textbf{S} $\uparrow$ & \makecell{0.307 \\ \tiny{(0.035)}} & \makecell{0.328 \\ \tiny{(0.0245)}} & \posdiff{+0.021} & \makecell{0.275 \\ \tiny{(0.067)}} & \makecell{0.245 \\ \tiny{(0.039)}} & \negdiff{-0.03} & \makecell{0.076 \\ \tiny{(0.02)}} & \makecell{0.237 \\ \tiny{(0.082)}} & \posdiff{+0.16}\\

& & \textbf{L} $\downarrow$ & \makecell{0.276 \\ \tiny{(0.0173)}} & \makecell{0.263 \\ \tiny{(0.014)}} & \posdiff{-0.013} & \makecell{0.217 \\ \tiny{(0.006)}} & \makecell{0.206 \\ \tiny{(0.019)}} & \posdiff{-0.011} & \makecell{0.134 \\ \tiny{(0.013)}} & \makecell{0.1 \\ \tiny{(0.024)}} & \posdiff{-0.034}\\

\bottomrule
\end{tabular}
}
\end{table}

\subsection{Ablation Studies}
We study factors impacting attacker information using ablations on CIFAR-10 with a LoRA-adapted ViT-Small.

\subsubsection{Impact of LoRA Rank}
\label{subsubsec:lorarankabl}

\begin{figure}[!ht]
    \centering
    \includegraphics[width=0.99\linewidth]{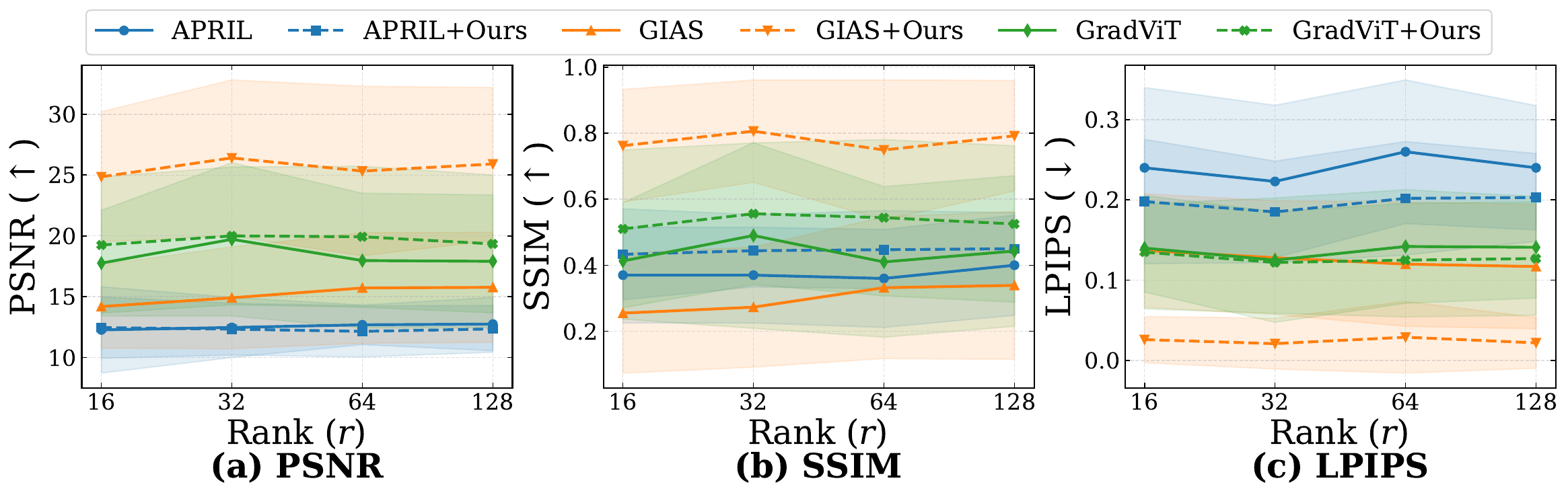}
    \caption{Performance comparison across multiple ranks $r \in \{16, 32, 64, 128\}$ for GIAS, GradViT and APRIL.}
    \label{fig:rankabl}
\end{figure}

A smaller LoRA rank increases ambiguity by reducing gradient observations, while a larger rank offers more data without resolving spatial uncertainty. As shown in Fig. \ref{fig:rankabl}, adding our positional-gradient anchors improves reconstruction most strongly for the generative baseline. Even at low rank, $r=16$, GIAS+Ours improves PSNR by 10.67 dB and SSIM by 0.507, while reducing LPIPS by 0.109. APRIL benefits more modestly, improving SSIM and LPIPS but showing only a small PSNR gain. Nevertheless, the spatial anchoring strategy improves performance for all the methods. Across ranks, GIAS+Ours remains consistently stronger than vanilla GIAS, which suggests that decreasing LoRA rank alone does not guarantee elimination of privacy risks.

\subsubsection{Impact of Batch Size}
\label{subsubsec:ablbs}
\begin{figure}[!ht]
    \centering
    \includegraphics[width=0.99\linewidth]{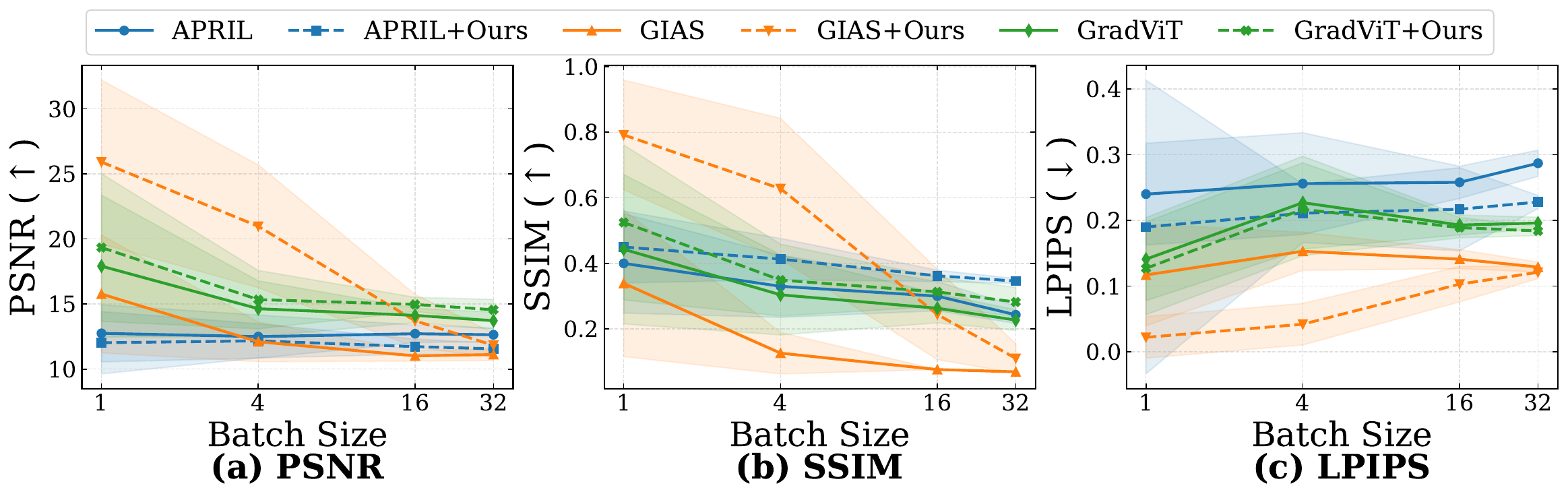}
    \caption{Performance comparison across multiple batch sizes $BS \in \{1, 4, 16, 32\}$ for GIAS, GradViT and APRIL.}
    \label{fig:bsabl}
\end{figure}

Larger batches make inversion harder because each observed gradient is a mixture of multiple images, which weakens the correspondence between any reconstructed sample and the update. Fig. \ref{fig:bsabl} shows that our spatial anchoring is most effective at small and moderate batch sizes. For GIAS, the improvement is large at batch size 1 and remains substantial at batch size 4, where applying our proposed spatial anchoring strategy provides a PSNR increase from 12.10 to 20.97, whereas SSIM increases from 0.126 to 0.628, and LPIPS decreases from 0.153 to 0.042. At batch sizes 16 and 32, the gains become smaller because batch averaging dominates the spatial signal; nevertheless, LPIPS remains lower for GIAS+Ours, showing that positional gradients improve perceptual consistency and pose greater privacy risks even at large batch sizes. Similar trends in change are also observed for APRIL and GradViT. We also provide empirical results for multiple batch sizes on the high-resolution CelebA dataset in supplementary Sec. \ref{sec:celebalargebs}.

We provide further ablations on group lazy regularization, initialization strategies, and server steps, alongside low-rank analyses (Supp. Sec. \ref{sec:glrabl}, \ref{sec:lorainit}, \ref{sec:bbinitabl}, \ref{sec:stepsabl}, \ref{sec:higherlorarank}). We also evaluate performance under secure aggregation, PEFTLeak, VLM vision encoders, and generator-client distribution gaps (Supp. Sec. \ref{sec:secagg}, \ref{sec:comppeftleak}, \ref{sec:vlmperf}, \ref{sec:distdist}, \ref{sec:femnistresults}). Overall, our method maintains high-fidelity reconstructions at lower LoRA ranks and minimal compute, as well as under secure aggregation, distribution gaps, and VLM encoders.

\section{Performance Against Defence Strategies}

\noindent
We further evaluate reconstruction robustness under lossy gradient-level privacy defences \cite{Zhu020dlg, wei2021glrfl}, comparing against the generative baseline GIAS. We consider three defence settings applied to the shared gradients: additive Gaussian noise ($\mathcal{N}$) \cite{Zhu020dlg}, gradient clipping \cite{wei2021glrfl}, and their combination. Gaussian noise with standard deviations $\sigma=0.01$ and $\sigma=0.1$ is added to the gradients, while clipping limits the gradient norm to $C = 4.0$ \cite{li2022ggl}. In the combined setting, noisy gradients are clipped before reconstruction.

\begin{figure}[!ht]
\centering
\scriptsize
\setlength{\tabcolsep}{1pt}
\newcommand{\defenseimg}[1]{%
\IfFileExists{#1}{\includegraphics[width=0.08\linewidth]{#1}}{\fbox{\rule{0pt}{0.08\linewidth}\rule{0.08\linewidth}{0pt}}}%
}
\begin{tabular}{c|cccc|cccc}
\toprule
\textbf{Method} & \multicolumn{4}{c|}{\textbf{GIAS}} & \multicolumn{4}{c}{\textbf{GIAS+Ours}} \\
\cmidrule(lr){1-5}\cmidrule(lr){6-9}
\textbf{GT} & $\mathcal{N}$ ($0.01$) & $\mathcal{N}$ ($0.1$) & \textbf{GC} & $\mathcal{N}$+\textbf{GC} & $\mathcal{N}$ ($0.01$) & $\mathcal{N}$ ($0.1$) & \textbf{GC} & $\mathcal{N}$+\textbf{GC} \\
\midrule
\defenseimg{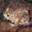}
& \defenseimg{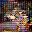}
& \defenseimg{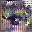}
& \defenseimg{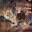}
& \defenseimg{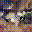}
& \defenseimg{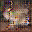}
& \defenseimg{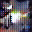}
& \defenseimg{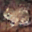}
& \defenseimg{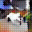} \\
\defenseimg{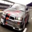}
& \defenseimg{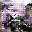}
& \defenseimg{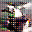}
& \defenseimg{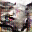}
& \defenseimg{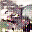}
& \defenseimg{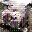}
& \defenseimg{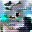}
& \defenseimg{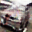}
& \defenseimg{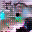} \\ %
\bottomrule
\end{tabular}
\caption{Qualitative comparison of GIAS and GIAS+Ours under gradient-level defence strategies. Numbers within parentheses denote the standard deviation of the noise applied to the gradients.}
\label{fig:defense_reconstructions}
\end{figure}

As shown in Fig. \ref{fig:defense_reconstructions}, under mild perturbations and gradient clipping, GIAS+Ours achieves superior semantic fidelity over GIAS due to our spatial anchoring strategy. However, under strong perturbations, the recoverable information in the shared gradients is substantially reduced, resulting in noticeable degradation in reconstruction quality for both methods.  The quantitative results in Tab. \ref{tab:defencetable} further confirm that strong lossy gradient perturbations effectively mitigate reconstruction attacks on low-rank adapted ViTs.

\begin{table}[!ht]
\centering
\caption{Quantitative comparison of GIAS and GIAS+Ours under gradient-level defence strategies. $\mathcal{N}$, GC, P, S and L denote additive noise, gradient clipping, PSNR, SSIM and LPIPS, respectively. Numbers within brackets indicate the standard deviation.}
\scriptsize
\setlength{\tabcolsep}{0.5pt}
\newcommand{\defenseimg}[1]{%
\IfFileExists{#1}{\includegraphics[width=0.08\linewidth]{#1}}{\fbox{\rule{0pt}{0.08\linewidth}\rule{0.08\linewidth}{0pt}}}%
}
\begin{tabular}{c|cccc|cccc}
\toprule
\multirow{2}{*}{\rotatebox{90}{\textbf{Metric}}} & \multicolumn{4}{c|}{\textbf{GIAS}} & \multicolumn{4}{c}{\textbf{GIAS+Ours}} \\
\cmidrule(lr){2-5}\cmidrule(lr){6-9}
 & $\mathcal{N}$ ($0.01$) & $\mathcal{N}$ ($0.1$) & \textbf{GC} & $\mathcal{N}$+\textbf{GC} & $\mathcal{N}$ ($0.01$) & $\mathcal{N}$ ($0.1$) & \textbf{GC} & $\mathcal{N}$+\textbf{GC} \\
\midrule
P &  
\makecell{13.11 \tiny{(3.73)}} &
\makecell{9.86 \tiny{(1.92)}} &
\makecell{13.9 \tiny{(4.12)}} &
\makecell{9.72 \tiny{(1.88)}} &
\makecell{14.19 \tiny{(4.33)}} &
\makecell{9.28 \tiny{(1.38)}} &
\makecell{25.39 \tiny{(4.81)}} &
\makecell{9.31 \tiny{(1.41)}} \\
S &  
\makecell{0.23 \tiny{(0.19)}} &
\makecell{0.04 \tiny{(0.06)}} &
\makecell{0.30 \tiny{(0.24)}} &
\makecell{0.05 \tiny{(0.06)}} &
\makecell{0.31 \tiny{(0.21)}} &
\makecell{0.03 \tiny{(0.05)}} &
\makecell{0.77 \tiny{(0.13)}}  &
\makecell{0.03 \tiny{(0.06)}}\\
L &  
\makecell{0.13 \tiny{(0.08)}} &
\makecell{0.17 \tiny{(0.06)}} &
\makecell{0.13 \tiny{(0.06)}} &
\makecell{0.15 \tiny{(0.07)}} &
\makecell{0.11 \tiny{(0.07)}} &
\makecell{0.19 \tiny{(0.06)}} &
\makecell{0.02 \tiny{(0.02)}}  &
\makecell{0.19 \tiny{(0.06)}}\\
\bottomrule
\end{tabular}
\label{tab:defencetable}
\end{table}

\section{Empirical Privacy-Utility Trade-Off}
\label{sec:putradeoff}
{In this section, we investigate whether perturbing client gradients through clipping and Gaussian noise reduces reconstruction fidelity while preserving the global model's predictive utility. We train the model throughout the complete FedAvg process, {where the data is distributed among the clients following a Dirichlet distribution with parameter $\alpha=0.1$}, under different gradient perturbation strengths. For each configuration, we evaluate utility using test accuracy and quantify empirical leakage using the PSNR between the reconstructed and original samples obtained under a fixed gradient inversion attack, along with PEFTLeak \mbox{\cite{Sami2025gipeft}} (see Supp. Sec. \mbox{\ref{sec:comppeftleak}} for comparison of the proposed method with PEFTLeak).}
\begin{figure}[!ht]
    \centering
    \begin{minipage}[!ht]{1.0\linewidth}
        \centering
        \includegraphics[width=0.8\linewidth]{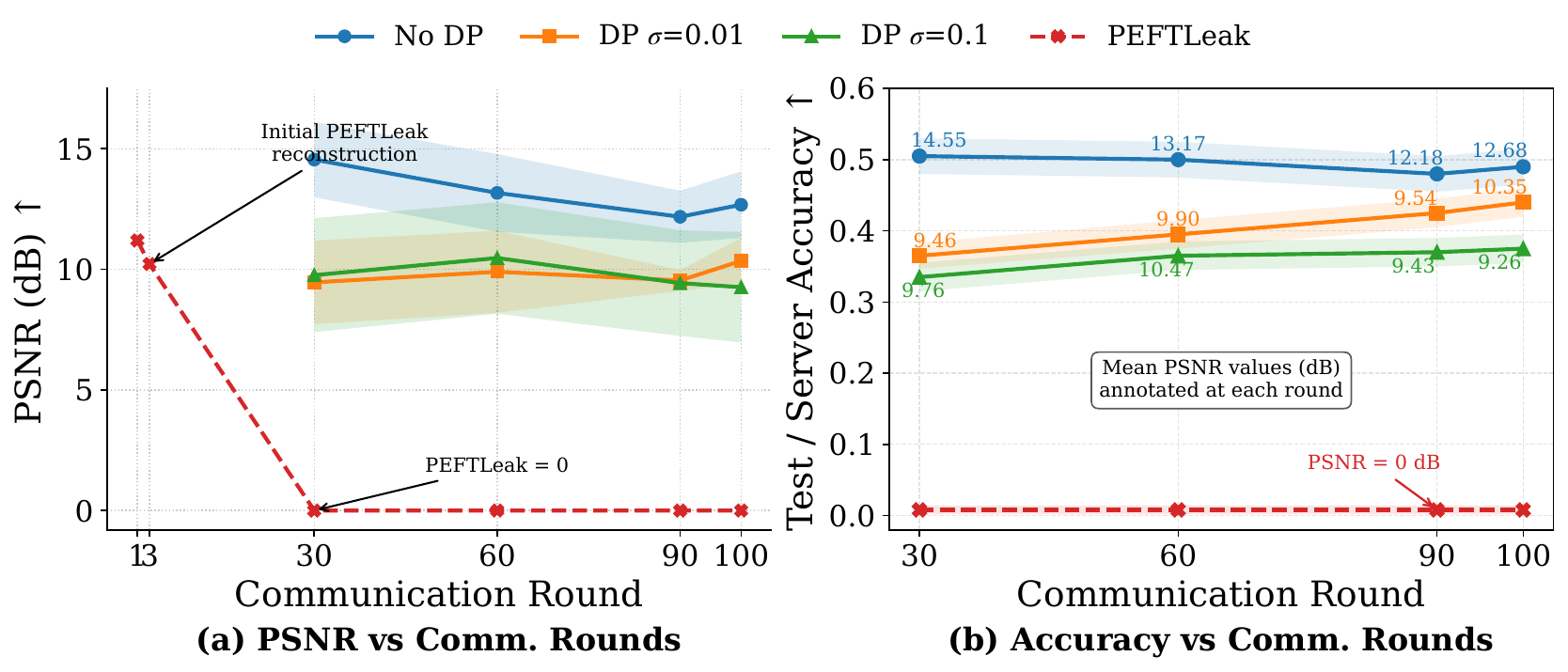}
        \caption{Lower PSNR indicates reduced privacy leakage, while higher test accuracy reflects better model utility. PEFTLeak drops to zero in both attack and utility as the poisoned weight changes, and it fails to reconstruct any information. See Supp. Sec. \ref{sec:moreputoff} for SSIM and LPIPS plots.}
        \label{fig:putoff}
    \end{minipage}
\end{figure}
{In Fig. \mbox{\ref{fig:putoff}}, we observe that increasing the perturbation strength reduces the gradient information available to the server and reconstruction quality, as indicated by a lower PSNR; it distorts the optimization signal too, reducing downstream accuracy, highlighting the inherent privacy–utility trade-off.}

\section{Limitations}
\label{sec:limitations}
{
While differential-privacy induced gradient perturbation effectively mitigates gradient leakage under standard reconstruction attacks, it introduces an inherent trade-off by reducing downstream task performance. Furthermore, ensuring robustness against adaptive adversaries and complex multimodal architectures (like VLMs) in heterogeneous federated settings remains a key research direction.
}

\section{Conclusion}

From our analysis, we find that LoRA-based PEFT reduces the number of trainable parameters exposed to the server but does not eliminate privacy leakage. Instead, frozen positional embeddings weaken spatial cues, increasing inversion ambiguity and non-identifiability. To address this, we propose a spatially anchored generative inversion framework that makes positional embeddings trainable across transformer blocks, providing additional spatial constraints while leveraging a generative prior for realistic reconstructions. Experiments on CIFAR-10, CelebA, and CARS demonstrate substantial improvements in reconstruction quality, including under aggregated multi-step updates and multiple client samples on CIFAR-10. These results highlight that PEFT-aware privacy defences should consider not only the number of trainable parameters but also the type of information they expose, particularly spatial information in federated ViTs.

\section*{Acknowledgement}

This work is supported by the P3DX project, seed-funded by the Ministry of Electronics and Information Technology (MeitY), GoI. The authors would also like to acknowledge compute support received from Kotak-IISc AI-ML Centre (KIAC), IISc.

{\small
\bibliographystyle{ieeenat_fullname}
\bibliography{main}

\begin{thebibliography}{62}
\providecommand{\natexlab}[1]{#1}
\providecommand{\url}[1]{\texttt{#1}}
\expandafter\ifx\csname urlstyle\endcsname\relax
  \providecommand{\doi}[1]{doi: #1}\else
  \providecommand{\doi}{doi: \begingroup \urlstyle{rm}\Url}\fi

\bibitem[Abnar and Zuidema(2020)]{abnar2020}
Samira Abnar and Willem Zuidema.
\newblock Quantifying attention flow in transformers.
\newblock In \emph{Proceedings of the 58th Annual Meeting of the Association
  for Computational Linguistics}, pages 4190--4197, Online, 2020. Association
  for Computational Linguistics.

\bibitem[Achlioptas(2003)]{achlioptas2003database}
Dimitris Achlioptas.
\newblock Database-friendly random projections: Johnson-lindenstrauss with
  binary coins.
\newblock \emph{Journal of Computer and System Sciences}, 66\penalty0
  (4):\penalty0 671–687, 2003.

\bibitem[Aghajanyan et~al.(2021)Aghajanyan, Gupta, and
  Zettlemoyer]{aghajanyan2021intrinsic}
Armen Aghajanyan, Sonal Gupta, and Luke Zettlemoyer.
\newblock Intrinsic dimensionality explains the effectiveness of language model
  fine-tuning.
\newblock In \emph{Proceedings of the 59th Annual Meeting of the Association
  for Computational Linguistics and the 11th International Joint Conference on
  Natural Language Processing (Volume 1: Long Papers)}, pages 7319--7328,
  Online, 2021. Association for Computational Linguistics.

\bibitem[Balunovic et~al.(2022)Balunovic, Dimitrov, Jovanovic, and
  Vechev]{BalunovicD0V22lamp}
Mislav Balunovic, Dimitar~I. Dimitrov, Nikola Jovanovic, and Martin~T. Vechev.
\newblock {LAMP:} extracting text from gradients with language model priors.
\newblock In \emph{Advances in Neural Information Processing Systems 35: Annual
  Conference on Neural Information Processing Systems}, 2022.

\bibitem[Benning and Burger(2018)]{BenningB18mrminvprobs}
Martin Benning and Martin Burger.
\newblock Modern regularization methods for inverse problems.
\newblock \emph{Acta Numer.}, 27:\penalty0 1--111, 2018.

\bibitem[Bonawitz et~al.(2017)Bonawitz, Ivanov, Kreuter, Marcedone, McMahan,
  Patel, Ramage, Segal, and Seth]{bonawitz2017randommasking}
Keith Bonawitz, Vladimir Ivanov, Ben Kreuter, Antonio Marcedone, H.~Brendan
  McMahan, Sarvar Patel, Daniel Ramage, Aaron Segal, and Karn Seth.
\newblock Practical secure aggregation for privacy-preserving machine learning.
\newblock In \emph{Proceedings of the 2017 ACM SIGSAC Conference on Computer
  and Communications Security}, page 1175–1191, New York, NY, USA, 2017.
  Association for Computing Machinery.

\bibitem[Bonawitz et~al.(2019)Bonawitz, Eichner, Grieskamp, Huba, Ingerman,
  Ivanov, Kiddon, Kone{\v{c}}n{\'y}, Mazzocchi, McMahan, Overveldt, Petrou,
  Ramage, and Roselander]{BonawitzEGHIIKK19fedlscale}
Kallista~A. Bonawitz, Hubert Eichner, Wolfgang Grieskamp, Dzmitry Huba, Alex
  Ingerman, Vladimir Ivanov, Chlo{\'{e}} Kiddon, Jakub Kone{\v{c}}n{\'y},
  Stefano Mazzocchi, Brendan McMahan, Timon~Van Overveldt, David Petrou, Daniel
  Ramage, and Jason Roselander.
\newblock Towards federated learning at scale: System design.
\newblock In \emph{Proceedings of the Second Conference on Machine Learning and
  Systems, SysML 2019, Stanford, CA, USA, March 31 - April 2, 2019}. mlsys.org,
  2019.

\bibitem[Cand{\`{e}}s et~al.(2006)Cand{\`{e}}s, Romberg, and
  Tao]{CandesRT06RUP}
Emmanuel~J. Cand{\`{e}}s, Justin~K. Romberg, and Terence Tao.
\newblock Robust uncertainty principles: exact signal reconstruction from
  highly incomplete frequency information.
\newblock \emph{{IEEE} Transactions on Information Theory}, 52\penalty0
  (2):\penalty0 489--509, 2006.

\bibitem[Chaudhry and Mallipeddi(2025)]{Chaudhry2025egap}
Yuvraj~Singh Chaudhry and Rammohan Mallipeddi.
\newblock E-gap: Evolutionary gradient attack on privacy.
\newblock \emph{Computers and Electrical Engineering}, 124:\penalty0 110399,
  2025.

\bibitem[Deng et~al.(2021)Deng, Wang, Li, Wang, Shang, Liu, Rajasekaran, and
  Ding]{DengWLWSLRD21TAG}
Jieren Deng, Yijue Wang, Ji Li, Chenghong Wang, Chao Shang, Hang Liu,
  Sanguthevar Rajasekaran, and Caiwen Ding.
\newblock {TAG:} gradient attack on transformer-based language models.
\newblock In \emph{Findings of the Association for Computational Linguistics:
  {EMNLP}}, pages 3600--3610. Association for Computational Linguistics, 2021.

\bibitem[Dimitrov et~al.(2024)Dimitrov, Baader, M\"{u}ller, and
  Vechev]{dimitar2024spear}
Dimitar~I. Dimitrov, Maximilian Baader, Mark~Niklas M\"{u}ller, and Martin
  Vechev.
\newblock Spear: exact gradient inversion of batches in federated learning.
\newblock In \emph{Proceedings of the 38th International Conference on Neural
  Information Processing Systems}, Red Hook, NY, USA, 2024. Curran Associates
  Inc.

\bibitem[Dosovitskiy et~al.(2021)Dosovitskiy, Beyer, Kolesnikov, Weissenborn,
  Zhai, Unterthiner, Dehghani, Minderer, Heigold, Gelly, Uszkoreit, and
  Houlsby]{dosovitskiy2021vit}
Alexey Dosovitskiy, Lucas Beyer, Alexander Kolesnikov, Dirk Weissenborn,
  Xiaohua Zhai, Thomas Unterthiner, Mostafa Dehghani, Matthias Minderer, Georg
  Heigold, Sylvain Gelly, Jakob Uszkoreit, and Neil Houlsby.
\newblock An image is worth 16x16 words: Transformers for image recognition at
  scale.
\newblock In \emph{International Conference on Learning Representations}, 2021.

\bibitem[Facco et~al.(2017)Facco, d’Errico, Rodriguez, and
  Laio]{Facco2017twonn}
Elena Facco, Maria d’Errico, Alex Rodriguez, and Alessandro Laio.
\newblock Estimating the intrinsic dimension of datasets by a minimal
  neighborhood information.
\newblock \emph{Scientific Reports}, 7\penalty0 (1), 2017.

\bibitem[Fang et~al.(2023)Fang, Chen, Wang, Wang, and Xia]{fang2023gifd}
Hao Fang, Bin Chen, Xuan Wang, Zhi Wang, and Shu-Tao Xia.
\newblock Gifd: A generative gradient inversion method with feature domain
  optimization.
\newblock In \emph{Proceedings of the IEEE/CVF International Conference on
  Computer Vision}, pages 4967--4976, 2023.

\bibitem[Feng and Tram\`{e}r(2024)]{feng2024privacybckdoor}
Shanglun Feng and Florian Tram\`{e}r.
\newblock Privacy backdoors: stealing data with corrupted pretrained models.
\newblock In \emph{Proceedings of the 41st International Conference on Machine
  Learning}. JMLR.org, 2024.

\bibitem[Feng et~al.(2024)Feng, Ma, Wang, Chegne, Ma, Abuadbba, and
  Bai]{feng2024grab}
Xinguo Feng, Zhongkui Ma, Zihan Wang, Eu~Joe Chegne, Mengyao Ma, Alsharif
  Abuadbba, and Guangdong Bai.
\newblock Uncovering gradient inversion risks in practical language model
  training.
\newblock In \emph{Proceedings of the 2024 on ACM SIGSAC Conference on Computer
  and Communications Security}, page 3525–3539, New York, NY, USA, 2024.
  Association for Computing Machinery.

\bibitem[Ferrando et~al.(2022)Ferrando, G{\'a}llego, and
  Costa-juss{\`a}]{ferrando2022measuring}
Javier Ferrando, Gerard~I. G{\'a}llego, and Marta~R. Costa-juss{\`a}.
\newblock Measuring the mixing of contextual information in the transformer.
\newblock In \emph{Proceedings of the 2022 Conference on Empirical Methods in
  Natural Language Processing}, pages 8698--8714, Abu Dhabi, United Arab
  Emirates, 2022. Association for Computational Linguistics.

\bibitem[Fowl et~al.(2022)Fowl, Geiping, Czaja, Goldblum, and
  Goldstein]{fowl2022robbing}
Liam~H Fowl, Jonas Geiping, Wojciech Czaja, Micah Goldblum, and Tom Goldstein.
\newblock Robbing the fed: Directly obtaining private data in federated
  learning with modified models.
\newblock In \emph{International Conference on Learning Representations}, 2022.

\bibitem[Fowl et~al.(2023)Fowl, Geiping, Reich, Wen, Czaja, Goldblum, and
  Goldstein]{fowl2023decepticons}
Liam~H Fowl, Jonas Geiping, Steven Reich, Yuxin Wen, Wojciech Czaja, Micah
  Goldblum, and Tom Goldstein.
\newblock Decepticons: Corrupted transformers breach privacy in federated
  learning for language models.
\newblock In \emph{The Eleventh International Conference on Learning
  Representations}, 2023.

\bibitem[Geiping et~al.(2020)Geiping, Bauermeister, Dr\"{o}ge, and
  Moeller]{geiping2020igfl}
Jonas Geiping, Hartmut Bauermeister, Hannah Dr\"{o}ge, and Michael Moeller.
\newblock Inverting gradients - how easy is it to break privacy in federated
  learning?
\newblock In \emph{Proc. of the 34th International Conference on Neural
  Information Processing Systems}, 2020.

\bibitem[Goodfellow et~al.(2014)Goodfellow, Pouget-Abadie, Mirza, Xu,
  Warde-Farley, Ozair, Courville, and Bengio]{goodfellow2014gans}
Ian~J. Goodfellow, Jean Pouget-Abadie, Mehdi Mirza, Bing Xu, David
  Warde-Farley, Sherjil Ozair, Aaron Courville, and Yoshua Bengio.
\newblock Generative adversarial nets.
\newblock In \emph{Proceedings of the 28th International Conference on Neural
  Information Processing Systems - Volume 2}, page 2672–2680, 2014.

\bibitem[Gupta et~al.(2022)Gupta, Huang, Zhong, Gao, Li, and
  Chen]{gupta2022film}
Samyak Gupta, Yangsibo Huang, Zexuan Zhong, Tianyu Gao, Kai Li, and Danqi Chen.
\newblock Recovering private text in federated learning of language models.
\newblock In \emph{Proceedings of the 36th International Conference on Neural
  Information Processing Systems}, 2022.

\bibitem[Hatamizadeh et~al.(2022)Hatamizadeh, Yin, Roth, Li, Kautz, Xu, and
  Molchanov]{ali2022gradvit}
Ali Hatamizadeh, Hongxu Yin, Holger Roth, Wenqi Li, Jan Kautz, Daguang Xu, and
  Pavlo Molchanov.
\newblock { GradViT: Gradient Inversion of Vision Transformers }.
\newblock In \emph{2022 IEEE/CVF Conference on Computer Vision and Pattern
  Recognition (CVPR)}, pages 10011--10020. IEEE Computer Society, 2022.

\bibitem[Hu et~al.(2022)Hu, Shen, Wallis, Allen{-}Zhu, Li, Wang, Wang, and
  Chen]{HuSWALWWC22lora}
Edward~J. Hu, Yelong Shen, Phillip Wallis, Zeyuan Allen{-}Zhu, Yuanzhi Li,
  Shean Wang, Lu Wang, and Weizhu Chen.
\newblock Lora: Low-rank adaptation of large language models.
\newblock In \emph{The Tenth International Conference on Learning
  Representations, {ICLR} 2022, Virtual Event, April 25-29, 2022}.
  OpenReview.net, 2022.

\bibitem[Jeon et~al.(2021)Jeon, Kim, Lee, Oh, and Ok]{gias}
Jinwoo Jeon, Jaechang Kim, Kangwook Lee, Sewoong Oh, and Jungseul Ok.
\newblock Gradient inversion with generative image prior.
\newblock In \emph{Proceedings of the 35th International Conference on Neural
  Information Processing Systems}, 2021.

\bibitem[Johnson and Lindenstrauss(1984)]{Johnson1984jllemma}
William~B. Johnson and Joram Lindenstrauss.
\newblock Extensions of lipschitz mappings into a hilbert space.
\newblock \emph{Conference on Modern Analysis and Probability}, page 189–206,
  1984.

\bibitem[Kariyappa et~al.(2023)Kariyappa, Guo, Maeng, Xiong, Suh, Qureshi, and
  Lee]{sanjay2023cpa}
Sanjay Kariyappa, Chuan Guo, Kiwan Maeng, Wenjie Xiong, G.~Edward Suh,
  Moinuddin~K Qureshi, and Hsien-Hsin~S. Lee.
\newblock Cocktail party attack: breaking aggregation-based privacy in
  federated learning using independent component analysis.
\newblock In \emph{Proceedings of the 40th International Conference on Machine
  Learning}. JMLR.org, 2023.

\bibitem[Karras et~al.(2017)Karras, Aila, Laine, and
  Lehtinen]{karras2017progressive}
Tero Karras, Timo Aila, Samuli Laine, and Jaakko Lehtinen.
\newblock Progressive growing of gans for improved quality, stability, and
  variation.
\newblock \emph{arXiv preprint arXiv:1710.10196}, 2017.

\bibitem[Karras et~al.(2020)Karras, Laine, Aittala, Hellsten, Lehtinen, and
  Aila]{karras2020stylegan2}
Tero Karras, Samuli Laine, Miika Aittala, Janne Hellsten, Jaakko Lehtinen, and
  Timo Aila.
\newblock Analyzing and improving the image quality of stylegan.
\newblock In \emph{Proceedings of the IEEE/CVF Conference on Computer Vision
  and Pattern Recognition (CVPR)}, pages 8110--8119, 2020.

\bibitem[Karras et~al.(2026)Karras, Aila, Laine, and
  Lehtinen]{karras2026celebAhq}
Tero Karras, Timo Aila, Samuli Laine, and Jaakko Lehtinen.
\newblock Celeba-hq, 2026.

\bibitem[Krause et~al.(2013)Krause, Stark, Deng, and Fei-Fei]{Krause2013cars}
Jonathan Krause, Michael Stark, Jia Deng, and Li Fei-Fei.
\newblock 3d object representations for fine-grained categorization.
\newblock In \emph{Proceedings of the IEEE International Conference on Computer
  Vision (ICCV) Workshops}, 2013.

\bibitem[LeCun et~al.(1989)LeCun, Boser, Denker, Henderson, Howard, Hubbard,
  and Jackel]{lecun1989cnn}
Y. LeCun, B. Boser, J.~S. Denker, D. Henderson, R.~E. Howard, W. Hubbard, and
  L.~D. Jackel.
\newblock Backpropagation applied to handwritten zip code recognition.
\newblock \emph{Neural Computation}, 1\penalty0 (4):\penalty0 541--551, 1989.

\bibitem[Lee et~al.(2025)Lee, Ham, and Kim]{LeeHK25mpvg}
Wonjun Lee, Bumsub Ham, and Suhyun Kim.
\newblock Maximizing the position embedding for vision transformers with global
  average pooling.
\newblock In \emph{Thirty-Ninth {AAAI} Conference on Artificial Intelligence,
  Thirty-Seventh Conference on Innovative Applications of Artificial
  Intelligence, Fifteenth Symposium on Educational Advances in Artificial
  Intelligence, {AAAI} 2025, Philadelphia, PA, USA, February 25 - March 4,
  2025}, pages 18154--18162. {AAAI} Press, 2025.

\bibitem[Levina and Bickel(2004)]{levina2004mle}
Elizaveta Levina and Peter Bickel.
\newblock Maximum likelihood estimation of intrinsic dimension.
\newblock In \emph{Advances in Neural Information Processing Systems}. MIT
  Press, 2004.

\bibitem[Li et~al.(2018)Li, Farkhoor, Liu, and Yosinski]{li2018intdim}
Chunyuan Li, Heerad Farkhoor, Rosanne Liu, and Jason Yosinski.
\newblock Measuring the intrinsic dimension of objective landscapes.
\newblock In \emph{6th International Conference on Learning Representations,
  {ICLR} 2018, Vancouver, BC, Canada, April 30 - May 3, 2018, Conference Track
  Proceedings}. OpenReview.net, 2018.

\bibitem[Li et~al.(2022)Li, Zhang, Liu, and Liu]{li2022ggl}
Zhuohang Li, Jiaxin Zhang, Luyang Liu, and Jian Liu.
\newblock Auditing privacy defenses in federated learning via generative
  gradient leakage.
\newblock In \emph{2022 IEEE/CVF Conference on Computer Vision and Pattern
  Recognition (CVPR)}, pages 10122--10132, 2022.

\bibitem[Liu et~al.(2026)Liu, Zhang, Huang, Zheng, Yang, and Zhang]{Liu2026nrr}
Sheng Liu, Xuefeng Zhang, Tao Huang, Guolong Zheng, Wencheng Yang, and Ji
  Zhang.
\newblock Probing privacy risks in deep neural networks: A jacobian rank-based
  analysis of intermediate layers.
\newblock \emph{Intelligent Computing}, 5, 2026.

\bibitem[Liu et~al.(2015)Liu, Luo, Wang, and Tang]{liu2015celebA}
Ziwei Liu, Ping Luo, Xiaogang Wang, and Xiaoou Tang.
\newblock Deep learning face attributes in the wild.
\newblock In \emph{Proceedings of International Conference on Computer Vision
  (ICCV)}, 2015.

\bibitem[Liu et~al.(2022)Liu, Chen, Zhao, Yu, Liu, Bao, Jiang, Nie, Xu, and
  Yang]{Liu2022carefl}
Zelei Liu, Yuanyuan Chen, Yansong Zhao, Han Yu, Yang Liu, Renyi Bao, Jinpeng
  Jiang, Zaiqing Nie, Qian Xu, and Qiang Yang.
\newblock Contribution-aware federated learning for smart healthcare.
\newblock \emph{Proceedings of the AAAI Conference on Artificial Intelligence},
  36\penalty0 (11):\penalty0 12396–12404, 2022.

\bibitem[Lu et~al.(2022)Lu, Zhang, Zhao, He, and Cheng]{jiahao2022april}
Jiahao Lu, Xi~Sheryl Zhang, Tianli Zhao, Xiangyu He, and Jian Cheng.
\newblock April: Finding the achilles' heel on privacy for vision transformers.
\newblock In \emph{2022 IEEE/CVF Conference on Computer Vision and Pattern
  Recognition (CVPR)}, pages 10041--10050, 2022.

\bibitem[McMahan et~al.(2017)McMahan, Moore, Ramage, Hampson, and
  Arcas]{mcmahan17fedavg}
Brendan McMahan, Eider Moore, Daniel Ramage, Seth Hampson, and Blaise Aguera~y
  Arcas.
\newblock {Communication-Efficient Learning of Deep Networks from Decentralized
  Data}.
\newblock In \emph{Proceedings of the 20th International Conference on
  Artificial Intelligence and Statistics}, pages 1273--1282. PMLR, 2017.

\bibitem[Ong et~al.(2026)Ong, Bobrowski, Reinert, and Skraba]{ong2026l2n2}
Eng{-}Jon Ong, Omer Bobrowski, Gesine Reinert, and Primoz Skraba.
\newblock A universal nearest-neighbor estimator for intrinsic dimensionality.
\newblock \emph{CoRR}, abs/2603.10493, 2026.

\bibitem[Petrov et~al.(2024)Petrov, Dimitrov, Baader, M\"{u}ller, and
  Vechev]{petrov2024dager}
Ivo Petrov, Dimitar~I. Dimitrov, Maximilian Baader, Mark~Niklas M\"{u}ller, and
  Martin Vechev.
\newblock Dager: exact gradient inversion for large language models.
\newblock In \emph{Proceedings of the 38th International Conference on Neural
  Information Processing Systems}, 2024.

\bibitem[Roy and Vetterli(2007)]{roy2007effective}
Olivier Roy and Martin Vetterli.
\newblock The effective rank: {A} measure of effective dimensionality.
\newblock In \emph{15th European Signal Processing Conference, {EUSIPCO} 2007,
  Poznan, Poland, September 3-7, 2007}, pages 606--610. {IEEE}, 2007.

\bibitem[Rumelhart et~al.(1986)Rumelhart, Hinton, and
  Williams]{Rumelhart1986fcnn}
David~E. Rumelhart, Geoffrey~E. Hinton, and Ronald~J. Williams.
\newblock Learning representations by back-propagating errors.
\newblock \emph{Nature}, 323\penalty0 (6088):\penalty0 533–536, 1986.

\bibitem[Sami et~al.(2025)Sami, Sen, Roy{-}Chowdhury, Krishnamurthy, and
  Guler]{Sami2025gipeft}
Hasin~Us Sami, Swapneel Sen, Amit~K. Roy{-}Chowdhury, Srikanth~V.
  Krishnamurthy, and Basak Guler.
\newblock Gradient inversion attacks on parameter-efficient fine-tuning.
\newblock In \emph{{IEEE/CVF} Conference on Computer Vision and Pattern
  Recognition, {CVPR} 2025, Nashville, TN, USA, June 11-15, 2025}, pages
  10224--10234. Computer Vision Foundation / {IEEE}, 2025.

\bibitem[Sami et~al.(2026)Sami, Sen, and Guler]{sami2026minegrad}
Hasin~Us Sami, Swapneel Sen, and Basak Guler.
\newblock Minegrad: Gradient inversion attacks on lo{RA} fine-tuning.
\newblock In \emph{The 29th International Conference on Artificial Intelligence
  and Statistics}, 2026.

\bibitem[Sánchez~Sánchez et~al.(2024)Sánchez~Sánchez, Huertas~Celdrán,
  Xie, Bovet, Martínez~Pérez, and Stiller]{SnchezSnchez2024fedtrust}
Pedro~Miguel Sánchez~Sánchez, Alberto Huertas~Celdrán, Ning Xie, Gér\^ome
  Bovet, Gregorio Martínez~Pérez, and Burkhard Stiller.
\newblock Federatedtrust: A solution for trustworthy federated learning.
\newblock \emph{Future Generation Computer Systems}, 152:\penalty0 83–98,
  2024.

\bibitem[Tariq et~al.(2024)Tariq, Serhani, Sallabi, Barka, Qayyum, Khater, and
  Shuaib]{Tariq2024tflsurvey}
Asadullah Tariq, Mohamed~Adel Serhani, Farag~M. Sallabi, Ezedin~S. Barka, Tariq
  Qayyum, Heba~M. Khater, and Khaled~A. Shuaib.
\newblock Trustworthy federated learning: A comprehensive review, architecture,
  key challenges, and future research prospects.
\newblock \emph{IEEE Open Journal of the Communications Society}, 5:\penalty0
  4920–4998, 2024.

\bibitem[Wei et~al.(2021)Wei, Liu, Wu, Su, and Iyengar]{wei2021glrfl}
Wenqi Wei, Ling Liu, Yanzhao Wu, Gong Su, and Arun Iyengar.
\newblock { Gradient-Leakage Resilient Federated Learning }.
\newblock In \emph{2021 IEEE 41st International Conference on Distributed
  Computing Systems (ICDCS)}, pages 797--807, Los Alamitos, CA, USA, 2021. IEEE
  Computer Society.

\bibitem[Ye et~al.(2024)Ye, Luo, Zhou, and Tang]{Ye2024stepsiseGIA}
Zipeng Ye, Wenjian Luo, Qi Zhou, and Yubo Tang.
\newblock High-fidelity gradient inversion in distributed learning.
\newblock \emph{Proceedings of the AAAI Conference on Artificial Intelligence},
  38\penalty0 (18):\penalty0 19983–19991, 2024.

\bibitem[Yin et~al.(2021)Yin, Mallya, Vahdat, Alvarez, Kautz, and
  Molchanov]{yin2021gradinversion}
Hongxu Yin, Arun Mallya, Arash Vahdat, Jose~M. Alvarez, Jan Kautz, and Pavlo
  Molchanov.
\newblock { See through Gradients: Image Batch Recovery via GradInversion }.
\newblock In \emph{2021 IEEE/CVF Conference on Computer Vision and Pattern
  Recognition (CVPR)}, pages 16332--16341, Los Alamitos, CA, USA, 2021. IEEE
  Computer Society.

\bibitem[Yu et~al.(2015)Yu, Zhang, Song, Seff, and Xiao]{YuZSSX15LSUN}
Fisher Yu, Yinda Zhang, Shuran Song, Ari Seff, and Jianxiong Xiao.
\newblock {LSUN:} construction of a large-scale image dataset using deep
  learning with humans in the loop.
\newblock \emph{CoRR}, abs/1506.03365, 2015.

\bibitem[Yu et~al.(2023)Yu, Wang, Wang, Li, Liu, Duan, Ji, and Chen]{lape}
Runyi Yu, Zhennan Wang, Yinhuai Wang, Kehan Li, Chang Liu, Haoyi Duan,
  Xiangyang Ji, and Jie Chen.
\newblock Lape: Layer-adaptive position embedding for vision transformers with
  independent layer normalization.
\newblock In \emph{2023 IEEE/CVF International Conference on Computer Vision
  (ICCV)}, pages 5863--5873, 2023.

\bibitem[Zhai et~al.(2023)Zhai, Mustafa, Kolesnikov, and Beyer]{Zhai2023siglip}
Xiaohua Zhai, Basil Mustafa, Alexander Kolesnikov, and Lucas Beyer.
\newblock Sigmoid loss for language image pre-training.
\newblock In \emph{Proceedings of the IEEE/CVF International Conference on
  Computer Vision (ICCV)}, pages 11975--11986, 2023.

\bibitem[Zhang et~al.(2024)Zhang, Zeng, Zhang, Wang, Wang, Zhou, Liang, and
  Qu]{zhang2024flhetbench}
Junyuan Zhang, Shuang Zeng, Miao Zhang, Runxi Wang, Feifei Wang, Yuyin Zhou,
  Paul~Pu Liang, and Liangqiong Qu.
\newblock Flhetbench: Benchmarking device and state heterogeneity in federated
  learning.
\newblock In \emph{2024 IEEE/CVF Conference on Computer Vision and Pattern
  Recognition (CVPR)}, pages 12098--12108, 2024.

\bibitem[Zhang et~al.(2025)Zhang, Liu, and Chen]{Zhang2025loraone}
Yuanhe Zhang, Fanghui Liu, and Yudong Chen.
\newblock Lora-one: One-step full gradient could suffice for fine-tuning large
  language models, provably and efficiently.
\newblock In \emph{Forty-second International Conference on Machine Learning,
  {ICML} 2025, Vancouver, BC, Canada, July 13-19, 2025}. {PMLR} /
  OpenReview.net, 2025.

\bibitem[Zhang et~al.(2023)Zhang, Yang, Dai, Wang, Yu, Qu, and
  Xu]{zhang2023fedpetuning}
Zhuo Zhang, Yuanhang Yang, Yong Dai, Qifan Wang, Yue Yu, Lizhen Qu, and Zenglin
  Xu.
\newblock {F}ed{PET}uning: When federated learning meets the
  parameter-efficient tuning methods of pre-trained language models.
\newblock In \emph{Findings of the Association for Computational Linguistics:
  ACL 2023}, pages 9963--9977, Toronto, Canada, 2023. Association for
  Computational Linguistics.

\bibitem[Zhao et~al.(2020)Zhao, Mopuri, and Bilen]{zhao2020idlg}
Bo Zhao, Konda~Reddy Mopuri, and Hakan Bilen.
\newblock idlg: Improved deep leakage from gradients.
\newblock \emph{CoRR}, abs/2001.02610, 2020.

\bibitem[Zhao et~al.(2024)Zhao, Sharma, Elkordy, Ezzeldin, Avestimehr, and
  Bagchi]{zhao2024loki}
Joshua~C. Zhao, Atul Sharma, Ahmed~Roushdy Elkordy, Yahya~H. Ezzeldin, Salman
  Avestimehr, and Saurabh Bagchi.
\newblock { Loki: Large-scale Data Reconstruction Attack against Federated
  Learning through Model Manipulation }.
\newblock In \emph{2024 IEEE Symposium on Security and Privacy (SP)}, pages
  1287--1305, Los Alamitos, CA, USA, 2024. IEEE Computer Society.

\bibitem[Zhu and Blaschko(2021)]{ZhuB21rgap}
Junyi Zhu and Matthew~B. Blaschko.
\newblock {R-GAP:} recursive gradient attack on privacy.
\newblock In \emph{9th International Conference on Learning Representations,
  {ICLR} 2021, Virtual Event, Austria, May 3-7, 2021}. OpenReview.net, 2021.

\bibitem[Zhu and Han(2020)]{Zhu020dlg}
Ligeng Zhu and Song Han.
\newblock Deep leakage from gradients.
\newblock In \emph{Federated Learning - Privacy and Incentive}, pages 17--31.
  Springer, 2020.

\end{thebibliography}
}

\clearpage
\newpage
\appendix

\begin{strip}
    \centering
    \hrule
    \vspace{0.2cm}
    {\LARGE \bfseries Supplementary Material for: \\ [0.2cm]}
    {\Large \bfseries Breaking the Privacy Promise: Gradient Inversion Attacks on Federated Low-Rank Adapted Vision Transformers}
    \vspace{0.2cm}
    \hrule
    \vspace{0.4cm}
\end{strip}

\section{Dimension Estimation for Sample-Wise Gradients}
\label{app:gradient_dimension_estimators}

This section summarizes the estimators used to quantify the dimensionality of sample-wise gradients. For each private example $(x_i,y_i)$, we compute the observed per-sample gradient
\begin{equation}
    g_i = \operatorname{vec}\!\left(\nabla_{\Theta_{\mathrm{obs}}}\ell(f_\theta(x_i),y_i)\right) \in \mathbb{R}^{p},
\label{eqn:gradvec}
\end{equation}
where $\Theta_{\mathrm{obs}}$ denotes the parameters exposed to the server, such as all fine-tuned parameters, LoRA parameters, positional embeddings, or their union. The resulting gradient vectors $\mathcal{G}=\{g_i\}_{i=1}^{N}$ is analyzed after applying the same preprocessing across all compared settings. Unless otherwise specified, we flatten and concatenate the selected gradient tensors and centre the resulting vectors. %

\subsection{Nearest-neighbour intrinsic dimension}
Let $T_j(g_i)$ denote the distance from $g_i$ to its $j$-th nearest neighbour in $\mathcal{G}\setminus\{g_i\}$. The maximum-likelihood estimator (MLE) of intrinsic dimension uses local neighbour-distance ratios under a Poisson-process approximation \cite{levina2004mle}:
\begin{equation}
    \widehat{d}_{\mathrm{MLE}}(k)
    = \frac{1}{N}\sum_{i=1}^{N}
    \left[\frac{1}{k-1}\sum_{j=1}^{k-1}
    \log\frac{T_k(g_i)}{T_j(g_i)}\right]^{-1}.
\end{equation}
We report MLE for $k = 20$ as suggested in the work \cite{levina2004mle}, because a small $k$ emphasizes local manifold geometry, whereas a larger $k$ can mix different semantic or class-dependent regions of the gradient volume.

TwoNN estimates intrinsic dimension using only the first two nearest neighbours \cite{Facco2017twonn}. For each point, define $\mu_i=T_2(g_i)/T_1(g_i)$. Under local uniformity, the empirical cumulative distribution $F(\mu)$ satisfies $F(\mu)\approx 1-\mu^{-d}$. Thus, plotting $-\log(1-F(\mu_i))$ against $\log\mu_i$ yields an approximately linear relation whose slope estimates $d$. TwoNN is useful for sample-wise gradients because it uses the smallest possible neighbourhood and is therefore less sensitive to curvature and density changes at larger scales.

L2N2 \cite{ong2026l2n2} is also based on nearest-neighbour distance ratios, but aggregates log-log transformed ratio statistics instead of fitting the empirical cumulative distribution directly \cite{ong2026l2n2}. In practice, we compute statistics of the form
\begin{equation}
    L_{k,j}(g_i)=\log\!\left(\log\frac{T_k(g_i)}{T_j(g_i)}\right), \qquad k>j,
\end{equation}
and convert their empirical average into an intrinsic-dimension estimate using the L2N2 calibration for the chosen $(k,j)$ pair. This provides a complementary nearest-neighbour estimate whose assumptions differ from MLE and TwoNN, making it a useful robustness check when the gradient vectors are spread non-uniformly.

\subsection{Jacobian-based effective ranks}

\noindent
Nearest-neighbour estimators are bounded by the neighbourhood of the gradient vectors, as they are heavily dependent upon the value of $k$. We additionally measure the local dimension of the gradient map around each sample $x_i$ with label $y_i$. Let
\begin{equation}
    \phi(x_i,y_i)=\operatorname{vec}\!\left(\nabla_{\Theta_{\mathrm{obs}}}\ell(f_\theta(x_i),y_i)\right),
\end{equation}
\begin{equation}
    J_i = \frac{\partial \phi(x_i,y_i)}{\partial x_i}
    \in \mathbb{R}^{p\times d_x},
\end{equation}
where $d_x$ is the input dimension and the label is held fixed. The singular values of $J_i$, denoted $\sigma_{i,1}\geq\sigma_{i,2}\geq\cdots$, describe the independent input-space directions that induce observable changes in the shared gradient. We compute the entropy-based effective rank \cite{roy2007effective}
\begin{equation}
    r_{\mathrm{eff}}(J_i)=\exp\!\left(-\sum_m q_{i,m}\log q_{i,m}\right),
\end{equation}
where
\begin{equation}
    q_{i,m}=\frac{\sigma_{i,m}^{2}}{\sum_l \sigma_{i,l}^{2}},
\end{equation}
and the cumulative-energy ranks
\begin{equation}
    E_i(r)=\frac{\sum_{m=1}^{r}\sigma_{i,m}^{2}}{\sum_l\sigma_{i,l}^{2}},
\end{equation}

We also compute the ranks at different energy levels ($90\%, 95\%, 99\%$), by thresholding at
\begin{equation}
    r_{\tau}(J_i)=\min\{r:E_i(r)\geq\tau\},
    \qquad \tau\in\{0.90,0.95,0.99\}.
\end{equation}
and these thresholds are denoted as $r_{90}$, $r_{95}$, and $r_{99}$. Low values indicate that only a few input directions control most of the observable gradient variation, while larger values indicate a richer local gradient response. In LoRA-only settings, the projection onto a small trainable parameter subspace can collapse the spectrum of $J_i$; adding trainable positional embeddings can increase the number of spatially meaningful directions visible to the server.

\paragraph{Interpretation.}
The four families of measurements are complementary. MLE, TwoNN, and L2N2 estimate the non-linear intrinsic dimension of the empirical gradient vectors $\mathcal{G}$, while Jacobian effective rank and $r_{90}/r_{95}/r_{99}$ estimate local linear dimensionality around individual samples. For fair comparisons, all estimates should be computed with the same data subset, observed parameter set, distance metric, normalization, and random seed. We report the mean and standard deviation across bootstrap resamples for both types of measurements in Fig. \ref{fig:intdim}.

\subsection{Random Projection Preprocessing}
\label{app:gradient_random_projection}

The raw sample-wise gradient vectors are extremely high-dimensional: even for ViT-Small, full fine-tuning exposes roughly $21$ million gradient coordinates, while rank-$128$ LoRA exposes only about $3.5$ million coordinates. To make nearest-neighbour search and bootstrap evaluation tractable, we first project each centred gradient vector to $m=8192$ dimensions using the Achlioptas random projection scheme \cite{achlioptas2003database}, following the Johnson-Lindenstrauss (JL) lemma \cite{Johnson1984jllemma}. For a gradient vector $g_i\in\mathbb{R}^{p}$, we compute $\tilde{g}_i=Rg_i\in\mathbb{R}^{8192}$ with
\begin{equation}
    R_{ab}=\sqrt{\frac{3}{m}}
    \begin{cases}
    +1, & \text{with probability } 1/6,\\
    0, & \text{with probability } 2/3,\\
    -1, & \text{with probability } 1/6.
    \end{cases}
\end{equation}
The projected vectors $\tilde{\mathcal{G}}=\{\tilde{g}_i\}_{i=1}^{N}$ are then used as inputs to MLE, TwoNN, L2N2, and the Jacobian-based spectrum analyses. In the formulas below, we write $g_i$ and $\mathcal{G}$ for the projected vectors $\tilde{g}_i$ and $\tilde{\mathcal{G}}$ for notational simplicity.

For each bootstrapped run, we randomly select at most $N=20000$ sample-wise gradients. Using the standard JL sufficient condition
\begin{equation}
    m \geq \frac{4\log N}{\epsilon^2/2-\epsilon^3/3},
\end{equation}
choosing $\epsilon=0.102$ gives
\begin{equation}
    \frac{4\log(20000)}{0.102^2/2-0.102^3/3}
    \approx 8171 < 8192.
\end{equation}
Thus, $8192$ projected dimensions satisfy the JL bound for $20000$ sampled gradients, preserving pairwise distances up to approximately a $1\pm 0.102$ multiplicative distortion before applying the dimension estimators.

\section{Comparison of Performance of the Broadcasted Backbone}
\label{sec:backbonecomp}

In this section, we evaluate the proposed backbone across multiple benchmark datasets to assess its suitability with respect to our \textbf{trustworthiness criteria}. As discussed previously, \emph{accuracy} and \emph{robustness} constitute two fundamental pillars of trustworthy learning \cite{Tariq2024tflsurvey, SnchezSnchez2024fedtrust}. We therefore evaluate the proposed backbone along these dimensions before it is broadcast by the server to the participating clients. This evaluation ensures that the backbone provides meaningful utility to the clients while maintaining sufficient robustness for the intended learning task.

Importantly, this evaluation is also necessary to distinguish the proposed honest-but-curious setting from a malicious server. A server could deliberately assign a poorly performing backbone to the clients, thereby degrading their learning performance while still exploiting the resulting client updates to facilitate data reconstruction. Such behaviour provides little or no utility to the clients while benefiting the server's reconstruction objective, and can therefore be regarded as malicious. Consequently, before broadcasting the backbone to the clients, we establish that it satisfies the required accuracy and robustness criteria, ensuring that the server's objective of reconstructing client data does not come at the expense of the clients' learning utility.

\begin{table}[!ht]
\centering
\caption{Comparison of the performance of the proposed backbone used in our experiments.}
\label{tab:backbone_settings}
\resizebox{\linewidth}{!}{
\begin{tabular}{c | c | c c | c c |c | c | c}
\toprule
\multirow{2}{*}{\textbf{Backbone}} & 
\multirow{2}{*}{\textbf{LoRA}} &
\multicolumn{2}{c|}{\textbf{PE Present (Layers)}} &
\multicolumn{2}{c|}{\textbf{PE Trainable (Layers)}} &
\multirow{2}{*}{\shortstack{\textbf{Patch}\\\textbf{Size}}} &
\multirow{2}{*}{\shortstack{\textbf{Full Params}\\\textbf{Trainable}}} &
\multirow{2}{*}{\shortstack{\textbf{Accuracy (\%)}\\ $\mu(\sigma)$}}\\
& 
&
\textbf{$1^{\text{st}}$} &
\textbf{$2^{\text{nd}}$-$\mathrm{L}^{\text{th}}$} &
\textbf{$1^{\text{st}}$} &
\textbf{$2^{\text{nd}}$-$\mathrm{L}^{\text{th}}$} &
&
&
\\
\midrule
\multicolumn{9}{c}{Trained From Scratch}\\\midrule
\multicolumn{9}{c}{\textbf{CIFAR-10}}\\\midrule
ViT & No  & Yes & No  & Yes & N/A & 4 & Yes & 95.6 (0.1) \\
Ours & No  & Yes & Yes & Yes & Yes & 4 & Yes & \textbf{95.7} (0.2) \\ \midrule
\multicolumn{9}{c}{Finetuned on ImageNet Pre-trained weights}\\\midrule
\multicolumn{9}{c}{\textbf{CIFAR-10}}\\\midrule
ViT & Yes & Yes & No  & No  & N/A & 4 & No  & 97.7 (0.1) \\
Ours & Yes & Yes & Yes & Yes & Yes & 4 & No  & \textbf{97.8} (0.1) \\
Ours & Yes & Yes & Yes & No  & No  & 4 & No  & 97.7 (0.1) \\  \midrule
\multicolumn{9}{c}{\textbf{Stanford CARS}}\\\midrule
\multicolumn{9}{c}{Finetuned on ImageNet pre-trained weights}\\\midrule
ViT & Yes & Yes & No  & No  & N/A & 4 & No  & 84.4 (0.3) \\
Ours & Yes & Yes & Yes & Yes & Yes & 4 & No  & \textbf{85.0} (0.4) \\
Ours & Yes & Yes & Yes & No  & No  & 4 & No  & 81.5 (0.9) \\ 
\bottomrule
\end{tabular}
}
\label{tab:backboneres}
\end{table}
In Tab.~\ref{tab:backboneres}, we observe that the proposed backbone, when equipped with trainable positional embeddings, outperforms the baseline ViT. This observation is consistent with the findings of LaPE \cite{lape} and MPVG \cite{LeeHK25mpvg}, which also demonstrate the benefits of adapting positional embeddings during training. These results justify the use of the proposed architecture, illustrated in Fig.~\ref{fig:archifigure}, within our FL training protocol. Importantly, by establishing that the backbone provides competitive or improved performance to the clients, we ensure that the server's use of this architecture does not constitute malicious behaviour by deliberately distributing a poor-performing model. Instead, the proposed protocol benefits the participating clients while supporting a trustworthy server-client interaction.

\section{Group Lazy Regularization}
\label{sec:glr}

Gradient inversion is a highly non-convex optimization problem, and reconstructions obtained from different random initializations often converge to distinct local optima despite corresponding to the same underlying image. Consequently, different optimization trajectories typically preserve the same semantic content while exhibiting slight spatial variations. Group lazy regularization \cite{yin2021gradinversion} introduces a consensus reconstruction that serves as a common optimization target for all trajectories.

Given a set of reconstructed images $\{\hat{x}_g\}_{g \in G}$, the consensus image is computed as the pixel-wise average

\begin{equation}
\mathbb{E}(\hat{x}_{g \in G})
=
\frac{1}{|G|}
\sum_{g \in G}
\hat{x}_g.
\end{equation}

The group consistency regularizer is defined as

\begin{equation}
\mathcal{R}_{\mathrm{group-lazy}}
=
\lambda_{G}
\left\|
\hat{x}
-
\mathbb{E}(\hat{x}_{g \in G})
\right\|_2,
\end{equation}

where $\lambda_{G}$ controls the regularization strength. For our work, we observed that using $\lambda_G = 1e-3$ yielded the best results. The consensus image acts as a soft spatial prior, guiding each reconstruction toward a semantically consistent solution and improving optimization stability. %

\subsection{Impact of Group Lazy Regularization}
\label{sec:glrabl}

\textbf{Effect of Group Lazy Regularization.} In Tab.~\ref{tab:glr_ablation}, we assess the effects of turning on Group Lazy Regularization (GLR) with a coefficient $\lambda_{G}$ of $10^{-4}$. We observe that enabling GLR marginally enhances reconstruction quality for both the baseline GIAS and GIAS paired with our trainable positional embeddings. These findings imply that group lazy regularization offers benefits in the baseline environment as well as with trainable positional embeddings, although the improvements are not statistically significant (Welch's t-test with $N=30$). Nevertheless, we retain GLR in our framework because it serves as a regularizer that encourages coherent updates within each parameter group, leading to a more stable optimization process. Since it incurs negligible computational overhead and never degrades reconstruction quality in our experiments, we include it in all subsequent evaluations.

\begin{table}[!ht]
\centering
\caption{Effect of enabling group lazy regularization ($\lambda_G=10^{-4}$). Parentheses denote standard deviation. Welch's $t$-test ($N=30$) compares enabling group lazy regularization for both methods.}
\label{tab:glr_ablation}
\small
\setlength{\tabcolsep}{3pt}
\begin{tabular}{llccc}
\toprule
Method & GLR &
PSNR $\uparrow$ &
SSIM $\uparrow$ &
LPIPS $\downarrow$ \\
\midrule

\multirow{3}{*}{GIAS}
& False &
\makecell{15.22 \\ \scriptsize{(4.78)}} &
\makecell{0.313 \\ \scriptsize{(0.226)}} &
\makecell{0.094 \\ \scriptsize{(0.056)}} \\

& True &
\makecell{15.77 \\ \scriptsize{(4.52)}} &
\makecell{0.340 \\ \scriptsize{(0.221)}} &
\makecell{0.120 \\ \scriptsize{(0.075)}} \\ \midrule

\multicolumn{2}{c}{$p$-value} &
\cellcolor{red!20}0.649 &
\cellcolor{red!20}0.642 &
\cellcolor{red!20}0.134 \\

\midrule

\multirow{3}{*}{GIAS + Ours}
& False &
\makecell{25.35 \\ \scriptsize{(8.00)}} &
\makecell{0.784 \\ \scriptsize{(0.235)}} &
\makecell{0.021 \\ \scriptsize{(0.033)}} \\

& True &
\makecell{25.91 \\ \scriptsize{(6.29)}} &
\makecell{0.790 \\ \scriptsize{(0.170)}} &
\makecell{0.022 \\ \scriptsize{(0.033)}} \\ \midrule

\multicolumn{2}{c}{$p$-value} &
\cellcolor{red!20}0.764 &
\cellcolor{red!20}0.910 &
\cellcolor{red!20}0.907 \\

\bottomrule
\end{tabular}
\end{table}

\begin{figure}
    \centering
    \begin{tabular}{m{2cm}m{1cm}ccc}
\toprule
Method & GLR &
1 &
2 &
3 \\
\midrule
\multicolumn{2}{c}{Ground Truth} & 
\includegraphics[width=0.1\linewidth]{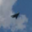} & \includegraphics[width=0.1\linewidth]{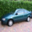} & \includegraphics[width=0.1\linewidth]{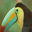} \\ \midrule
\multirow{2}{*}{GIAS}
& False &
\includegraphics[width=0.1\linewidth]{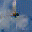} & \includegraphics[width=0.1\linewidth]{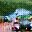} & \includegraphics[width=0.1\linewidth]{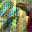} \\ \cmidrule{2-5}

& True &
\includegraphics[width=0.1\linewidth]{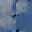} & \includegraphics[width=0.1\linewidth]{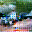} & \includegraphics[width=0.1\linewidth]{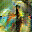} \\ \midrule

\midrule

\multirow{2}{*}{GIAS + Ours}
& False &
\includegraphics[width=0.1\linewidth]{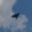} & \includegraphics[width=0.1\linewidth]{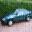} & \includegraphics[width=0.1\linewidth]{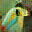} \\ \cmidrule{2-5}

& True &
\includegraphics[width=0.1\linewidth]{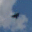} & \includegraphics[width=0.1\linewidth]{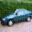} & \includegraphics[width=0.1\linewidth]{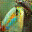} \\ \midrule

\bottomrule
\end{tabular}
    \caption{Visual comparison of reconstructions with and without group lazy regularization (GLR) using $\lambda_G=10^{-4}$. Results are shown for both the baseline GIAS and GIAS+Ours. Enabling GLR produces comparable reconstruction quality for the baseline while yielding slightly improved visual fidelity for GIAS+Ours, consistent with the quantitative results in Tab.~\ref{tab:glr_ablation}.}
    \label{fig:glr_ablation}
\end{figure}

\noindent
\textbf{Effect of the $\lambda_G$ value.} Tab.~\ref{tab:glr_value_ablation} investigates the effect of the group lazy regularization coefficient $\lambda_G$. Increasing $\lambda_G$ from $10^{-4}$ to $10^{-3}$ improves the average reconstruction quality of our method, raising PSNR from 25.91 to 27.64 dB, SSIM from 0.79 to 0.85, and reducing LPIPS from 0.022 to 0.011. This trend suggests that our optimization can benefit from a stronger group lazy regularization term. However, Welch's $t$-test with $N=30$ yields $p_{\text{PSNR}}=0.292$, $p_{\text{SSIM}}=0.168$, and $p_{\text{LPIPS}}=0.108$, indicating that these improvements are not statistically significant ($p>0.05$ for all metrics). Consequently, although $\lambda_G=10^{-3}$ achieves slightly better mean performance for our method, we adopt $\lambda_G=10^{-4}$ in all subsequent experiments because it yields the best performance for the baseline GIAS. Using the same regularization coefficient for both methods ensures identical optimization settings and enables a fair comparison between the proposed method and the baseline.

\begin{table}[!ht]
\centering
\caption{Effect of the group lazy-regularization coefficient value when GLR optimization is enabled. Parentheses denote standard deviation. Welch's $t$-test ($N=30$) compares the two $\lambda_G$ values.}
\label{tab:glr_value_ablation}
\small
\setlength{\tabcolsep}{3pt}
\begin{tabular}{llccc}
\toprule
Method & $\lambda_G$ &
PSNR $\uparrow$ &
SSIM $\uparrow$ &
LPIPS $\downarrow$ \\
\midrule

\multirow{3}{*}{GIAS}
& $10^{-4}$ &
\makecell{15.77 \\ \scriptsize{(4.52)}} &
\makecell{0.340 \\ \scriptsize{(0.221)}} &
\makecell{0.120 \\ \scriptsize{(0.075)}} \\

& $10^{-3}$ &
\makecell{14.01 \\ \scriptsize{(3.29)}} &
\makecell{0.268 \\ \scriptsize{(0.164)}} &
\makecell{0.102 \\ \scriptsize{(0.052)}} \\ \midrule

\multicolumn{2}{c}{$p$-value} &
\cellcolor{red!20}0.090 &
\cellcolor{red!20}0.158 &
\cellcolor{red!20}0.285 \\

\midrule

\multirow{3}{*}{GIAS + Ours}
& $10^{-4}$ &
\makecell{25.91 \\ \scriptsize{(6.29)}} &
\makecell{0.790 \\ \scriptsize{(0.170)}} &
\makecell{0.022 \\ \scriptsize{(0.033)}} \\

& $10^{-3}$ &
\makecell{27.64 \\ \scriptsize{(6.31)}} &
\makecell{0.850 \\ \scriptsize{(0.163)}} &
\makecell{0.011 \\ \scriptsize{(0.016)}} \\ \midrule

\multicolumn{2}{c}{$p$-value} &
\cellcolor{red!20}0.292 &
\cellcolor{red!20}0.168 &
\cellcolor{red!20}0.108 \\

\bottomrule
\end{tabular}
\end{table}

\begin{figure}
    \centering
    \begin{tabular}{m{2cm}m{1cm}ccc}
\toprule
Method & $\lambda_G$ &
1 &
2 &
3 \\
\midrule
\multicolumn{2}{c}{Ground Truth} & 
\includegraphics[width=0.1\linewidth]{images/appendix/group_lazy/GT/gt_0000_label0.png} & \includegraphics[width=0.1\linewidth]{images/appendix/group_lazy/GT/gt_0000_label1.png} & \includegraphics[width=0.1\linewidth]{images/appendix/group_lazy/GT/gt_0000_label2.png} \\ \midrule
\multirow{2}{*}{GIAS}
& $10^{-4}$ &
\includegraphics[width=0.1\linewidth]{images/appendix/group_lazy/True/gias/rec_0000_label0.png} & \includegraphics[width=0.1\linewidth]{images/appendix/group_lazy/True/gias/rec_0000_label1.png} & \includegraphics[width=0.1\linewidth]{images/appendix/group_lazy/True/gias/rec_0000_label2.png} \\ \cmidrule{2-5}

& $10^{-3}$ &
\includegraphics[width=0.1\linewidth]{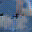} & \includegraphics[width=0.1\linewidth]{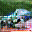} & \includegraphics[width=0.1\linewidth]{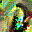} \\ \midrule

\midrule

\multirow{2}{*}{GIAS + Ours}
& $10^{-4}$ &
\includegraphics[width=0.1\linewidth]{images/appendix/group_lazy/True/giasours/rec_0000_label0.png} & \includegraphics[width=0.1\linewidth]{images/appendix/group_lazy/True/giasours/rec_0000_label1.png} & \includegraphics[width=0.1\linewidth]{images/appendix/group_lazy/True/giasours/rec_0000_label2.png} \\ \cmidrule{2-5}

& $10^{-3}$ &
\includegraphics[width=0.1\linewidth]{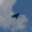} & \includegraphics[width=0.1\linewidth]{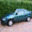} & \includegraphics[width=0.1\linewidth]{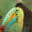} \\ \midrule

\bottomrule
\end{tabular}
    \caption{Visual comparison of reconstructions obtained using different group lazy regularization coefficients, $\lambda_G\in\{10^{-4},10^{-3}\}$. While a larger coefficient ($10^{-3}$) produces slightly sharper reconstructions for GIAS+Ours, the qualitative differences are modest. We therefore adopt $\lambda_G=10^{-4}$ in the remaining experiments to maintain identical optimization settings with the baseline GIAS.}
    \label{fig:lambda_g_abl}
\end{figure}

\section{Experimental Details for CelebA and CARS Datasets}
\label{sec:expxtra}

For the CelebA \cite{liu2015celebA} and Stanford CARS \cite{Krause2013cars} datasets, we use StyleGAN2 \cite{karras2020stylegan2} models pretrained on the CelebA-HQ \cite{karras2017progressive, karras2026celebAhq} and LSUN Cars \cite{YuZSSX15LSUN} datasets, respectively. For the CelebA experiments, we randomly select 20 celebrities to construct a 20-class classification task, whereas the Stanford CARS dataset comprises 196 unique classes. While the face generator is pretrained on the CelebA-HQ domain, the generator used for Stanford CARS is pretrained on LSUN Cars. Consequently, the CARS dataset setting exhibits a larger difference between the distribution of the generator training data and that of the client data. For our experiments, we keep the image resolution at $128\times128$ for the CelebA dataset and $64\times64$ for the Stanford CARS dataset. The optimization steps for the latent vector $z$ and the generator parameters were run for 2000 and 8000 iterations, respectively.

\section{Qualitative Evidence on Applying Trainable Positional Embeddings to Optimization-Based GIA}
\label{sec:tpeoptgia}

In this section, we present qualitative evidence on the effect of applying trainable positional embeddings to optimization-based gradient inversion attacks (GIAs), such as GradViT \cite{ali2022gradvit}. Since positional embeddings provide spatial information that is essential for correctly aligning the reconstructed image patches, we investigate whether making them trainable can also improve the reconstruction performance of conventional optimization-based GIA methods. 

\begin{figure}[!ht]
\centering
\scriptsize
\setlength{\tabcolsep}{1pt}

\begin{tabular}{c cccccccc}
\toprule
\textbf{Method}
& \textbf{1}
& \textbf{2}
& \textbf{3}
& \textbf{4}
& \textbf{5}
& \textbf{6}
& \textbf{7}
& \textbf{8}\\
\midrule

GT
& \includegraphics[width=0.09\linewidth]{images/fig5_gt/CIFAR10_vit_small_user0_gt_gt_0000_label4.png}
& \includegraphics[width=0.09\linewidth]{images/fig5_gt/CIFAR10_vit_small_user1_gt_gt_0000_label0.png}
& \includegraphics[width=0.09\linewidth]{images/fig5_gt/CIFAR10_vit_small_user2_gt_gt_0000_label2.png}
& \includegraphics[width=0.09\linewidth]{images/fig5_gt/CIFAR10_vit_small_user3_gt_gt_0000_label8.png}
& \includegraphics[width=0.09\linewidth]{images/fig5_gt/CIFAR10_vit_small_user4_gt_gt_0000_label7.png}
& \includegraphics[width=0.09\linewidth]{images/fig5_gt/CIFAR10_vit_small_user5_gt_gt_0000_label4.png}
& \includegraphics[width=0.09\linewidth]{images/fig5_gt/CIFAR10_vit_small_user6_gt_gt_0000_label0.png}
& \includegraphics[width=0.09\linewidth]{images/fig5_gt/CIFAR10_vit_small_user7_gt_gt_0000_label7.png}
\\

GradViT
& \includegraphics[width=0.09\linewidth]{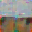}
& \includegraphics[width=0.09\linewidth]{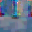}
& \includegraphics[width=0.09\linewidth]{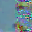}
& \includegraphics[width=0.09\linewidth]{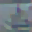}
& \includegraphics[width=0.09\linewidth]{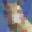}
& \includegraphics[width=0.09\linewidth]{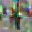}
& \includegraphics[width=0.09\linewidth]{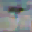}
& \includegraphics[width=0.09\linewidth]{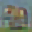}
\\

GradViT + Ours
& \includegraphics[width=0.09\linewidth]{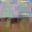}
& \includegraphics[width=0.09\linewidth]{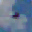}
& \includegraphics[width=0.09\linewidth]{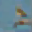}
& \includegraphics[width=0.09\linewidth]{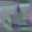}
& \includegraphics[width=0.09\linewidth]{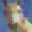}
& \includegraphics[width=0.09\linewidth]{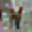}
& \includegraphics[width=0.09\linewidth]{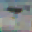}
& \includegraphics[width=0.09\linewidth]{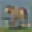}
\\
\bottomrule
\end{tabular}
\caption{Qualitative reconstruction for GradViT \cite{ali2022gradvit} comparison across 8 samples on the CIFAR-10 dataset. We can observe that the proposed framework also improves the spatial information obscuration issue for optimization-based gradient inversion attacks too, on the backbone ViT-Small model.}
\label{fig:gradvitqualgrid}
\end{figure}

We present qualitative results in Fig.~\ref{fig:gradvitqualgrid} to analyze the effect of trainable positional embeddings on the clarity and spatial consistency of the reconstructed samples. We observe that the proposed method can correct the patch misalignment that occurs in GradViT when the client model is fine-tuned using low-rank adapters. This further demonstrates the capability of our method to address the duplicate data issue caused by missing spatial information in the gradients, extending its applicability to optimization-based GIAs as well.

\section{More Qualitative Results on Larger Batch Sizes}
\label{sec:largebsmore}

\begin{figure}[!ht]
\centering
\scriptsize
\setlength{\tabcolsep}{1pt}

\begin{tabular}{c| cccccccc}
\toprule
\textbf{Method} &
\multicolumn{4}{c|}{4 Samples from a batch of 4}& \multicolumn{4}{c}{4 samples from a batch of 16} \\ 
\midrule

GT
& \includegraphics[width=0.09\linewidth]{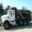}
& \includegraphics[width=0.09\linewidth]{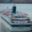}
& \includegraphics[width=0.09\linewidth]{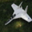}
& \includegraphics[width=0.09\linewidth]{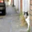}
& \includegraphics[width=0.09\linewidth]{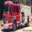}
& \includegraphics[width=0.09\linewidth]{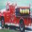}
& \includegraphics[width=0.09\linewidth]{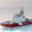}
& \includegraphics[width=0.09\linewidth]{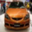}
\\

GIAS
& \includegraphics[width=0.09\linewidth]{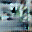}
& \includegraphics[width=0.09\linewidth]{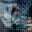}
& \includegraphics[width=0.09\linewidth]{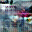}
& \includegraphics[width=0.09\linewidth]{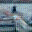}
& \includegraphics[width=0.09\linewidth]{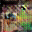}
& \includegraphics[width=0.09\linewidth]{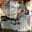}
& \includegraphics[width=0.09\linewidth]{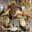}
& \includegraphics[width=0.09\linewidth]{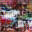}
\\

GIAS + Ours
& \includegraphics[width=0.09\linewidth]{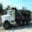}
& \includegraphics[width=0.09\linewidth]{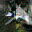}
& \includegraphics[width=0.09\linewidth]{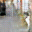}
& \includegraphics[width=0.09\linewidth]{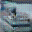}
& \includegraphics[width=0.09\linewidth]{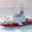}
& \includegraphics[width=0.09\linewidth]{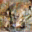}
& \includegraphics[width=0.09\linewidth]{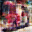}
& \includegraphics[width=0.09\linewidth]{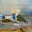}
\\

\bottomrule
\end{tabular}

\caption{Qualitative reconstruction comparison for GIAS and GIAS+Ours for batch sizes 4 and 16 in the single-step gradient setting with rank 128.}
\label{fig:morelargebatch}
\end{figure}

In this section, we present more results for qualitative comparison of GIAS and the proposed method for batch sizes of 4 and 16 in Fig. \ref{fig:morelargebatch} in the single-step gradient setting with LoRA rank $r=128$.

\section{More Qualitative Results on High-Resolution ($128 \times 128$) CelebA}
\label{sec:moreceleba}

\begin{figure}[!ht]
\centering
\scriptsize
\setlength{\tabcolsep}{1pt}

\begin{tabular}{m{1.5cm} cccccc}
\toprule
\textbf{Method}
& \multicolumn{6}{c}{CelebA}\\
\midrule

GT
& \includegraphics[width=0.12\linewidth]{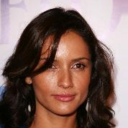}
& \includegraphics[width=0.12\linewidth]{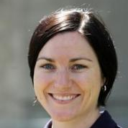}
& \includegraphics[width=0.12\linewidth]{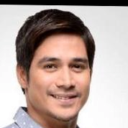}
& \includegraphics[width=0.12\linewidth]{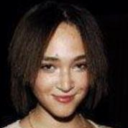}
& \includegraphics[width=0.12\linewidth]{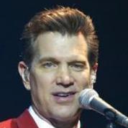}
& \includegraphics[width=0.12\linewidth]{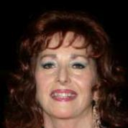}
\\

GIAS
& \includegraphics[width=0.12\linewidth]{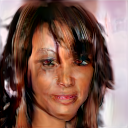}
& \includegraphics[width=0.12\linewidth]{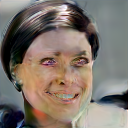}
& \includegraphics[width=0.12\linewidth]{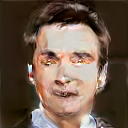}
& \includegraphics[width=0.12\linewidth]{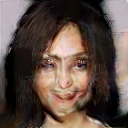}
& \includegraphics[width=0.12\linewidth]{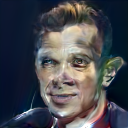}
& \includegraphics[width=0.12\linewidth]{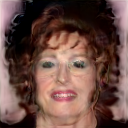}
\\

GIAS + Ours
& \includegraphics[width=0.12\linewidth]{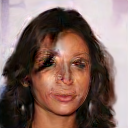}
& \includegraphics[width=0.12\linewidth]{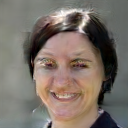}
& \includegraphics[width=0.12\linewidth]{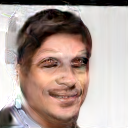}
& \includegraphics[width=0.12\linewidth]{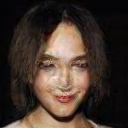}
& \includegraphics[width=0.12\linewidth]{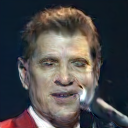}
& \includegraphics[width=0.12\linewidth]{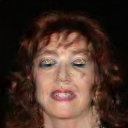}
\\\midrule

GT
& \includegraphics[width=0.12\linewidth]{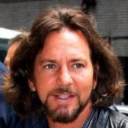}
& \includegraphics[width=0.12\linewidth]{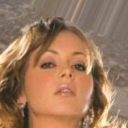}
& \includegraphics[width=0.12\linewidth]{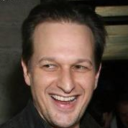}
& \includegraphics[width=0.12\linewidth]{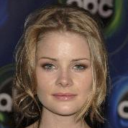}
& \includegraphics[width=0.12\linewidth]{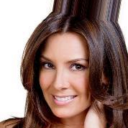}
& \includegraphics[width=0.12\linewidth]{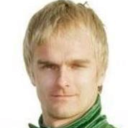}
\\

GIAS
& \includegraphics[width=0.12\linewidth]{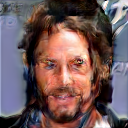}
& \includegraphics[width=0.12\linewidth]{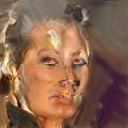}
& \includegraphics[width=0.12\linewidth]{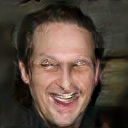}
& \includegraphics[width=0.12\linewidth]{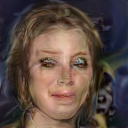}
& \includegraphics[width=0.12\linewidth]{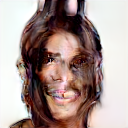}
& \includegraphics[width=0.12\linewidth]{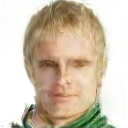}
\\

GIAS + Ours
& \includegraphics[width=0.12\linewidth]{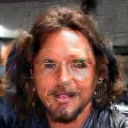}
& \includegraphics[width=0.12\linewidth]{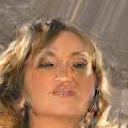}
& \includegraphics[width=0.12\linewidth]{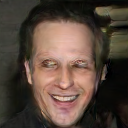}
& \includegraphics[width=0.12\linewidth]{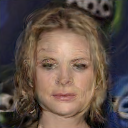}
& \includegraphics[width=0.12\linewidth]{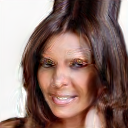}
& \includegraphics[width=0.12\linewidth]{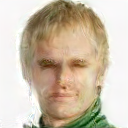}
\\ \midrule
\bottomrule
\end{tabular}

\caption{Qualitative comparison on samples from the CelebA dataset with image resolution $128 \times 128$.}
\label{fig:morecelebAimgs}
\end{figure}

In this section, we present more reconstruction results on high-resolution images from the CelebA dataset in Fig. \ref{fig:morecelebAimgs}.

\section{Empirical Analysis of Effect of Difference in Distribution on Reconstruction Quality}
\label{sec:distdist}

In this section, we evaluate whether a difference in distribution between the pre-training data of the generator (say, $p_{gen}$) and the local data of the victim client (say, $p_{cl}$) has any effect on the attack performance of the proposed method. For this empirical analysis, we simulate the attack on the CIFAR-100 dataset with DCGAN pretrained on CIFAR-10, and also on the Stanford CARS dataset with StyleGAN pretrained on the LSUN Cars dataset. As we have mentioned previously in the main manuscript, the dataset on which the GAN is pre-trained and the data of the client are similar but do not have overlapping data; we consider this to be an adequate setup to evaluate the aforementioned idea.

\subsection{Quantitative Comparison}

Although we have already provided empirical results in Sec. \ref{subsubsec:fedsgdattack} for the Stanford CARS dataset, we will reproduce the same here again for the continuation of the flow of the manuscript. In Tab. \ref{tab:c100cars}, we observe from the PSNR, SSIM and LPIPS metric values that, compared to the baseline GIAS, the proposed method is capable of bridging the gap between the two data distributions $p_{gen}$ and $p_{cl}$, consequently improving attack performance. 

\begin{table}[!ht]
    \centering
    \scriptsize
    \caption{Quantitative comparison on CIFAR-100 and CARS datasets.}
    \label{tab:c100cars}
    \vspace{-1em}
    \begin{tabular}{c|c|c|c|c|c|c}
    \toprule
     & \multicolumn{3}{c|}{\textbf{CARS}} & \multicolumn{3}{c}{\textbf{CIFAR-100}} \\ \midrule
    Method & PSNR & SSIM & LPIPS & PSNR & SSIM & LPIPS\\\midrule
     GIAS   & 
     \makecell{9.04 \\ \tiny{(1.90)}} &
     \makecell{0.07 \\ \tiny{(0.08)}} &
     \makecell{0.357 \\ \tiny{(0.074)}}&
     \makecell{13.90 \\ \tiny{(4.64)}}& 
     \makecell{0.307 \\ \tiny{(0.238)}}&
     \makecell{0.099 \\ \tiny{(0.073)}}\\ \midrule
     GIAS+Ours  & 
     \makecell{\textbf{20.783} \\ \tiny{(3.166)}} &
     \makecell{\textbf{0.72} \\ \tiny{(0.128)}} &
     \makecell{\textbf{0.089} \\ \tiny{(0.058)}} &
     \makecell{\textbf{24.16} \\ \tiny{(5.293)}}&
     \makecell{\textbf{0.794} \\ \tiny{(0.201)}} &
     \makecell{\textbf{0.024} \\ \tiny{(0.041)}}\\
     \bottomrule
    \end{tabular}
\end{table}

\subsection{Qualitative Comparison}

\begin{figure}[!ht]
\centering
\scriptsize
\setlength{\tabcolsep}{1pt}

\begin{tabular}{m{1.5cm} cccccc}
\toprule
\textbf{Method}
& \multicolumn{6}{c}{CARS}\\
\midrule

GT
& \includegraphics[width=0.12\linewidth]{images/cars_imgs/0_GT.png}
& \includegraphics[width=0.12\linewidth]{images/cars_imgs/3_GT.png}
& \includegraphics[width=0.12\linewidth]{images/cars_imgs/4_GT.png}
& \includegraphics[width=0.12\linewidth]{images/cars_imgs/5_GT.png}
& \includegraphics[width=0.12\linewidth]{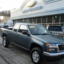}
& \includegraphics[width=0.12\linewidth]{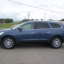}
\\

GIAS
& \includegraphics[width=0.12\linewidth]{images/cars_imgs/0_NPE.png}
& \includegraphics[width=0.12\linewidth]{images/cars_imgs/3_NPE.png}
& \includegraphics[width=0.12\linewidth]{images/cars_imgs/4_NPE.png}
& \includegraphics[width=0.12\linewidth]{images/cars_imgs/5_NPE.png}
& \includegraphics[width=0.12\linewidth]{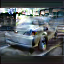}
& \includegraphics[width=0.12\linewidth]{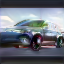}
\\

GIAS + Ours
& \includegraphics[width=0.12\linewidth]{images/cars_imgs/0_PE.png}
& \includegraphics[width=0.12\linewidth]{images/cars_imgs/3_PE.png}
& \includegraphics[width=0.12\linewidth]{images/cars_imgs/4_PE.png}
& \includegraphics[width=0.12\linewidth]{images/cars_imgs/5_PE.png}
& \includegraphics[width=0.12\linewidth]{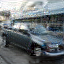}
& \includegraphics[width=0.12\linewidth]{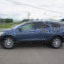}
\\\midrule

GT
& \includegraphics[width=0.12\linewidth]{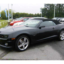}
& \includegraphics[width=0.12\linewidth]{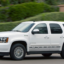}
& \includegraphics[width=0.12\linewidth]{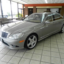}
& \includegraphics[width=0.12\linewidth]{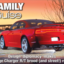}
& \includegraphics[width=0.12\linewidth]{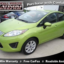}
& \includegraphics[width=0.12\linewidth]{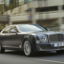}
\\

GIAS
& \includegraphics[width=0.12\linewidth]{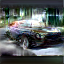}
& \includegraphics[width=0.12\linewidth]{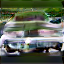}
& \includegraphics[width=0.12\linewidth]{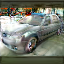}
& \includegraphics[width=0.12\linewidth]{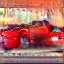}
& \includegraphics[width=0.12\linewidth]{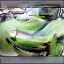}
& \includegraphics[width=0.12\linewidth]{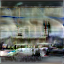}
\\

GIAS + Ours
& \includegraphics[width=0.12\linewidth]{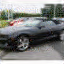}
& \includegraphics[width=0.12\linewidth]{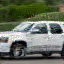}
& \includegraphics[width=0.12\linewidth]{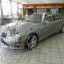}
& \includegraphics[width=0.12\linewidth]{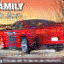}
& \includegraphics[width=0.12\linewidth]{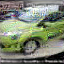}
& \includegraphics[width=0.12\linewidth]{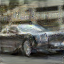}
\\ \midrule
\multicolumn{7}{c}{Images at larger scale}\\\midrule
GT
& \multicolumn{2}{c}{\includegraphics[width=0.24\linewidth]{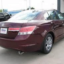}}
& \multicolumn{2}{c}{\includegraphics[width=0.24\linewidth]{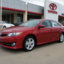}}
& \multicolumn{2}{c}{\includegraphics[width=0.24\linewidth]{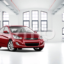}}
\\

GIAS
& \multicolumn{2}{c}{\includegraphics[width=0.24\linewidth]{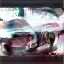}}
& \multicolumn{2}{c}{\includegraphics[width=0.24\linewidth]{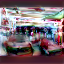}}
& \multicolumn{2}{c}{\includegraphics[width=0.24\linewidth]{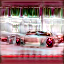}}
\\

GIAS + Ours
& \multicolumn{2}{c}{\includegraphics[width=0.24\linewidth]{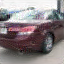}}
& \multicolumn{2}{c}{\includegraphics[width=0.24\linewidth]{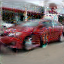}}
& \multicolumn{2}{c}{\includegraphics[width=0.24\linewidth]{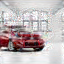}}
\\

\bottomrule
\end{tabular}

\caption{Qualitative comparison on the Stanford CARS datasets.}
\label{fig:carsimgs}
\end{figure}

\begin{figure}[!ht]
\centering
\scriptsize
\setlength{\tabcolsep}{1pt}

\begin{tabular}{c cccccccc}
\toprule
\textbf{Method}
& \multicolumn{8}{c}{CIFAR-100}\\
\midrule

GT
& \includegraphics[width=0.09\linewidth]{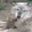}
& \includegraphics[width=0.09\linewidth]{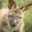}
& \includegraphics[width=0.09\linewidth]{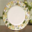}
& \includegraphics[width=0.09\linewidth]{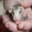}
& \includegraphics[width=0.09\linewidth]{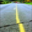}
& \includegraphics[width=0.09\linewidth]{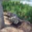}
& \includegraphics[width=0.09\linewidth]{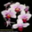}
& \includegraphics[width=0.09\linewidth]{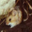}
\\

GIAS
& \includegraphics[width=0.09\linewidth]{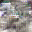}
& \includegraphics[width=0.09\linewidth]{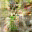}
& \includegraphics[width=0.09\linewidth]{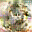}
& \includegraphics[width=0.09\linewidth]{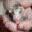}
& \includegraphics[width=0.09\linewidth]{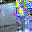}
& \includegraphics[width=0.09\linewidth]{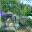}
& \includegraphics[width=0.09\linewidth]{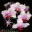}
& \includegraphics[width=0.09\linewidth]{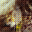}
\\

GIAS + Ours
& \includegraphics[width=0.09\linewidth]{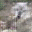}
& \includegraphics[width=0.09\linewidth]{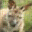}
& \includegraphics[width=0.09\linewidth]{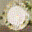}
& \includegraphics[width=0.09\linewidth]{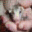}
& \includegraphics[width=0.09\linewidth]{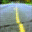}
& \includegraphics[width=0.09\linewidth]{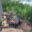}
& \includegraphics[width=0.09\linewidth]{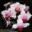}
& \includegraphics[width=0.09\linewidth]{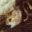}
\\ \midrule

GT
& \includegraphics[width=0.09\linewidth]{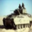}
& \includegraphics[width=0.09\linewidth]{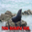}
& \includegraphics[width=0.09\linewidth]{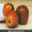}
& \includegraphics[width=0.09\linewidth]{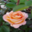}
& \includegraphics[width=0.09\linewidth]{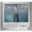}
& \includegraphics[width=0.09\linewidth]{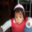}
& \includegraphics[width=0.09\linewidth]{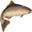}
& \includegraphics[width=0.09\linewidth]{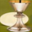}
\\

GIAS
& \includegraphics[width=0.09\linewidth]{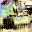}
& \includegraphics[width=0.09\linewidth]{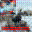}
& \includegraphics[width=0.09\linewidth]{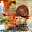}
& \includegraphics[width=0.09\linewidth]{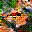}
& \includegraphics[width=0.09\linewidth]{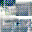}
& \includegraphics[width=0.09\linewidth]{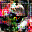}
& \includegraphics[width=0.09\linewidth]{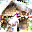}
& \includegraphics[width=0.09\linewidth]{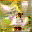}
\\

GIAS+Ours
& \includegraphics[width=0.09\linewidth]{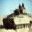}
& \includegraphics[width=0.09\linewidth]{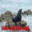}
& \includegraphics[width=0.09\linewidth]{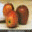}
& \includegraphics[width=0.09\linewidth]{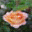}
& \includegraphics[width=0.09\linewidth]{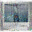}
& \includegraphics[width=0.09\linewidth]{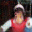}
& \includegraphics[width=0.09\linewidth]{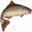}
& \includegraphics[width=0.09\linewidth]{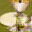}
\\

\bottomrule
\end{tabular}

\caption{Qualitative comparison on the CIFAR-100 datasets.}
\label{fig:c100imgs}
\end{figure}

We further present the qualitative results in Fig.~\ref{fig:carsimgs} and Fig.~\ref{fig:c100imgs}, illustrating the reconstruction quality on the Stanford CARS and CIFAR-100 datasets, respectively. We observe that, although the proposed attack is capable of reconstructing several samples from the Stanford CARS dataset with recognizable visual content, it can produce somewhat noisy reconstructions in a few cases. Nevertheless, these results demonstrate the out-of-distribution reconstruction capability of the proposed method. In contrast, for the CIFAR-100 dataset, the proposed attack can effectively reconstruct almost all samples with high visual fidelity. These qualitative results further highlight the effect of distributional differences on the attack performance, suggesting that greater similarity between the underlying data distributions facilitates more effective reconstruction.

\noindent
\textbf{Discussion:} Stanford CARS contains 196 classes, whereas LSUN Cars is treated as a single-class dataset, despite both datasets containing car images. Consequently, the two datasets exhibit a substantial distributional difference due to their distinct data sources, visual content characteristics, and semantic diversity. In contrast, CIFAR-10 and CIFAR-100 contain 10 and 100 classes, respectively, but share a more closely related underlying image distribution. Although the proposed attack outperforms the baseline on the Stanford CARS datasets, the performance improvement is considerably more pronounced for the CIFAR datasets. This observation provides empirical evidence that the distributional relationship between the target data and the data used by the attack can substantially influence reconstruction performance, with greater distributional similarity facilitating a larger improvement over the baseline.

Furthermore, the car datasets contain images with substantially higher spatial resolution than the CIFAR datasets, whereas in our experiments we use a ViT-Small backbone. Using a larger backbone may improve the attack performance further by increasing the number of trainable parameters and, consequently, the amount of gradient information available to the attacker, as discussed previously in Sec.~1.

\section{Statistical Guarantees in the FedAvg Setting}
\label{sec:statguarfedavg}

In the multi-step gradient update setting, each reported result in Tab. \ref{tab:steps_results} is averaged over nine independent runs with different random initializations. Statistical significance is assessed using a two-sided Welch's t-test ($\alpha=0.05$), which does not assume equal variances between the compared methods.

\begin{table}[!ht]
\centering
\scriptsize
\caption{Statistical significance analysis of the improvements reported in Tab.~\ref{tab:steps_results}. P-values are computed using a two-sided Welch's t-test over nine independent random initializations. Bold values denote statistically significant differences ($p<0.05$). Metrics P, S, and L refer to PSNR, SSIM, and LPIPS, respectively.}
\begin{tabular}{ccc|ccc}
\toprule
$E$ & $B$ & Metric & APRIL & GradViT & GIAS \\
\midrule
1 & 8 & P & \cellcolor{red!20}0.548 & \cellcolor{red!20}0.622 & \cellcolor{green!20}\textbf{0.004} \\
1 & 8 & S & \cellcolor{red!20}0.064 & \cellcolor{red!20}1.000 & \cellcolor{green!20}\textbf{0.003} \\
1 & 8 & L & \cellcolor{red!20}0.758 & \cellcolor{red!20}1.000 & \cellcolor{green!20}\textbf{0.011} \\
1 & 4 & P & \cellcolor{red!20}0.069 & \cellcolor{red!20}0.536 & \cellcolor{green!20}\textbf{0.002} \\
1 & 4 & S & \cellcolor{red!20}\textbf{0.016} & \cellcolor{red!20}0.541 & \cellcolor{green!20}\textbf{0.001} \\
1 & 4 & L & \cellcolor{red!20}0.272 & \cellcolor{red!20}1.000 & \cellcolor{green!20}\textbf{0.003} \\
2 & 8 & P & \cellcolor{red!20}0.379 & \cellcolor{red!20}0.400 & \cellcolor{green!20}\textbf{0.002} \\
2 & 8 & S & \cellcolor{red!20}0.101 & \cellcolor{red!20}0.085 & \cellcolor{green!20}\textbf{0.001} \\
2 & 8 & L & \cellcolor{red!20}0.602 & \cellcolor{red!20}1.000 & \cellcolor{green!20}\textbf{0.005} \\
2 & 4 & P & \cellcolor{red!20}0.799 & \cellcolor{red!20}0.464 & \cellcolor{green!20}\textbf{5.19e-04} \\
2 & 4 & S & \cellcolor{red!20}0.162 & \cellcolor{red!20}0.267 & \cellcolor{green!20}\textbf{2.92e-04} \\
2 & 4 & L & \cellcolor{red!20}0.100 & \cellcolor{red!20}0.130 & \cellcolor{green!20}\textbf{0.003} \\
\bottomrule
\end{tabular}
\label{tab:welchttestmultistep}
\end{table}

\section{More Results on Large Batch Sizes on CelebA dataset}
\label{sec:celebalargebs}

\begin{figure}[!ht]
\centering
\scriptsize
\setlength{\tabcolsep}{1pt}

\begin{tabular}{c cccccccc}
\toprule
\textbf{Method}
& \textbf{1} & \textbf{2} & \textbf{3} & \textbf{4} & \textbf{5} & \textbf{6} & \textbf{7} & \textbf{8} \\
\midrule
\multicolumn{9}{c}{Batch Size $B=4$}\\\midrule

GT
& \includegraphics[width=0.09\linewidth]{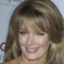}
& \includegraphics[width=0.09\linewidth]{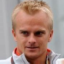}
& \includegraphics[width=0.09\linewidth]{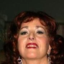}
& \includegraphics[width=0.09\linewidth]{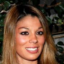}
& \includegraphics[width=0.09\linewidth]{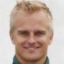}
& \includegraphics[width=0.09\linewidth]{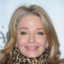}
& \includegraphics[width=0.09\linewidth]{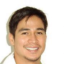}
& \includegraphics[width=0.09\linewidth]{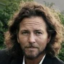}
\\
GIAS
& \includegraphics[width=0.09\linewidth]{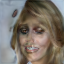}
& \includegraphics[width=0.09\linewidth]{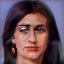}
& \includegraphics[width=0.09\linewidth]{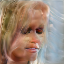}
& \includegraphics[width=0.09\linewidth]{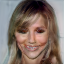}
& \includegraphics[width=0.09\linewidth]{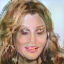}
& \includegraphics[width=0.09\linewidth]{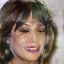}
& \includegraphics[width=0.09\linewidth]{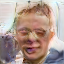}
& \includegraphics[width=0.09\linewidth]{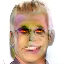}
\\
GIAS + Ours
& \includegraphics[width=0.09\linewidth]{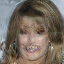}
& \includegraphics[width=0.09\linewidth]{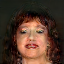}
& \includegraphics[width=0.09\linewidth]{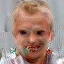}
& \includegraphics[width=0.09\linewidth]{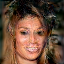}
& \includegraphics[width=0.09\linewidth]{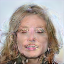}
& \includegraphics[width=0.09\linewidth]{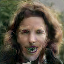}
& \includegraphics[width=0.09\linewidth]{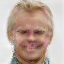}
& \includegraphics[width=0.09\linewidth]{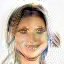}
\\
\midrule

GT
& \includegraphics[width=0.09\linewidth]{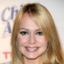}
& \includegraphics[width=0.09\linewidth]{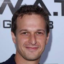}
& \includegraphics[width=0.09\linewidth]{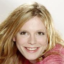}
& \includegraphics[width=0.09\linewidth]{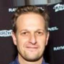}
& \includegraphics[width=0.09\linewidth]{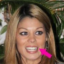}
& \includegraphics[width=0.09\linewidth]{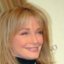}
& \includegraphics[width=0.09\linewidth]{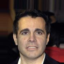}
& \includegraphics[width=0.09\linewidth]{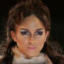}
\\
GIAS
& \includegraphics[width=0.09\linewidth]{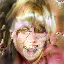}
& \includegraphics[width=0.09\linewidth]{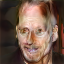}
& \includegraphics[width=0.09\linewidth]{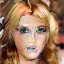}
& \includegraphics[width=0.09\linewidth]{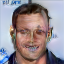}
& \includegraphics[width=0.09\linewidth]{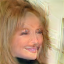}
& \includegraphics[width=0.09\linewidth]{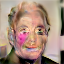}
& \includegraphics[width=0.09\linewidth]{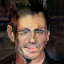}
& \includegraphics[width=0.09\linewidth]{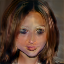}
\\
GIAS + Ours
& \includegraphics[width=0.09\linewidth]{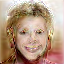}
& \includegraphics[width=0.09\linewidth]{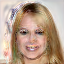}
& \includegraphics[width=0.09\linewidth]{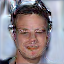}
& \includegraphics[width=0.09\linewidth]{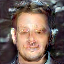}
& \includegraphics[width=0.09\linewidth]{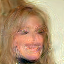}
& \includegraphics[width=0.09\linewidth]{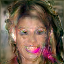}
& \includegraphics[width=0.09\linewidth]{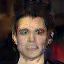}
& \includegraphics[width=0.09\linewidth]{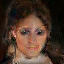}
\\
\bottomrule
\end{tabular}

\begin{tabular}{c cccccccc}
\toprule
\textbf{Method}
& \textbf{1} & \textbf{2} & \textbf{3} & \textbf{4} & \textbf{5} & \textbf{6} & \textbf{7} & \textbf{8} \\
\midrule
\multicolumn{9}{c}{Batch Size $B=16$}\\\midrule

GT
& \includegraphics[width=0.09\linewidth]{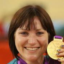}
& \includegraphics[width=0.09\linewidth]{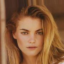}
& \includegraphics[width=0.09\linewidth]{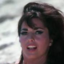}
& \includegraphics[width=0.09\linewidth]{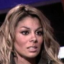}
& \includegraphics[width=0.09\linewidth]{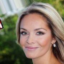}
& \includegraphics[width=0.09\linewidth]{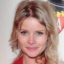}
& \includegraphics[width=0.09\linewidth]{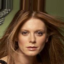}
& \includegraphics[width=0.09\linewidth]{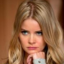}
\\
& \includegraphics[width=0.09\linewidth]{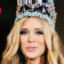}
& \includegraphics[width=0.09\linewidth]{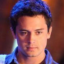}
& \includegraphics[width=0.09\linewidth]{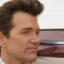}
& \includegraphics[width=0.09\linewidth]{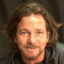}
& \includegraphics[width=0.09\linewidth]{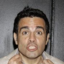}
& \includegraphics[width=0.09\linewidth]{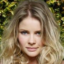}
& \includegraphics[width=0.09\linewidth]{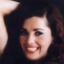}
& \includegraphics[width=0.09\linewidth]{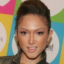}
\\
\midrule

GIAS
& \includegraphics[width=0.09\linewidth]{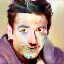}
& \includegraphics[width=0.09\linewidth]{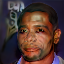}
& \includegraphics[width=0.09\linewidth]{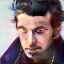}
& \includegraphics[width=0.09\linewidth]{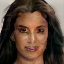}
& \includegraphics[width=0.09\linewidth]{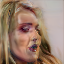}
& \includegraphics[width=0.09\linewidth]{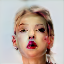}
& \includegraphics[width=0.09\linewidth]{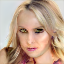}
& \includegraphics[width=0.09\linewidth]{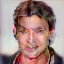}
\\
& \includegraphics[width=0.09\linewidth]{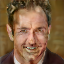}
& \includegraphics[width=0.09\linewidth]{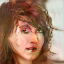}
& \includegraphics[width=0.09\linewidth]{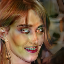}
& \includegraphics[width=0.09\linewidth]{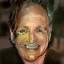}
& \includegraphics[width=0.09\linewidth]{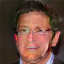}
& \includegraphics[width=0.09\linewidth]{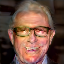}
& \includegraphics[width=0.09\linewidth]{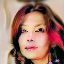}
& \includegraphics[width=0.09\linewidth]{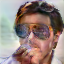}
\\
\midrule

GIAS + Ours
& \includegraphics[width=0.09\linewidth]{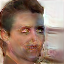}
& \includegraphics[width=0.09\linewidth]{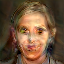}
& \includegraphics[width=0.09\linewidth]{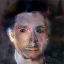}
& \includegraphics[width=0.09\linewidth]{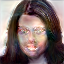}
& \includegraphics[width=0.09\linewidth]{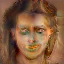}
& \includegraphics[width=0.09\linewidth]{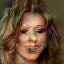}
& \includegraphics[width=0.09\linewidth]{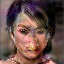}
& \includegraphics[width=0.09\linewidth]{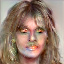}
\\
& \includegraphics[width=0.09\linewidth]{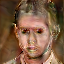}
& \includegraphics[width=0.09\linewidth]{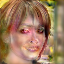}
& \includegraphics[width=0.09\linewidth]{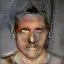}
& \includegraphics[width=0.09\linewidth]{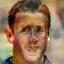}
& \includegraphics[width=0.09\linewidth]{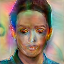}
& \includegraphics[width=0.09\linewidth]{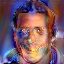}
& \includegraphics[width=0.09\linewidth]{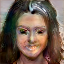}
& \includegraphics[width=0.09\linewidth]{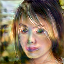}
\\
\bottomrule
\end{tabular}

\caption{Qualitative reconstruction comparison on the CelebA dataset. Top: Batch Size $B=4$. Bottom: Batch Size $B=16$.}
\label{fig:celeba_qual_grid_largebs}
\end{figure}

In this section, we present empirical quantitative results for batch sizes 1, 4, 16 and 32 in Tab. \ref{tab:celebabstab} and also qualitative reconstruction results on larger batch sizes of 4 and 16 in Fig. \ref{fig:celeba_qual_grid_largebs}. For this ablation, we used CelebA images of resolution $64 \times 64$. We observe that even on these larger resolution images, the proposed modification to the transformer architecture, that is, incorporation of trainable positional embedding improves performance for large batch sizes over the baseline GIAS.

\begin{table}[!ht]
    \centering
    \scriptsize
    \caption{Comparison of performance for multiple batch sizes on $64 \times 64$ CelebA dataset.}
    \resizebox{\linewidth}{!}{
    \begin{tabular}{c|ccc|ccc}
    \toprule
         \multirow{3}{*}{Method} & \multicolumn{6}{c}{Batch Size}\\ \cmidrule{2-7}
         & \multicolumn{3}{c|}{1} & \multicolumn{3}{c}{4}\\\cmidrule{2-7} 
         & PSNR & SSIM & LPIPS
         & PSNR & SSIM & LPIPS\\ \midrule
        GIAS 
        & 12.48 & 0.300 & 0.198
        & 11.00 & 0.192 & 0.238 \\
        GIAS+Ours 
        & \textbf{22.44} & \textbf{0.752} & \textbf{0.064}
        & \textbf{17.25} & \textbf{0.470} & \textbf{0.139} \\ \midrule
        & \multicolumn{3}{c|}{16} & \multicolumn{3}{c}{32} \\\cmidrule{2-7}
         & PSNR & SSIM & LPIPS
         & PSNR & SSIM & LPIPS\\ \midrule
         GIAS
         & 10.80 & 0.181 & 0.236
        & 10.63 & 0.178 & 0.225\\
        GIAS+Ours
        & \textbf{12.79} & \textbf{0.239} & \textbf{0.203}
        & \textbf{11.89} & \textbf{0.190} & \textbf{0.221}\\ \midrule
        \bottomrule
    \end{tabular}
    }
    \label{tab:celebabstab}
\end{table}

\section{Impact of Low Rank Adapter Initialization}
\label{sec:lorainit}

We further study the effect on initialization of the Low Rank adapters, for which we have three choices: (1) standard, that is, initializing $A \sim \mathcal{U}\left[-\frac{1}{\sqrt{d}},\frac{1}{\sqrt{d}}\right]$, and $B$ with zeros, (2) initializing both $A, B \sim \mathcal{U}\left[-\frac{1}{\sqrt{d}},\frac{1}{\sqrt{d}}\right]$, %
(3) initializing $A$ and $B$ following the strategy described in LoRA-One \cite{Zhang2025loraone}. Tab. \ref{tab:lorainit} presents the effect of LoRA initialization on the attack metrics for the single-step update scenario with batch size 1 for LoRA rank 128. We observe that while applying uniform initialization for both the adapters definitely improves performance for both GIAS and GIAS+Ours, the same is not true for LoRA-One initialization. In fact, we observe that using LoRA-One initialization in our case yields sub-optimal performance for both cases.

\begin{table}[!ht]
\centering
\scriptsize
\caption{Effect of different initialization strategies on reconstruction performance of GIAS.}
\setlength{\tabcolsep}{1pt}
\begin{tabular}{l|ccc|ccc}
\toprule
\multirow{2}{*}{\textbf{Initialization}} & \multicolumn{3}{c|}{GIAS} & \multicolumn{3}{c}{GIAS+Ours} \\ \cmidrule{2-7}
& \textbf{PSNR} $\uparrow$ & \textbf{SSIM} $\uparrow$ & \textbf{LPIPS} $\downarrow$ & \textbf{PSNR} $\uparrow$ & \textbf{SSIM} $\uparrow$ & \textbf{LPIPS} $\downarrow$ \\
\midrule
Standard    &  \makecell{13.48 \tiny{(3.74)}} & \makecell{0.24 \tiny{(0.19)}}  & \makecell{0.126 \tiny{(0.077)}} & \makecell{23.47 \tiny{(6.04)}} & \makecell{0.77 \tiny{(0.20)}} & \makecell{0.025 \tiny{(0.031)}} \\
Uniform     &  \makecell{15.77 \tiny{(4.52)}} & \makecell{0.34 \tiny{(0.221)}} & \makecell{0.12 \tiny{(0.075)}} & \makecell{25.91 \tiny{(6.29)}} &  \makecell{0.79 \tiny{(0.17)}} & \makecell{0.022 \tiny{(0.033)}} \\
LoRA-One \cite{Zhang2025loraone}    & \makecell{10.32 \tiny{(1.62)}} & \makecell{0.08 \tiny{(0.05)}} & \makecell{0.198 \tiny{(0.064)}} & \makecell{10.32 \tiny{(1.44)}} & \makecell{0.07 \tiny{(0.05)}} & \makecell{0.197 \tiny{(0.064)}} \\
\bottomrule
\end{tabular}
\label{tab:lorainit}
\end{table}

\section{Ablation on ViT Backbone Initialization}
\label{sec:bbinitabl}

In this section, we investigate the effect of model initialization on attack efficacy. Tab.~\ref{tab:vit_init_ablation} studies the influence of the ViT backbone initialization on reconstruction quality. When initialized from a CIFAR-pretrained checkpoint, both GIAS \cite{gias} and GIAS+Ours achieve relatively modest reconstruction performance, with our method providing only a limited improvement over the baseline. In contrast, initializing the backbone from an ImageNet-pretrained checkpoint substantially improves the performance of GIAS+Ours, increasing PSNR from 17.67 to 25.91 dB and SSIM from 0.468 to 0.79 while reducing LPIPS from 0.174 to 0.022. Although the baseline GIAS also benefits from ImageNet pretraining, the gain is considerably smaller. These results suggest that richer semantic representations learned through large-scale pretraining provide a more informative optimization landscape for our trainable positional embeddings, allowing the proposed method to exploit pretrained features more effectively than the baseline.

\begin{table}[!ht]
\centering
\caption{Ablation on the ViT backbone initialization. The generator initialization is kept fixed across all experiments.}
\label{tab:vit_init_ablation}
\small
\setlength{\tabcolsep}{4pt}
\begin{tabular}{ccccc}
\toprule
\textbf{ViT Init.} & \textbf{Method} & \textbf{PSNR} $\uparrow$ & \textbf{SSIM} $\uparrow$ & \textbf{LPIPS} $\downarrow$ \\
\midrule
\multirow{2}{*}{CIFAR} & GIAS   & 
\makecell{15.08 \\ \scriptsize{(7.51)}} & 
\makecell{0.356 \\ \scriptsize{(0.312)}} & 
\makecell{0.187 \\ \scriptsize{(0.145)}} \\
 & GIAS+Ours  & 
 \makecell{17.67 \\ \scriptsize{(9.195)}} & 
 \makecell{0.468 \\ \scriptsize{(0.382)}} & 
 \makecell{0.174 \\ \scriptsize{(0.161)}} \\ \midrule
\multirow{2}{*}{ImageNet} & GIAS  & 
\makecell{15.77 \\ \scriptsize{(4.52)}} &
\makecell{0.34 \\ \scriptsize{(0.221)}} &
\makecell{0.12 \\ \scriptsize{(0.075)}} \\
 & GIAS+Ours & 
 \makecell{25.91 \\ \scriptsize{(6.29)}} &
\makecell{0.79 \\ \scriptsize{(0.17)}} &
\makecell{0.022 \\ \scriptsize{(0.033)}} \\
\bottomrule
\end{tabular}
\end{table}

\section{Impact of Number of Steps in Server on Attack Performance}
\label{sec:stepsabl}

\begin{figure}[!ht]
    \centering
    \includegraphics[width=\linewidth]{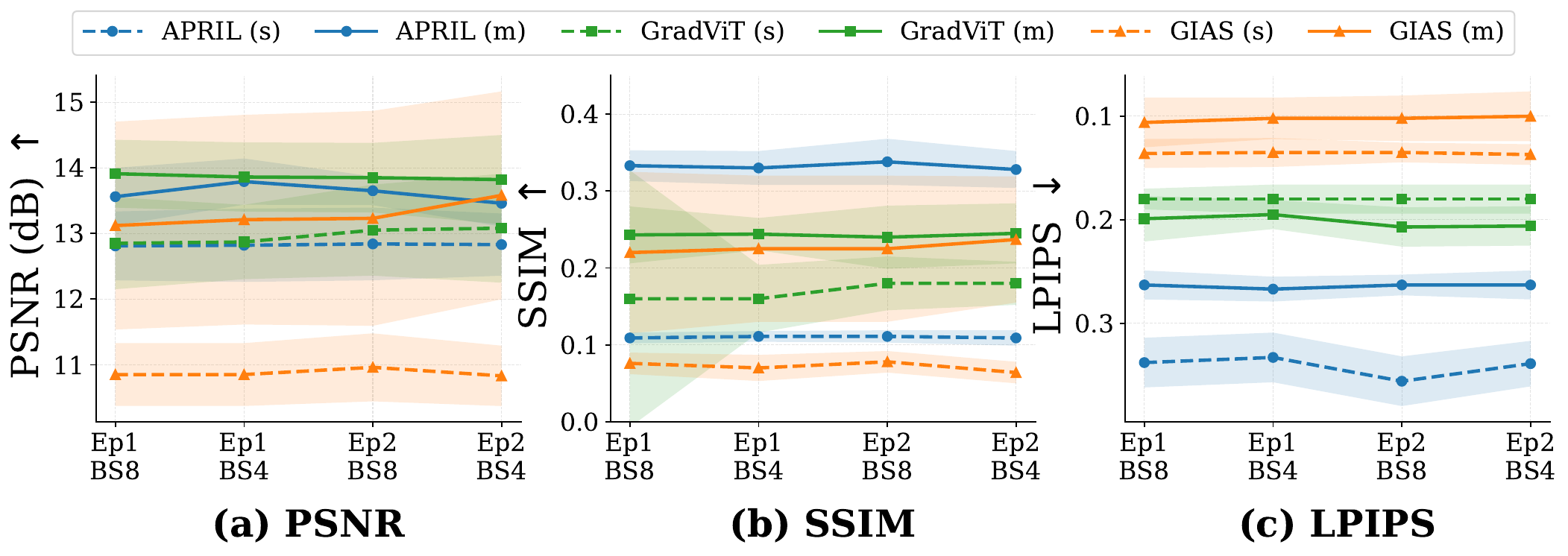}
    \caption{Ablation on the effect of the number of steps in the server. Parentheses denote standard deviation. $\Delta$ tracks performance change from (s) to (m). "Ep $x$ BS $y$" denotes that the optimization in the user was conducted for $x$ epochs with a batch size of $y$, before the gradient was sent to the server.}
    \label{fig:steps_results_full}
\end{figure}

In this section, we ablate the effect of performing single-step versus multi-step optimization in the server-side attacker. Here, single-step (s) and multi-step (m) denote the number of optimization steps performed by the attacker to emulate the client's local updates. From Fig. \ref{fig:steps_results_full}, %
we observe that moving from (s) to (m) consistently improves all metrics for GIAS: PSNR increases by 2.27-2.75 dB, SSIM increases by 0.144-0.173, and LPIPS decreases by 0.03-0.037. The improvement is stable across both one and two local epochs, and across batch sizes 4 and 8. Similar improvements are also observed for APRIL and GradViT. This indicates that multi-step updates provide better estimation of the target client gradients.

\section{Qualitative Evidence of the Effectiveness of the Trainable Positional Embeddings for Different LoRA Ranks}
\label{sec:higherlorarank}

\begin{figure}[!ht]
\centering
\scriptsize
\setlength{\tabcolsep}{1pt}

\begin{tabular}{c cccccccc}
\toprule
\textbf{Method}
& \textbf{1}
& \textbf{2}
& \textbf{3}
& \textbf{4}
& \textbf{5}
& \textbf{6}
& \textbf{7}
& \textbf{8}\\
\midrule
\multicolumn{9}{c}{LoRA Rank $r=64$}\\\midrule

GT
& \includegraphics[width=0.09\linewidth]{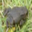}
& \includegraphics[width=0.09\linewidth]{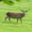}
& \includegraphics[width=0.09\linewidth]{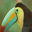}
& \includegraphics[width=0.09\linewidth]{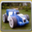}
& \includegraphics[width=0.09\linewidth]{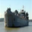}
& \includegraphics[width=0.09\linewidth]{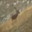}
& \includegraphics[width=0.09\linewidth]{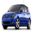}
& \includegraphics[width=0.09\linewidth]{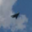}
\\

GIAS
& \includegraphics[width=0.09\linewidth]{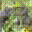}
& \includegraphics[width=0.09\linewidth]{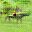}
& \includegraphics[width=0.09\linewidth]{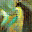}
& \includegraphics[width=0.09\linewidth]{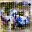}
& \includegraphics[width=0.09\linewidth]{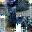}
& \includegraphics[width=0.09\linewidth]{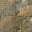}
& \includegraphics[width=0.09\linewidth]{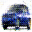}
& \includegraphics[width=0.09\linewidth]{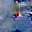}
\\

GIAS + Ours
& \includegraphics[width=0.09\linewidth]{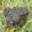}
& \includegraphics[width=0.09\linewidth]{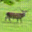}
& \includegraphics[width=0.09\linewidth]{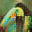}
& \includegraphics[width=0.09\linewidth]{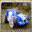}
& \includegraphics[width=0.09\linewidth]{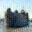}
& \includegraphics[width=0.09\linewidth]{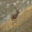}
& \includegraphics[width=0.09\linewidth]{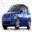}
& \includegraphics[width=0.09\linewidth]{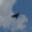}
\\ \midrule

GT
& \includegraphics[width=0.09\linewidth]{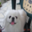}
& \includegraphics[width=0.09\linewidth]{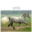}
& \includegraphics[width=0.09\linewidth]{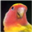}
& \includegraphics[width=0.09\linewidth]{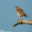}
& \includegraphics[width=0.09\linewidth]{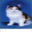}
& \includegraphics[width=0.09\linewidth]{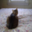}
& \includegraphics[width=0.09\linewidth]{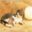}
& \includegraphics[width=0.09\linewidth]{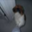}
\\

GIAS
& \includegraphics[width=0.09\linewidth]{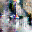}
& \includegraphics[width=0.09\linewidth]{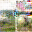}
& \includegraphics[width=0.09\linewidth]{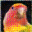}
& \includegraphics[width=0.09\linewidth]{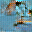}
& \includegraphics[width=0.09\linewidth]{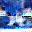}
& \includegraphics[width=0.09\linewidth]{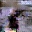}
& \includegraphics[width=0.09\linewidth]{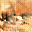}
& \includegraphics[width=0.09\linewidth]{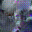}
\\

GIAS+Ours
& \includegraphics[width=0.09\linewidth]{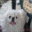}
& \includegraphics[width=0.09\linewidth]{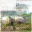}
& \includegraphics[width=0.09\linewidth]{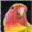}
& \includegraphics[width=0.09\linewidth]{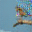}
& \includegraphics[width=0.09\linewidth]{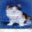}
& \includegraphics[width=0.09\linewidth]{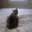}
& \includegraphics[width=0.09\linewidth]{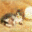}
& \includegraphics[width=0.09\linewidth]{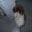}
\\

\midrule
\multicolumn{9}{c}{LoRA Rank $r=32$}\\\midrule
GT
& \includegraphics[width=0.09\linewidth]{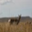}
& \includegraphics[width=0.09\linewidth]{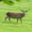}
& \includegraphics[width=0.09\linewidth]{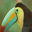}
& \includegraphics[width=0.09\linewidth]{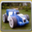}
& \includegraphics[width=0.09\linewidth]{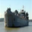}
& \includegraphics[width=0.09\linewidth]{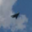}
& \includegraphics[width=0.09\linewidth]{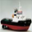}
& \includegraphics[width=0.09\linewidth]{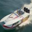}
\\

GIAS
& \includegraphics[width=0.09\linewidth]{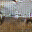}
& \includegraphics[width=0.09\linewidth]{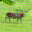}
& \includegraphics[width=0.09\linewidth]{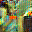}
& \includegraphics[width=0.09\linewidth]{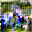}
& \includegraphics[width=0.09\linewidth]{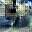}
& \includegraphics[width=0.09\linewidth]{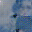}
& \includegraphics[width=0.09\linewidth]{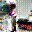}
& \includegraphics[width=0.09\linewidth]{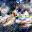}
\\

GIAS + Ours
& \includegraphics[width=0.09\linewidth]{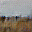}
& \includegraphics[width=0.09\linewidth]{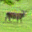}
& \includegraphics[width=0.09\linewidth]{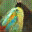}
& \includegraphics[width=0.09\linewidth]{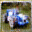}
& \includegraphics[width=0.09\linewidth]{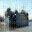}
& \includegraphics[width=0.09\linewidth]{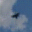}
& \includegraphics[width=0.09\linewidth]{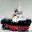}
& \includegraphics[width=0.09\linewidth]{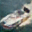}
\\ \midrule

GT
& \includegraphics[width=0.09\linewidth]{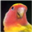}
& \includegraphics[width=0.09\linewidth]{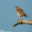}
& \includegraphics[width=0.09\linewidth]{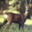}
& \includegraphics[width=0.09\linewidth]{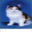}
& \includegraphics[width=0.09\linewidth]{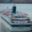}
& \includegraphics[width=0.09\linewidth]{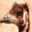}
& \includegraphics[width=0.09\linewidth]{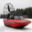}
& \includegraphics[width=0.09\linewidth]{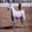}
\\

GIAS
& \includegraphics[width=0.09\linewidth]{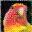}
& \includegraphics[width=0.09\linewidth]{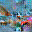}
& \includegraphics[width=0.09\linewidth]{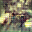}
& \includegraphics[width=0.09\linewidth]{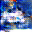}
& \includegraphics[width=0.09\linewidth]{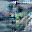}
& \includegraphics[width=0.09\linewidth]{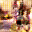}
& \includegraphics[width=0.09\linewidth]{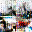}
& \includegraphics[width=0.09\linewidth]{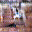}
\\

GIAS+Ours
& \includegraphics[width=0.09\linewidth]{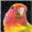}
& \includegraphics[width=0.09\linewidth]{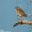}
& \includegraphics[width=0.09\linewidth]{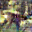}
& \includegraphics[width=0.09\linewidth]{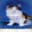}
& \includegraphics[width=0.09\linewidth]{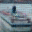}
& \includegraphics[width=0.09\linewidth]{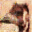}
& \includegraphics[width=0.09\linewidth]{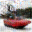}
& \includegraphics[width=0.09\linewidth]{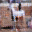}
\\
\bottomrule
\end{tabular}
\caption{Qualitative reconstruction comparison across 8 samples on the CIFAR-10 dataset with LoRA rank $r=64$ and $r=32$. We can observe that the proposed method is effective even at lower ranks and is capable of solving the spatial information obscuration issue for the backbone ViT-Small model.}
\label{fig:rankabl_qual_grid}
\end{figure}

In this section, we present qualitative empirical evidence of the attack performance when the backbone ViT is fine-tuned using LoRA with ranks $r=32$ and $r=64$ in Fig. \ref{fig:rankabl_qual_grid}. Consistent with the quantitative results presented in Sec.~\ref{subsubsec:lorarankabl}, the proposed method demonstrates improved reconstruction performance over the baseline GIAS method. In particular, the incorporation of trainable positional embeddings provides additional spatial information, enabling more accurate spatial reconstruction in cases where the baseline method fails to recover the underlying structure.

\section{Comparison of Performance under Secure Aggregation}
\label{sec:secagg}

\begin{figure}[!ht]
\centering
\scriptsize
\setlength{\tabcolsep}{1pt}

\begin{tabular}{c cccccccc}
\toprule
\textbf{Method}
& \textbf{1}
& \textbf{2}
& \textbf{3}
& \textbf{4}
& \textbf{5}
& \textbf{6}
& \textbf{7}
& \textbf{8} \\
\midrule
\multicolumn{9}{c}{Batch Size $B=1$ (4 Clients)}\\\midrule

GT
& \includegraphics[width=0.09\linewidth]{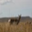}
& \includegraphics[width=0.09\linewidth]{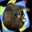}
& \includegraphics[width=0.09\linewidth]{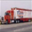}
& \includegraphics[width=0.09\linewidth]{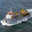}
& \includegraphics[width=0.09\linewidth]{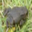}
& \includegraphics[width=0.09\linewidth]{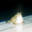}
& \includegraphics[width=0.09\linewidth]{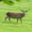}
& \includegraphics[width=0.09\linewidth]{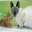}
\\
GIAS
& \includegraphics[width=0.09\linewidth]{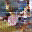}
& \includegraphics[width=0.09\linewidth]{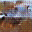}
& \includegraphics[width=0.09\linewidth]{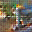}
& \includegraphics[width=0.09\linewidth]{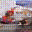}
& \includegraphics[width=0.09\linewidth]{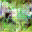}
& \includegraphics[width=0.09\linewidth]{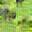}
& \includegraphics[width=0.09\linewidth]{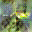}
& \includegraphics[width=0.09\linewidth]{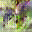}
\\
GIAS + Ours
& \includegraphics[width=0.09\linewidth]{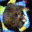}
& \includegraphics[width=0.09\linewidth]{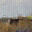}
& \includegraphics[width=0.09\linewidth]{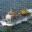}
& \includegraphics[width=0.09\linewidth]{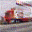}
& \includegraphics[width=0.09\linewidth]{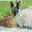}
& \includegraphics[width=0.09\linewidth]{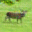}
& \includegraphics[width=0.09\linewidth]{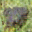}
& \includegraphics[width=0.09\linewidth]{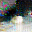}
\\
\midrule

GT
& \includegraphics[width=0.09\linewidth]{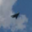}
& \includegraphics[width=0.09\linewidth]{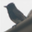}
& \includegraphics[width=0.09\linewidth]{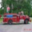}
& \includegraphics[width=0.09\linewidth]{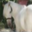}
& \includegraphics[width=0.09\linewidth]{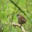}
& \includegraphics[width=0.09\linewidth]{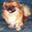}
& \includegraphics[width=0.09\linewidth]{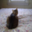}
& \includegraphics[width=0.09\linewidth]{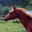}
\\
GIAS
& \includegraphics[width=0.09\linewidth]{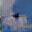}
& \includegraphics[width=0.09\linewidth]{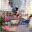}
& \includegraphics[width=0.09\linewidth]{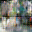}
& \includegraphics[width=0.09\linewidth]{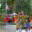}
& \includegraphics[width=0.09\linewidth]{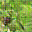}
& \includegraphics[width=0.09\linewidth]{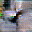}
& \includegraphics[width=0.09\linewidth]{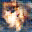}
& \includegraphics[width=0.09\linewidth]{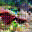}
\\
GIAS + Ours
& \includegraphics[width=0.09\linewidth]{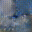}
& \includegraphics[width=0.09\linewidth]{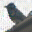}
& \includegraphics[width=0.09\linewidth]{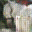}
& \includegraphics[width=0.09\linewidth]{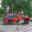}
& \includegraphics[width=0.09\linewidth]{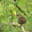}
& \includegraphics[width=0.09\linewidth]{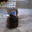}
& \includegraphics[width=0.09\linewidth]{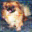}
& \includegraphics[width=0.09\linewidth]{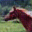}
\\
\bottomrule
\end{tabular}

\vspace{1.5em}

\begin{tabular}{c cccccccc}
\toprule
\textbf{Method}
& \textbf{1}
& \textbf{2}
& \textbf{3}
& \textbf{4}
& \textbf{5}
& \textbf{6}
& \textbf{7}
& \textbf{8} \\
\midrule
\multicolumn{9}{c}{Batch Size $B=4$ (4 Clients)}\\\midrule

GT
& \includegraphics[width=0.09\linewidth]{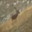}
& \includegraphics[width=0.09\linewidth]{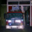}
& \includegraphics[width=0.09\linewidth]{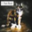}
& \includegraphics[width=0.09\linewidth]{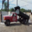}
& \includegraphics[width=0.09\linewidth]{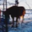}
& \includegraphics[width=0.09\linewidth]{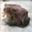}
& \includegraphics[width=0.09\linewidth]{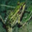}
& \includegraphics[width=0.09\linewidth]{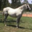}
\\
& \includegraphics[width=0.09\linewidth]{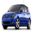}
& \includegraphics[width=0.09\linewidth]{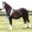}
& \includegraphics[width=0.09\linewidth]{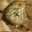}
& \includegraphics[width=0.09\linewidth]{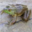}
& \includegraphics[width=0.09\linewidth]{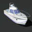}
& \includegraphics[width=0.09\linewidth]{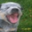}
& \includegraphics[width=0.09\linewidth]{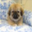}
& \includegraphics[width=0.09\linewidth]{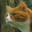}
\\
\midrule

GIAS
& \includegraphics[width=0.09\linewidth]{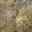}
& \includegraphics[width=0.09\linewidth]{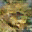}
& \includegraphics[width=0.09\linewidth]{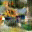}
& \includegraphics[width=0.09\linewidth]{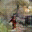}
& \includegraphics[width=0.09\linewidth]{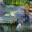}
& \includegraphics[width=0.09\linewidth]{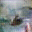}
& \includegraphics[width=0.09\linewidth]{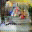}
& \includegraphics[width=0.09\linewidth]{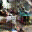}
\\
& \includegraphics[width=0.09\linewidth]{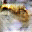}
& \includegraphics[width=0.09\linewidth]{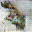}
& \includegraphics[width=0.09\linewidth]{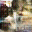}
& \includegraphics[width=0.09\linewidth]{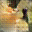}
& \includegraphics[width=0.09\linewidth]{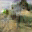}
& \includegraphics[width=0.09\linewidth]{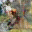}
& \includegraphics[width=0.09\linewidth]{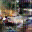}
& \includegraphics[width=0.09\linewidth]{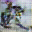}
\\
\midrule

GIAS + Ours
& \includegraphics[width=0.09\linewidth]{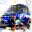}
& \includegraphics[width=0.09\linewidth]{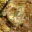}
& \includegraphics[width=0.09\linewidth]{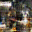}
& \includegraphics[width=0.09\linewidth]{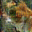}
& \includegraphics[width=0.09\linewidth]{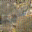}
& \includegraphics[width=0.09\linewidth]{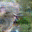}
& \includegraphics[width=0.09\linewidth]{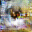}
& \includegraphics[width=0.09\linewidth]{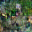}
\\
& \includegraphics[width=0.09\linewidth]{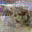}
& \includegraphics[width=0.09\linewidth]{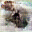}
& \includegraphics[width=0.09\linewidth]{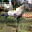}
& \includegraphics[width=0.09\linewidth]{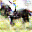}
& \includegraphics[width=0.09\linewidth]{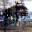}
& \includegraphics[width=0.09\linewidth]{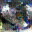}
& \includegraphics[width=0.09\linewidth]{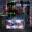}
& \includegraphics[width=0.09\linewidth]{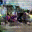}
\\
\bottomrule
\end{tabular}
\caption{Qualitative reconstruction comparison for Secure Aggregation scenarios. Top: Batch Size $B=1$ (in each client). Bottom: Batch Size $B=4$ (in each client).}
\label{fig:secagg_qual_grid}
\end{figure}

Secure Aggregation is a commonly used privacy-preserving mechanism in federated learning that prevents the server from directly observing the individual updates from each client. Instead, the server only receives the aggregated updates from multiple participating clients, making conventional gradient inversion attacks considerably more challenging. In this section, we evaluate the effectiveness of the proposed attack under the Secure Aggregation setting, where the objective is to reconstruct the data of the participating clients from the aggregated updates. We compare the performance of the proposed method with the baseline approaches on the CIFAR-10 dataset to investigate whether the proposed attack can effectively reconstruct client data despite the information loss introduced by aggregation, and provide the quantitative results in Tab. \ref{tab:c10secagg}, supported by qualitative evidence in Fig. \ref{fig:secagg_qual_grid}. For the experiments concerning Secure Aggregation, we consider the number of clients to be 4, and we ablate our investigation over two values of batch sizes, 1 and 4, in the single-step gradient update setup (FedSGD).

\begin{table}[!ht]
    \centering
    \scriptsize
    \caption{Quantitative comparison on CIFAR-10 under Secure Aggregation. The proposed trainable positional embeddings clearly benefit reconstruction even under secure aggregation.}
    \label{tab:c10secagg}
    \resizebox{\linewidth}{!}{
    \begin{tabular}{c|c|c|c|c|c|c}
    \toprule
     & \multicolumn{3}{c}{Batch Size 1} & \multicolumn{3}{c}{Batch Size 4} \\ \midrule
    Method & PSNR & SSIM & LPIPS& PSNR & SSIM & LPIPS\\\midrule
     GIAS   & 
     \makecell{11.56 \\ \tiny{(1.49)}}& 
     \makecell{0.132 \\ \tiny{(0.063)}}& 
     \makecell{0.135 \\ \tiny{(0.029)}}& 
     \makecell{11.08 \\ \tiny{(0.598)}}& 
     \makecell{0.08 \\ \tiny{(0.022)}}& 
     \makecell{0.136 \\ \tiny{(0.012)}} \\ \midrule
     GIAS+Ours  & 
     \makecell{\textbf{19.44} \\ \tiny{(4.28)}}& 
     \makecell{\textbf{0.593} \\ \tiny{(0.222)}}& 
     \makecell{\textbf{0.046} \\ \tiny{(0.036)}}& 
     \makecell{\textbf{13.516} \\ \tiny{(1.72)}}& 
     \makecell{\textbf{0.247} \\ \tiny{(0.106)}}& 
     \makecell{\textbf{0.1} \\ \tiny{(0.023)}} \\ \bottomrule
    \end{tabular}
    }
\end{table}

The most basic form of Secure Aggregation is implemented using \textit{pairwise masking} \cite{bonawitz2017randommasking}, where each pair of participating clients generates a shared random mask using a common random seed. Before transmitting their local gradients, each client adds or subtracts the corresponding pairwise masks from its gradient update. The masks are constructed such that, when the updates from all participating clients are aggregated at the server, the pairwise masks cancel each other out, leaving only the sum of the original client gradients. As a result, the server can obtain the aggregated update required for model training without directly observing the individual gradient updates of the participating clients.

On the CIFAR-10 dataset, we observe that the proposed method remains capable of recovering recognizable client samples from the aggregated updates, while achieving better reconstruction performance compared to the baseline methods. Although Secure Aggregation introduces additional ambiguity by combining the updates from multiple clients, the proposed method is able to leverage the learned generative model before constraining the reconstruction towards plausible data samples. As the number of aggregated client updates increases, the reconstruction problem becomes increasingly challenging due to the contribution of multiple client samples to the observed update. Nevertheless, the results demonstrate that the proposed method can still recover meaningful information from the aggregated gradients, highlighting that Secure Aggregation alone may not completely prevent data leakage under gradient inversion attacks.

\section{Comparison with Model Poisoning-based Attacks on Federated PEFT Scenario}
\label{sec:comppeftleak}
In this section, we compare our method with model poisoning methods that deliberately broadcast malicious weights to clients with the specific intention of reconstructing client data. A recent method that applies this objective to bottleneck-adapted Vision Transformers is PEFTLeak \cite{Sami2025gipeft}. PEFTLeak achieves this objective by specifically designing the bias and weights to facilitate the recovery of individual patches. However, such methods cannot maintain their attack performance when clients are allowed to update the transmitted weights, as evidenced by our experiments in Sec. \ref{sec:putradeoff}. In this section, we compare the performance of PEFTLeak with our proposed method on CIFAR datasets. 

\begin{table}[!ht]
    \centering
    \scriptsize
    \caption{Comparison of performance with PEFTLeak on the CIFAR-10 dataset for multiple batch sizes in the single-step update setting.}
    \resizebox{\linewidth}{!}{
    \begin{tabular}{c|ccc|ccc}
    \toprule
         \multirow{3}{*}{Method} & \multicolumn{6}{c}{Batch Size}\\ \cmidrule{2-7}
         & \multicolumn{3}{c|}{1} & \multicolumn{3}{c}{4} \\\cmidrule{2-7}
         & PSNR & SSIM & LPIPS
         & PSNR & SSIM & LPIPS\\ \midrule
        PEFTLeak 
        & 16.25 & \textbf{0.798} & 0.241 
        & 15.57 & \textbf{0.765} & 0.252 \\
        GIAS+Ours 
        & \textbf{25.91} & 0.79 & \textbf{0.022} 
        & \textbf{20.97} & 0.628 & \textbf{0.042}\\\midrule
        & \multicolumn{3}{c|}{16} & \multicolumn{3}{c}{32} \\\cmidrule{2-7}
        & PSNR & SSIM & LPIPS
         & PSNR & SSIM & LPIPS\\ \midrule
         PEFTLeak
         & \textbf{14.90} & \textbf{0.66} & 0.329
        & \textbf{14.36} & \textbf{0.585} & 0.384\\
        GIAS+Ours 
        & 13.71 & 0.244 & \textbf{0.103}
        & 11.83 & 0.110 & \textbf{0.121}\\ 
        \bottomrule
    \end{tabular}
    }
    \label{tab:vspeftleak}
\end{table}

From Tab. \ref{tab:vspeftleak}, we observe that the proposed method significantly outperforms PEFTLeak at low batch sizes in terms of PSNR and LPIPS in the single-step update setting, whereas PEFTLeak marginally outperforms the proposed method at higher batch sizes. However, PEFTLeak remains a less practical choice in such scenarios due to its detectability through the progression of the utility. This makes it difficult for PEFTLeak to maintain the observed attack performance throughout the entire federated training process.

\begin{figure}[!ht]
    \centering
    \resizebox{\linewidth}{!}{
    \begin{tabular}{ccc|ccc}
    \toprule
    GT & GIAS & GIAS+Ours & GT & GIAS & GIAS+Ours \\ \midrule
    \includegraphics[width=0.15\linewidth]{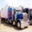}&\includegraphics[width=0.15\linewidth]{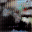}&\includegraphics[width=0.15\linewidth]{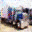}&
    \includegraphics[width=0.15\linewidth]{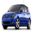}&\includegraphics[width=0.15\linewidth]{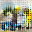}&\includegraphics[width=0.15\linewidth]{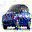}\\\midrule
    \includegraphics[width=0.15\linewidth]{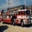}&\includegraphics[width=0.15\linewidth]{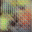}&\includegraphics[width=0.15\linewidth]{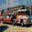}&
    \includegraphics[width=0.15\linewidth]{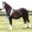}&\includegraphics[width=0.15\linewidth]{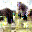}&\includegraphics[width=0.15\linewidth]{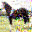}\\\midrule
    \includegraphics[width=0.15\linewidth]{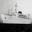}&\includegraphics[width=0.15\linewidth]{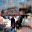}&\includegraphics[width=0.15\linewidth]{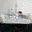}&
    \includegraphics[width=0.15\linewidth]{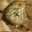}&\includegraphics[width=0.15\linewidth]{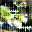}&\includegraphics[width=0.15\linewidth]{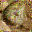}\\\midrule
    \includegraphics[width=0.15\linewidth]{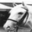}&\includegraphics[width=0.15\linewidth]{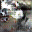}&\includegraphics[width=0.15\linewidth]{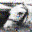}&
    \includegraphics[width=0.15\linewidth]{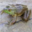}&\includegraphics[width=0.15\linewidth]{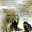}&\includegraphics[width=0.15\linewidth]{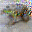}\\\midrule
    \includegraphics[width=0.15\linewidth]{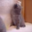}&\includegraphics[width=0.15\linewidth]{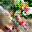}&\includegraphics[width=0.15\linewidth]{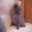}&
    \includegraphics[width=0.15\linewidth]{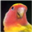}&\includegraphics[width=0.15\linewidth]{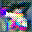}&\includegraphics[width=0.15\linewidth]{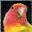}\\\midrule
    \includegraphics[width=0.15\linewidth]{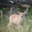}&\includegraphics[width=0.15\linewidth]{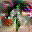}&\includegraphics[width=0.15\linewidth]{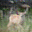}&
    \includegraphics[width=0.15\linewidth]{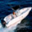}&\includegraphics[width=0.15\linewidth]{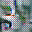}&\includegraphics[width=0.15\linewidth]{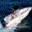}\\\midrule
    \includegraphics[width=0.15\linewidth]{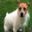}&\includegraphics[width=0.15\linewidth]{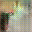}&\includegraphics[width=0.15\linewidth]{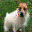}&
    \includegraphics[width=0.15\linewidth]{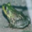}&\includegraphics[width=0.15\linewidth]{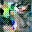}&\includegraphics[width=0.15\linewidth]{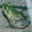}\\\midrule
    \includegraphics[width=0.15\linewidth]{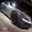}&\includegraphics[width=0.15\linewidth]{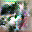}&\includegraphics[width=0.15\linewidth]{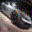}&
    \includegraphics[width=0.15\linewidth]{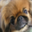}&\includegraphics[width=0.15\linewidth]{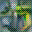}&\includegraphics[width=0.15\linewidth]{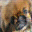}\\\midrule
    \includegraphics[width=0.15\linewidth]{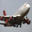}&\includegraphics[width=0.15\linewidth]{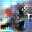}&\includegraphics[width=0.15\linewidth]{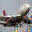}&
    \includegraphics[width=0.15\linewidth]{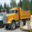}&\includegraphics[width=0.15\linewidth]{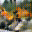}&\includegraphics[width=0.15\linewidth]{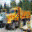}\\\midrule
    \includegraphics[width=0.15\linewidth]{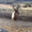}&\includegraphics[width=0.15\linewidth]{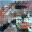}&\includegraphics[width=0.15\linewidth]{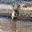}&
    \includegraphics[width=0.15\linewidth]{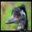}&\includegraphics[width=0.15\linewidth]{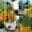}&\includegraphics[width=0.15\linewidth]{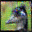}\\\midrule
    \includegraphics[width=0.15\linewidth]{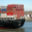}&\includegraphics[width=0.15\linewidth]{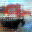}&\includegraphics[width=0.15\linewidth]{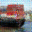}&
    \includegraphics[width=0.15\linewidth]{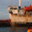}&\includegraphics[width=0.15\linewidth]{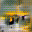}&\includegraphics[width=0.15\linewidth]{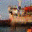}\\\midrule
    \includegraphics[width=0.15\linewidth]{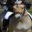}&\includegraphics[width=0.15\linewidth]{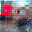}&\includegraphics[width=0.15\linewidth]{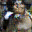}&
    \includegraphics[width=0.15\linewidth]{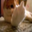}&\includegraphics[width=0.15\linewidth]{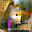}&\includegraphics[width=0.15\linewidth]{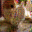}\\\bottomrule
    \end{tabular}
    }
    \caption{Qualitative evidence of reconstructions using LoRA gradients from ViT in SigLIP VLM client models.}
    \label{fig:vlm_attack}
\end{figure}

\section{Comparison of Attack Performance on VLMs}
\label{sec:vlmperf}

In this work, we primarily focus on vision transformer models, although there has also been substantial work in the language domain. However, work in the multimodal domain remains limited. Since our focus is primarily on the vision domain, we explore whether the proposed attack is also effective when used in conjunction with a language model. 

In this case as well, we assume that the server knows the language tokens fed to the language model at the client, either by knowing the class label along with the prompt template or by using an attack such as GRAB \cite{feng2024grab} or LAMP \cite{BalunovicD0V22lamp}. Furthermore, we use SigLIP \cite{Zhai2023siglip} as the multimodal model for training at the clients with a batch size of 4, resulting in 4 positive pairs and 12 negative pairs. For the attack simulation, we use the prompt ``\verb|This is an image of {class_name}|'' as the input to the text encoder, which gives us the text embeddings to compute the SigLIP loss. We can observe from the results presented in Fig. \ref{fig:vlm_attack} that the proposed method performs significantly better compared to the baseline method GIAS on the CIFAR-10 dataset.

\section{More Empirical Evidence of Privacy-Utility Trade-Off}
\label{sec:moreputoff}

In Sec. \ref{sec:putradeoff}, we had already provided empirical evidence through PSNR vs. Communication round and Server test accuracy plot. In this section, we present the plots of SSIM and LPIPS over the FedAvg \cite{mcmahan17fedavg} training duration at the 30$^{th}$, 60$^{th}$, 90$^{th}$ and 100$^{th}$ rounds. 

\begin{figure}[!ht]
    \centering
    \begin{minipage}[!ht]{1.0\linewidth}
        \centering
        \includegraphics[width=0.95\linewidth]{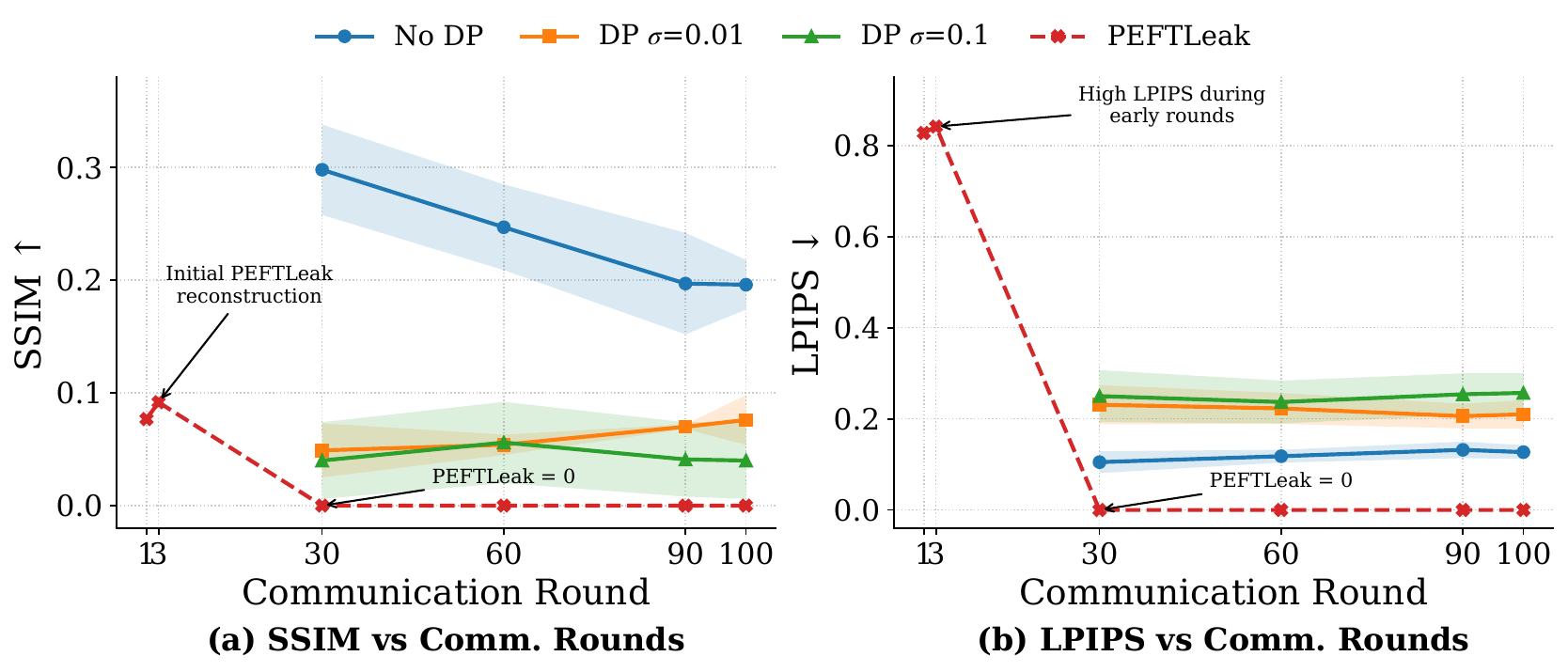}
        \caption{Higher SSIM and lower LPIPS indicate better reconstruction. We can observe that as the FedAvg training progresses, the attack performance with no defence actually drops, while the LPIPS remains the same over the training duration. The behaviour of PEFTLeak remains the same as PSNR and Accuracy for both SSIM and LPIPS over the whole training duration.}
        \label{fig:slputoff}
    \end{minipage}
\end{figure}

\section{Results on FEMNIST Dataset from LEAF benchmark}
\label{sec:femnistresults}

\begin{figure}[H]
\centering
\scriptsize
\setlength{\tabcolsep}{1pt}
\renewcommand{\arraystretch}{0.85}

\begin{tabular}{ccc|ccc}
\toprule
\textbf{GT} & \textbf{GIAS} & \makecell{\textbf{GIAS+}\textbf{Ours}} &
\textbf{GT} & \textbf{GIAS} & \makecell{\textbf{GIAS+}\textbf{Ours}} \\
\midrule

\includegraphics[width=0.14\linewidth,keepaspectratio]{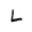} &
\includegraphics[width=0.14\linewidth,keepaspectratio]{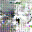} &
\includegraphics[width=0.14\linewidth,keepaspectratio]{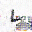} &
\includegraphics[width=0.14\linewidth,keepaspectratio]{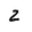} &
\includegraphics[width=0.14\linewidth,keepaspectratio]{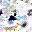} &
\includegraphics[width=0.14\linewidth,keepaspectratio]{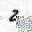} \\
\includegraphics[width=0.14\linewidth,keepaspectratio]{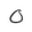} &
\includegraphics[width=0.14\linewidth,keepaspectratio]{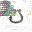} &
\includegraphics[width=0.14\linewidth,keepaspectratio]{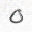} &
\includegraphics[width=0.14\linewidth,keepaspectratio]{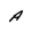} &
\includegraphics[width=0.14\linewidth,keepaspectratio]{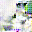} &
\includegraphics[width=0.14\linewidth,keepaspectratio]{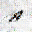} \\
\includegraphics[width=0.14\linewidth,keepaspectratio]{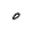} &
\includegraphics[width=0.14\linewidth,keepaspectratio]{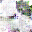} &
\includegraphics[width=0.14\linewidth,keepaspectratio]{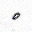} &
\includegraphics[width=0.14\linewidth,keepaspectratio]{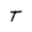} &
\includegraphics[width=0.14\linewidth,keepaspectratio]{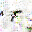} &
\includegraphics[width=0.14\linewidth,keepaspectratio]{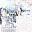} \\
\includegraphics[width=0.14\linewidth,keepaspectratio]{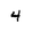} &
\includegraphics[width=0.14\linewidth,keepaspectratio]{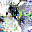} &
\includegraphics[width=0.14\linewidth,keepaspectratio]{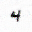} &
\includegraphics[width=0.14\linewidth,keepaspectratio]{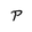} &
\includegraphics[width=0.14\linewidth,keepaspectratio]{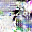} &
\includegraphics[width=0.14\linewidth,keepaspectratio]{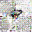} \\
\includegraphics[width=0.14\linewidth,keepaspectratio]{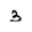} &
\includegraphics[width=0.14\linewidth,keepaspectratio]{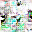} &
\includegraphics[width=0.14\linewidth,keepaspectratio]{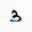} &
\includegraphics[width=0.14\linewidth,keepaspectratio]{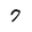} &
\includegraphics[width=0.14\linewidth,keepaspectratio]{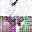} &
\includegraphics[width=0.14\linewidth,keepaspectratio]{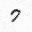} \\
\includegraphics[width=0.14\linewidth,keepaspectratio]{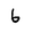} &
\includegraphics[width=0.14\linewidth,keepaspectratio]{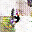} &
\includegraphics[width=0.14\linewidth,keepaspectratio]{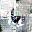} &
\includegraphics[width=0.14\linewidth,keepaspectratio]{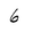} &
\includegraphics[width=0.14\linewidth,keepaspectratio]{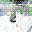} &
\includegraphics[width=0.14\linewidth,keepaspectratio]{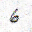} \\
\includegraphics[width=0.14\linewidth,keepaspectratio]{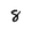} &
\includegraphics[width=0.14\linewidth,keepaspectratio]{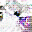} &
\includegraphics[width=0.14\linewidth,keepaspectratio]{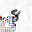} &
\includegraphics[width=0.14\linewidth,keepaspectratio]{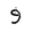} &
\includegraphics[width=0.14\linewidth,keepaspectratio]{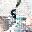} &
\includegraphics[width=0.14\linewidth,keepaspectratio]{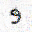} \\
\includegraphics[width=0.14\linewidth,keepaspectratio]{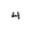} &
\includegraphics[width=0.14\linewidth,keepaspectratio]{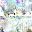} &
\includegraphics[width=0.14\linewidth,keepaspectratio]{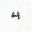} &
\includegraphics[width=0.14\linewidth,keepaspectratio]{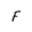} &
\includegraphics[width=0.14\linewidth,keepaspectratio]{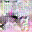} &
\includegraphics[width=0.14\linewidth,keepaspectratio]{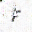} \\
\includegraphics[width=0.14\linewidth,keepaspectratio]{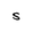} &
\includegraphics[width=0.14\linewidth,keepaspectratio]{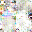} &
\includegraphics[width=0.14\linewidth,keepaspectratio]{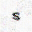} &
\includegraphics[width=0.14\linewidth,keepaspectratio]{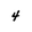} &
\includegraphics[width=0.14\linewidth,keepaspectratio]{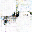} &
\includegraphics[width=0.14\linewidth,keepaspectratio]{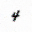} \\
\includegraphics[width=0.14\linewidth,keepaspectratio]{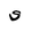} &
\includegraphics[width=0.14\linewidth,keepaspectratio]{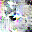} &
\includegraphics[width=0.14\linewidth,keepaspectratio]{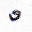} &
\includegraphics[width=0.14\linewidth,keepaspectratio]{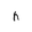} &
\includegraphics[width=0.14\linewidth,keepaspectratio]{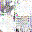} &
\includegraphics[width=0.14\linewidth,keepaspectratio]{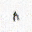} \\
\includegraphics[width=0.14\linewidth,keepaspectratio]{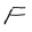} &
\includegraphics[width=0.14\linewidth,keepaspectratio]{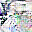} &
\includegraphics[width=0.14\linewidth,keepaspectratio]{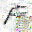} &
\includegraphics[width=0.14\linewidth,keepaspectratio]{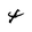} &
\includegraphics[width=0.14\linewidth,keepaspectratio]{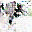} &
\includegraphics[width=0.14\linewidth,keepaspectratio]{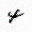} \\
\includegraphics[width=0.14\linewidth,keepaspectratio]{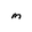} &
\includegraphics[width=0.14\linewidth,keepaspectratio]{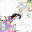} &
\includegraphics[width=0.14\linewidth,keepaspectratio]{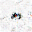} &
\includegraphics[width=0.14\linewidth,keepaspectratio]{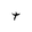} &
\includegraphics[width=0.14\linewidth,keepaspectratio]{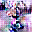} &
\includegraphics[width=0.14\linewidth,keepaspectratio]{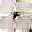} \\
\includegraphics[width=0.14\linewidth,keepaspectratio]{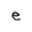} &
\includegraphics[width=0.14\linewidth,keepaspectratio]{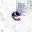} &
\includegraphics[width=0.14\linewidth,keepaspectratio]{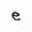} &
\includegraphics[width=0.14\linewidth,keepaspectratio]{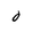} &
\includegraphics[width=0.14\linewidth,keepaspectratio]{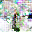} &
\includegraphics[width=0.14\linewidth,keepaspectratio]{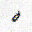} \\
\includegraphics[width=0.14\linewidth,keepaspectratio]{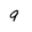} &
\includegraphics[width=0.14\linewidth,keepaspectratio]{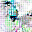} &
\includegraphics[width=0.14\linewidth,keepaspectratio]{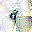} &
\includegraphics[width=0.14\linewidth,keepaspectratio]{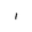} &
\includegraphics[width=0.14\linewidth,keepaspectratio]{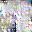} &
\includegraphics[width=0.14\linewidth,keepaspectratio]{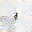} \\

\bottomrule
\end{tabular}

\caption{Qualitative comparison  of reconstruction on FEMNIST dataset between GIAS and GIAS + Ours.}
\label{fig:femnist_qualitative}
\end{figure}

We further evaluate the proposed spatial anchoring mechanism on the FEMNIST dataset from the LEAF benchmark. FEMNIST is a handwritten-character dataset that differs substantially from the natural image distribution used to train the DCGAN, making this setting a challenging out-of-distribution reconstruction task. We use a ViT-Small backbone adapted with low-rank matrices with rank $r = 128$ as the network shared by the server and the clients. In addition to that, we use a CIFAR-pretrained DCGAN generator as a prior. Qualitative comparisons of the reconstructions can be observed in Fig. \ref{fig:femnist_qualitative}, where we present multiple reconstruction samples from our experiments, showing clear evidence of reconstruction superiority compared to GIAS.

In addition to the qualitative results, the quantitative results in Tab. \ref{tab:femnist_results} show that the proposed method outperforms the baseline GIAS in the FedSGD setting in this challenging out-of-distribution task as well. The substantial improvement across PSNR, LPIPS, and SSIM indicates that the additional positional gradients remain useful even under a strong generator-data distribution mismatch.

\begin{table}[!ht]
\centering
\normalsize
\setlength{\tabcolsep}{1pt}
\caption{Comparison of performance on the FEMNIST dataset in the FedSGD setting.}
\label{tab:femnist_results}
\begin{tabular}{lccc}
\toprule
Method & PSNR $\uparrow$ & SSIM $\uparrow$ & LPIPS $\downarrow$ \\
\midrule
GIAS & \makecell{11.31 \\ \scriptsize{(3.27)}} & \makecell{0.180 \\ \scriptsize{(0.149)}} & \makecell{0.211 \\ \scriptsize{(0.082)}}\\
GIAS + Ours & \makecell{\textbf{22.11} \\ \scriptsize{(5.31)}} & \makecell{\textbf{0.662} \\ \scriptsize{(0.201)}} & \makecell{\textbf{0.033} \\ \scriptsize{(0.045)}}\\
\bottomrule
\end{tabular}
\end{table}

\end{document}